\documentclass[onecolumn]{cohere}

\usepackage{lipsum}

\usepackage{amsmath,amsfonts,bm}

\def\eqref#1{equation~\ref{#1}}

\def\1{\bm{1}}

\DeclareMathAlphabet{\mathsfit}{\encodingdefault}{\sfdefault}{m}{sl}
\SetMathAlphabet{\mathsfit}{bold}{\encodingdefault}{\sfdefault}{bx}{n}

\usepackage{stfloats} 
\usepackage{hyperref}
\usepackage{url}
\usepackage{booktabs}
\usepackage{graphicx}
\usepackage{multirow}
\usepackage{adjustbox}
\usepackage{float}
\usepackage{caption}
\usepackage{subcaption}
\usepackage[colorinlistoftodos]{todonotes}
\usepackage{lscape}
\usepackage{rotating}
\usepackage{enumitem}
\usepackage{wrapfig}
\usepackage{graphicx}
\usepackage{enumitem}
\usepackage{booktabs}
\usepackage{CJKutf8}
\usepackage{multirow}
\usepackage{amsfonts}
\usepackage{amsmath}
\usepackage{xcolor, colortbl}
\usepackage{multirow, hhline}
\usepackage{subcaption} 
\usepackage{arydshln}
\usepackage{siunitx}
\newcommand\blfootnote[1]{%
  \begingroup
  \renewcommand\thefootnote{}\footnote{#1}%
  \addtocounter{footnote}{-1}%
  \endgroup
}

\usepackage{nicematrix}

\title{RLHF Can Speak Many Languages: Unlocking Multilingual Preference Optimization for LLMs}

\author{
    name={John Dang},
    affiliation={Cohere For AI},
    email={johndang@cohere.com}
}
\author{
    name={Arash Ahmadian},
    affiliation={Cohere \& Cohere For AI},
    email={arash@cohere.com}
}
\author{
    name={Kelly Marchisio},
    affiliation={Cohere},
    email={kelly@cohere.com}
}
\author{
    name={Julia Kreutzer},
    affiliation={Cohere For AI},
    email={juliakreutzer@cohere.com}
}
\author{
    name={Ahmet Üstün},
    affiliation={Cohere For AI},
    email={ahmet@cohere.com}
}
\author{
    name={Sara Hooker},
    affiliation={Cohere For AI},
    email={sarahooker@cohere.com}
}

\date{\today}
\abstract{Preference optimization techniques have become a standard final stage for training state-of-art large language models (LLMs). However, despite widespread adoption, the vast majority of work to-date has focused on first-class citizen languages like English and Chinese. This captures a small fraction of the languages in the world, but also makes it unclear which aspects of current state-of-the-art research transfer to a multilingual setting. In this work, we perform an exhaustive study to achieve a new state-of-the-art in aligning multilingual LLMs. We introduce a novel, scalable method for generating high-quality multilingual feedback data to balance data coverage. We establish the benefits of cross-lingual transfer and increased dataset size in preference training. Our preference-trained model achieves a 54.4\% win-rate against Aya 23 8B, the current state-of-the-art multilingual LLM in its parameter class, and a 69.5\% win-rate or higher against widely used models like Gemma-1.1-7B-it, Llama-3-8B-Instruct, Mistral-7B-Instruct-v0.3. As a result of our study, we expand the frontier of alignment techniques to 23 languages covering  half of the world's population. 
}

\begin{document}
\blfootnote{Corresponding authors: \{\url{johndang}, \url{ahmet}, \url{sarahooker}\}\url{@cohere.com}}
\section{Introduction}
What languages are favored in technological progress is often deeply intertwined with historical patterns of technology access and resources~\citep{nekoto-etal-2020-participatory-forall, bird-2022-local,ayadata2024}. Preference optimization is a valuable and widely adopted post-training technique to align large language models (LLMs) with human preferences~\citep{NIPS2017_d5e2c0ad,stiennon2022learningRLHF,ouyang2022LLMRLHF,bai2022constitutional}which has been shown to lead to large gains in performance across a wide variety of NLP tasks \citep{wang2024mathshepherd,ivison2023camels,xu2024advancing, lightman2024lets}. To-date, the majority of progress in preference optimization has over-fit to a small handful of languages, resulting in large gaps in performance outside of English \citep{schwartz2022towards, Kotek2023GenderBA, Khandelwal2023CasteistBN, vashishtha2023evaluating,khondaker2023gptaraeval}, and also risks introducing security flaws for all users \citep{yong2024lowresource, nasr2023scalable, Li2023PrivacyIL, Lukas2023AnalyzingLO, deng2023multilingual}.

\begin{figure}[t]
  \centering
  \includegraphics[width=0.6\linewidth]{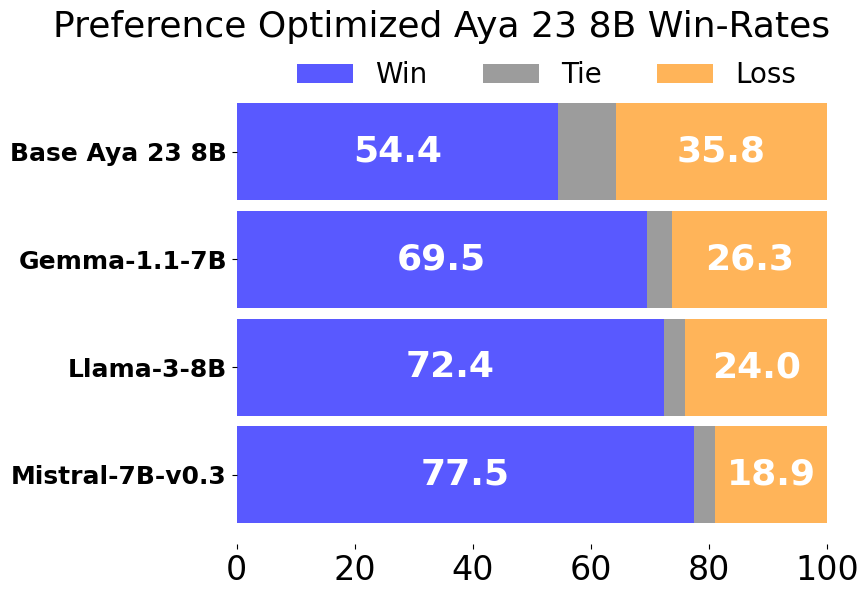}
  \caption{\textbf{Win-rates between our preference-trained model with the other state-of-the-art open weight models}: Averaged across 23 languages, our preference-trained model based on Aya-23-8B significantly outperforms the original Base Aya-23-8B, Gemma-1.1-7B-it, Meta-Llama3-8B-Instruct, and Mistral-7B-Instruct-v0.3.}
  \label{fig:win-rate-comp}
\end{figure}

While expanding the frontier of what languages are supported by AI is an increasingly urgent challenge, extending preference optimization to a multilingual setting is a non-trivial challenge. First, numerous works have shown that multilingual modeling typically faces both a data scarcity and data quality problem \citep{ayadata2024, ustun2024aya,dodge2021documenting, kreutzer-etal-2022-quality, luccioni-viviano-2021-whats}. This is even more pronounced for high-quality preference data which is virtually non-existent in many languages. Collecting multilingual preference datasets through human annotation is expensive and time intensive \citep{boubdir2023prompts, chaudhari2024rlhf}, and while prior works have proposed the use of LLMs to synthetically create preference datasets \citep{bai2022constitutional, yuan2024selfrewarding, pace2024westofn}, these efforts predominantly focus on English. The few efforts that have focused on multilinguality have relied on translation, introducing artifacts and resulting in a lack of diversity in preference pairs \citep{lai-etal-2023-okapi}, which is known to be critical to model performance \citep{naik2023diversity, Chung_2023, li2023making, lahoti2023improving,kirk2024understanding}.  


Second, preference optimization in many languages simultaneously is a difficult machine learning task. The lack of studies in preference optimization outside of English raises questions on how findings from monolingual optimization would transfer. Training dynamics of RLHF are known to be often unstable \citep{casper2023open, Gao2022ScalingLF,chaudhari2024rlhf}, which can be exacerbated by the involvement of multiple languages where preference data is from a heterogeneous distribution and negative transfer between languages is possible \citep{wang-etal-2020-negative,wang2019characterizing}. The few existing works on multilingual RLHF \citep{lai-etal-2023-okapi, lai2024llms} exhibit poor results and are outperformed by massively multilingual language models without any preference optimization \citep{ustun2024aya,aryabumi2024aya}.



The poor performance shown in the few existing multilingual preference optimization works begs the question: \emph{Is this a result of fundamental limitations with standard preference optimization techniques (especially in heterogeneous optimization settings like multilingual) or whether we are lacking high quality multilingual data?} 

In this work, we exhaustively study the aforementioned challenges. Our goal is to systematically understand key variables which might impact multilingual alignment, including the source and amount of available preference data, offline vs online RLHF techniques, and  the effect varying number of languages covered in preference optimization training data. We complete a comprehensive set of experiments with state-of-the-art alignment techniques including DPO~\citep{rafailov2023DPO} and RLOO~\citep{Kool2019Buy4R,ahmadian2024basics} starting from the 8-billion parameters instruction-finetuned Aya model covering 23 languages \citep[Aya-23-8B;][]{aryabumi2024aya}. Our findings can be summarized as follows:

\begin{enumerate}
    \item {
        \textbf{Preference optimization exhibits cross-lingual transfer}. We show that preference-optimization even with only English data improves performance in other languages. However, the addition of a few more languages significantly increases cross-lingual transfer, achieving win-rates on unseen languages up to 54.9\% when including 5 languages in training data compared to 46.3\% when training only on English.
    }

    \item {
        \textbf{Multilingual preference data is necessary for aligning multilingual LLMs}. We find that increasing the number of languages in preference optimization training data consistently improves multilingual performance compared to English-only training data, increasing win-rates by up to 7.0\% from 46.4\% to 53.4\% when all languages are included.
    }

    \item {
        \textbf{Online preference optimization outperforms offline optimization}. RLOO as an online method achieves better overall performance than DPO by a maximum 10.6\% difference in their average win-rates (54.4\% vs 43.8\%). Furthermore, we find that RLOO also enables better cross-lingual transfer, achieving up to 8.3\% increase over DPO in average win-rate on languages not included in training. 
    }

    \item {
        \textbf{Preference optimized Aya 23 8B outperforms other open weights models} Preference optimization leads to large gains in win-rates against both the original Aya model (54.4\% win-rate) and widely used open weights models including Gemma-1.1-7B-it \citep{gemmareport} (69.5\% win-rate), Meta-Llama3-8B-Instruct \citep{llama3modelcard} (72.4\% win-rate), Mistral-7b-Instruct-v0.3 (77.5\%win-rate) \citep{jiang2023mistral} averaged across all 23 languages as shown in Figure \ref{fig:win-rate-comp}.
    }
\end{enumerate}





\section{Methodology}

\subsection{Addressing Data Scarcity}

Prior works on multilingual preference training such as Okapi  \citep{lai-etal-2023-okapi}, typically involved using preference data translated from English, however, the Okapi models has since been outperformed by non-preference aligned models including the base Aya 23 8B model we experiment with here \citep{aryabumi2024aya}. While language coverage may be improved by translation \citep{ranaldi-pucci-2023-english,ustun2024aya}, the introduction of translation artifacts known as \textit{translationese}~\citep{bizzoni-etal-2020-human,vanmassenhove-etal-2021-machine} can hurt performance. Furthermore, repeatedly translating the same preference pairs can hurt preference diversity. The trade-off between increasing the amount of multilingual data and quality of multilingual data is difficult to isolate empirically ~\citep{yu-etal-2022-translate,dutta-chowdhury-etal-2022-towards}. We hypothesize that poor performance exhibited by prior works may be due to low quality data containing undesirable artifacts such as those introduced by translation.

Here, we attempt to avoid some of the issues with translation by creating preference pairs that intentionally aim to steer model generations away from \textit{translationese}.
First, we construct a diverse set of general instruction-following multilingual prompts by translating approximately 50K English prompts from ShareGPT\footnote{\url{https://sharegpt.com}} 
into the remaining 22 languages supported by Aya 23 8B. Automatic translation is done by using NLLB 3.3B \citep{nllb2022}.

\textbf{Source of completions} After translating prompts, we generate completions for each language by using multiple LLMs of varying multilingual capability. This ensures increased completion diversity compared to the alternative of simply translating the original English preference models. More specifically, we use Cohere's Command\footnote{\url{https://docs.cohere.com/docs/command-beta}} and Command R+\footnote{\url{https://docs.cohere.com/docs/command-r-plus}} models, where the latter is explicitly trained for multilingual performance\footnote{\url{https://docs.cohere.com/docs/command-r-plus\#unique-command-r-model-capabilities}}.  We note that Cohere’s terms of use\footnote{\url{https://cohere.com/terms-of-use}} prohibit training on model generations. However, we received a special exception for this work to use generations from Command models to train models. For Command, we generate English completions as the model is primarily proficient in English, and translate them into the other 22 languages. For Command R+, we generate a completion from the same prompt in-language.  This method enables obtaining a pair of multilingual competitions for each prompt with varying quality based on the difference in models' capabilities and the use of machine translation.


The use of \emph{translated} completions and comparing with high-quality \emph{direct} multilingual generations allows the model to steer away from translation artifacts. Note that the translated competitions are ranked as ``bad completions'' by 91\% hence, in most cases the preference ranking acts as a proxy label for translated completions.

\subsection{Offline vs Online Preference Training}

Reinforcement Learning from Human Feedback \citep[RLHF;][]{NIPS2017_d5e2c0ad,stiennon2022learningRLHF,ouyang2022LLMRLHF,bai2022constitutional}, which was proposed as the first framework for aligning language models to human preferences, has become a key for training state-of-the-art LLMs \citep{openai2023GPT4,touvron2023llama2,anthropic2024claude,reid2024gemini}. Canonically, PPO \citep{schulman2017proximal} has been used in RLHF as the online RL algorithm \citep{stiennon2022learningRLHF, ouyang2022instructgpt, nakano2021webgpt}. However, recent offline methods such as Direct Preference Optimization (DPO) \citep{rafailov2023DPO} and subsequent works in the same direction \citep{azar2024ipo,ethayarajh2024kto,choi2024srpo}, have proven increasingly popular due to reduced computational complexity. Traditional online methods such as PPO and REINFORCE~\citep{williams1992simple} require an additional network in addition to the policy, maintaining a reward model in memory, and also using the policy to generate doing training, all of which DPO does not require as it is fully offline.



A fractured experimental ecosystem and non-standardized datasets have made it difficult to evaluate the relative merits of both approaches comprehensively. However, recent work in an English setting \citep{tajwar2024preference,tang2024understanding} suggests that although the DPO-family of methods theoretically optimize the same objective as traditional RL algorithms, they under-perform compared to well-tuned traditional online RL methods \textit{due to the lack of online generations (on-policy or off-policy) and critique as provided by the reward model}. \citep{tajwar2024preference} also provides empirical evidence that explicit negative gradients during training improve over online methods without them. Given that multilingual datasets are far more heterogeneous than most datasets, it is unclear how these findings apply to massively multilingual settings.

To evaluate the impact usage of online and offline generations in multilingual preference optimization we benchmark \textbf{DPO} \citep{rafailov2023DPO} and  \textbf{REINFORCE-Leave-one-out} \citep[\textbf{RLOO};][]{Kool2019Buy4R,ahmadian2024basics}. \citet{ahmadian2024basics} shows that PPO may not be the right tool for RLHF, and that simpler REINFORCE-style methods such as Vanilla Policy Gradient and RLOO, are competitive or outperform PPO. In their experiments, RLOO outperforms both PPO and DPO, and incurs significantly lower computational overhead compared to PPO. Additionally, it has a contrastive loss by nature, which \citep{tajwar2024preference} improves learning compared to traditional RL. In Appendix Section \ref{appendix:rlhf-background}, we give a brief background and introduction to each method.

\textbf{Preference data mixtures} 
To evaluate if the multilingual preference data is essential in all languages and also to measure the cross-lingual transfer during preference training, we design various preference data mixtures where we control the number of languages covered and the amount of the preference data per language:   
\begin{enumerate}[wide, labelwidth=!, labelindent=0pt]
\item \underline{English-only}: English-only preference data mixture that includes 50K prompts with ranked generations. We term this this variant \textsc{EN-1-50K} and use it to understand \textbf{cross-lingual benefits that accrue from using only English data}. This is important, given it is the standard formulation of research to date.
\item\underline{5 language subset}: Multilingual mixture which includes English, Vietnamese, German, Turkish, and Portuguese with a total amount of 50K prompts (10K per language).
These 5 languages contain a mixture of higher and lower resource languages, with a range of language families and scripts. We refer to this variant as \textsc{ML-5-50K}. Training on only this mixture and evaluating on the remaining 18 languages allows us to measure the impact of multilingual preference data on \textbf{cross-lingual transfer to unseen languages} in comparison to English-only preference data.
\item\underline{All languages (fixed data budget)}: Multilingual mixture with all 23 languages supported by Aya 23 8B. This is our fixed budget variant where the total number of prompts is kept the same at 50K (approximately 2.2K examples per language) to compare with EN-1-50K and ML-23-50K. This variant \textbf{measures performance trade-offs when including all languages given the same data budget.} We refer to this variant as \textsc{ML-23-50K}.
\item\underline{All languages (not-fixed data budget)}: Our most comprehensive preference data mixture where we include all 23 languages with 10K prompts per language (230K in total). This mixture which we refer to as \textsc{ML-23-230K} enables us to evaluate the impact of a larger preference data budget in comparison with ML-23-50K. 
\end{enumerate}    

\section{Experimental Set-up}\label{sec:methodology}

\textbf{Multilingual Base Model} We perform all experiments with Aya 23 8B \citep{aryabumi2024aya} which is chosen because it  
\textbf{(1)} is massively multilingual, pre-trained and supervised fine-tuned for 23 languages, and
\textbf{(2)} it achieves state-of-the-art multilingual performance in 23 languages compared to other commonly used LLMs in its class, outperforming Mistral-7B-Instruct-v0.3 \citep{jiang2023mistral}, Gemma-1.1-7B-it \citep{gemmareport}, and Aya-101-13B \citep{ustun2024aya}. On multilingual benchmarks and open-ended generations, Aya 23 achieves a 65\% win-rate or higher in head-to-head comparisons with popular open sourcee models\citep{aryabumi2024aya}. Furthermore, Aya 23  is an open weights model that is not preference-trained, allowing us to isolate the impact of multilingual preference optimization in our experiments.

\setlength{\columnsep}{15pt}
\begin{wraptable}{R}{0.5\linewidth}
\centering
\vspace{-.5cm}
\begin{tabular}{lr}
\hline
\multicolumn{2}{c}{\textbf{Agreement between RM and GPT-4}} \\ 
\hline
\noalign{\smallskip} 
English           & 87.9\%                                         \\
Vietnamese        & 88.7\%                                         \\
Turkish           & 84.4\%                                         \\
Portuguese        & 91.0\%                                         \\
German            & 85.5\%                                         \\
        \hdashline 

Avg 23 Languages  & 87.3\%                                         \\ \hline
\end{tabular}
\caption{Ranking agreement rate between the Cohere Reward Model, which is used in our experiments to label preference data and rate online generations, and GPT-4 Turbo on randomly selected multilingual Dolly responses generated from Command and Command R+ (512 randomly selected per language). Unlike the reward models, GPT-4-Turbo is capable of outputting ties. GPT-4-Turbo selects tie result 3.1\% of the time on this dataset.} 
\label{tab:rm-stats}
\end{wraptable}

\textbf{\textbf{Reward Model}} \label{sec:reward-model}We use the Cohere reward model (\texttt{Cohere May 2024}) which is competitive with top-scoring state-of-the-art RMs on the RewardBench Leaderboard \citep{lambert2024rewardbench}, scoring 88.2, which currently is ranked 4th.\footnote{\url{https://huggingface.co/spaces/allenai/reward-bench}} This reward model achieves high LLM response ranking agreement with GPT-4-Turbo on English and non-English languages as shown in Table \ref{tab:rm-stats}. We intentionally use a separate reward model from the model we use for llm-as-a-judge evaluation (GPT-4-Turbo\footnote{\url{https://platform.openai.com/docs/models/gpt-4-turbo-and-gpt-4}}) given the known biases incurred by using the same model for both \citep{verga2024replacing,bansal2024peering}. 

\textbf{\textbf{Preference Optimization Training}} We train Aya 23 8B model for 2 epochs in both DPO and RLOO experiments. DPO runs are trained with KL-penalty $\beta = 0.5$, learning rate 5e-7, and AdamW optimizer \citep{kingma2014adam}. RLOO runs are trained with RLOO $k=2$,  KL-penalty $\beta = 0.01$, 
learning rate \num{5e-6}, AdamW optimizer, and online generation sampling temperature of 0.75. All runs are performed on a single node with either 8 x Nvidia A100 80GB or 8 x Nvidia H100 80GB GPUs with DeepSpeed ZeRO Stage 3 \citep{rajbhandari2020zero} and full fine-tuning of all 8 billion model parameters. We performed hyperparameter sweeps for learning rate $lr \in \{\num{5e-8}, \num{5e-7}, \num{5e-6}, \num{5e-5}\}$ and for KL-penalty $\beta \in \{0.05,0.1,0.5\}$ for both DPO and RLOO to the best of our ability. For a fair comparison with DPO, we utilize the same reward model for RLOO which is used to generate our synthetic multilingual preference dataset (for ranking the generations).   

\begin{figure*}[t]
  \begin{subfigure}[b]{0.32\textwidth}
    \includegraphics[width=\textwidth]{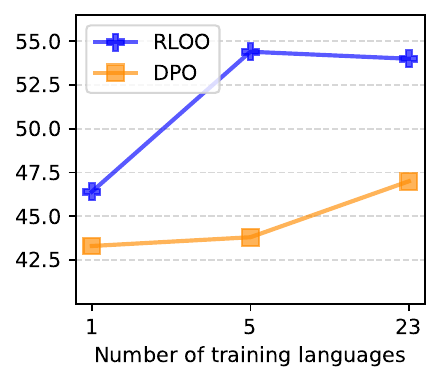}
    \caption{Avg. win\% on \textbf{23} languages}
    \label{fig:ml-23}
  \end{subfigure} 
  \begin{subfigure}[b]{0.32\textwidth}
    \includegraphics[width=\textwidth]{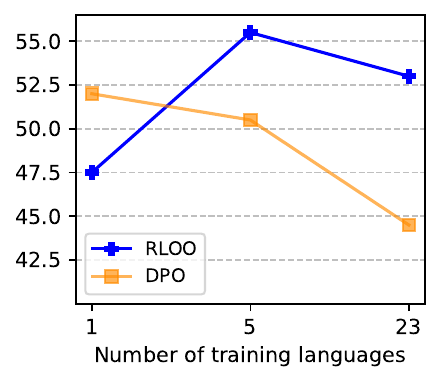}
    \caption{Win\% on \textbf{English}}
    \label{fig:en-only}
  \end{subfigure} 
  \begin{subfigure}[b]{0.32\textwidth}
    \includegraphics[width=\textwidth]{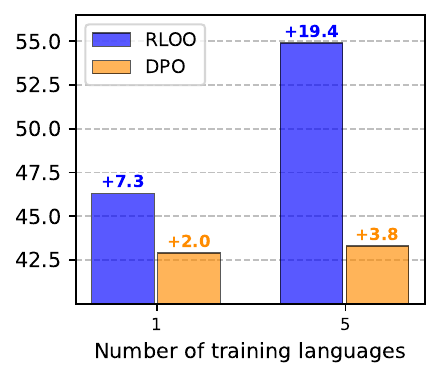}
    \caption{Avg. win\% on \textbf{unseen} langs.}
    \label{fig:cross-lingual}
  \end{subfigure} 
  \caption{DPO and RLOO win-rates as  number of languages in training data increases. We report win-rates for \textbf{(a)} the average of 23 languages, \textbf{(b)} only English, and \textbf{(c)} the average of unseen languages, reflecting the cross-lingual transfer.}
  \label{fig:results}
\end{figure*}

\subsection{Model Comparisons}\label{sec:baselines}
We evaluate against multiple state-of-the-art open-source 
models to ensure a comprehensive evaluation. Details of each model are below:
\begin{itemize}
\item \textbf{Meta-Llama-3-8B-Instruct}
~\citep{llama3modelcard} is an 8B parameter open-source instruction fine-tuned model which has been pre-trained on over 15T tokens. Over 5\% of the pretraining data is high-quality non-English data, covering over 30 languages. The model is supervised fine-tuned and preference optimized with both  offline (DPO) and online (PPO) algorithms

\item \textbf{Mistral-7B-Instruct-v0.3}
~\citep{jiang2023mistral} is an open-source instruct fine-tuned edition of the Mistral-7B pre-trained model. The model is trained on instruction datasets publicly available on the HuggingFace repository.

\item \textbf{Gemma-1.1-7B-it}
~\citep{gemmareport} is a 7B parameter instruction fine-tuned model trained with Gemini models' architectures, data, and training recipes~\citep{geminiteam2024gemini} on 6T tokens of data from web documents, mathematics, and code that are primarily English. In addition to the supervised fine-tuning, this model is also further fine-tuned using RLHF on collected pairs of preferences from human annotators. 

\end{itemize}

We note that while the models we evaluate do not explicitly claim to support multiple languages, in practice, they are heavily used by multilingual users relative to explicitly multilingual models like mT0 and BLOOMZ ~\citep{muennighoff2023crosslingual}. We also observe that they perform well in practice.

\subsection{Evaluation} 
We assess the preference-optimized models on the multilingual open-ended generation and summarization tasks using LLM-simulated evaluation:
\begin{enumerate}
    \item \textbf{Open-ended generations} For open-ended generations, we use dolly-machine-translated test set from the \textbf{Aya evaluation suite}~\citep{ayadata2024} which is a held-out test set of 200 instances from the Dolly-15k dataset \citep{DatabricksBlog2023DollyV2} translated into 101 languages. This test set was curated by avoiding instructions that include culturally-specific or geographic references. For languages that are available (Arabic, French, Hindi, Russian, and Spanish), we use the \textbf{dolly-human-edited} test set~\citep{ayadata2024}, an improved version of \textbf{dolly-machine-translated} post-edited by professional human annotators to correct any translation issues. 
\item \textbf{Summarization Task} For summarization, we use \textbf{XLSum} \citep{hasan-etal-2021-xl}, for the subset of 15 languages covered by the benchmark within the Aya 23 language coverage. 
\end{enumerate}

Across both tasks, we measure LLM-simulated win-rates which have been shown to be highly correlated with human evaluation for both monolingual English settings \citep{dubois2024alpacafarm} and multilingual settings \citep{ustun2024aya}. We use GPT-4-Turbo as an LLM-judge and follow the same procedure described by \citet{ustun2024aya}. To minimize bias, we randomize the order of model outputs. The judge prompt can be found in Appendix \ref{appendix:judge-prompt}. For evaluation, we use max prompt context length of 512, maximum generation length of 512, and sampling temperature of 0.75.

\section{Results and Discussion}\label{sec:results}

\textbf{Win-rates Against Open-Weights Models} 
Figure \ref{fig:win-rate-comp} and Table \ref{tab:ml-comp} show the win-rates of our preference-trained models against state-of-the-art open-source models. Importantly, the base Aya 23 8B already achieves high win-rates against Gemma-1.1 (62.1\%), Llama-3 (66.6\%), and Mistral-v0.3 (69\%) averaged across all 23 languages on Dolly open-ended generations. Preference-optimized Aya models extend this lead further. Concretely, our best variant of RLOO leads to 69.5\% (+7.4), 72.4\% (+5.8), and 77.5 (+8.5) win-rates against Gemma, Llama-3, and Mistral respectively.
\setlength{\columnsep}{15pt}
\begin{wraptable}{R}{0.5\linewidth}
\centering
\vspace{-.5cm}
\fontsize{9.5}{9.5}\selectfont
\setlength{\tabcolsep}{4pt}
\begin{tabular}{ll|rrr|}
\toprule
 &  & \multicolumn{3}{c}{\textbf{Average 23 Languages}}    \\
 & &
\multicolumn{1}{c}{\textbf{Win\%}} & \multicolumn{1}{c}{\textbf{Loss\%}} & \multicolumn{1}{c}{$\Delta$\textbf{W-L\%}} \\ 

\midrule

& \textsc{Gemma-1.1}    & 
\multicolumn{1}{r}{62.1} & \multicolumn{1}{r}{29.4} & \multicolumn{1}{c}{\cellcolor[HTML]{FFFFFF}32.7} \\

& \textsc{Llama-3}   & 
\multicolumn{1}{r}{66.6} & \multicolumn{1}{r}{29.4} & \multicolumn{1}{c}{\cellcolor[HTML]{D1EDDF}37.2} \\ 

\multirow{-3}{*}{\textsc{Base}} & \textsc{Mistral-v0.3} & 
\multicolumn{1}{r}{69.0} & \multicolumn{1}{r}{26.8} & \multicolumn{1}{c}{\cellcolor[HTML]{B5E1CB}42.2} \\  

\midrule

& \textsc{Gemma-1.1}    & 
\multicolumn{1}{r}{67.7} & \multicolumn{1}{r}{27.1} & \multicolumn{1}{c}{\cellcolor[HTML]{BEE5D2}40.6} \\

& \textsc{Llama-3}   & 
\multicolumn{1}{r}{71.0} & \multicolumn{1}{r}{24.7} & \multicolumn{1}{c}{\cellcolor[HTML]{9DD8BB}46.3} \\ 

\multirow{-3}{*}{\textsc{DPO}} & \textsc{Mistral-v0.3} & 
\multicolumn{1}{r}{74.7} & \multicolumn{1}{r}{21.8} & \multicolumn{1}{c}{\cellcolor[HTML]{78C9A1}52.9} \\  

\midrule

& \textsc{Gemma-1.1}    & 
\multicolumn{1}{r}{69.5} & \multicolumn{1}{r}{26.3} & \multicolumn{1}{c}{\cellcolor[HTML]{AFDFC7}43.2} \\

& \textsc{Llama-3}  & 
\multicolumn{1}{r}{72.4} & \multicolumn{1}{r}{24.0} & \multicolumn{1}{c}{\cellcolor[HTML]{91D3B3}48.4} \\ 

\multirow{-3}{*}{\textsc{RLOO}} & \textsc{Mistral-v0.3}  & 
\multicolumn{1}{r}{77.5} & \multicolumn{1}{r}{18.9} & \multicolumn{1}{c}{\cellcolor[HTML]{57BB8A}58.6} \\

\bottomrule
\end{tabular}
\caption{Open-ended generation (Dolly) win-rates for the base Aya 23 8B, and DPO/RLOO preference optimized Aya 23 8B models against Gemma-1.1-7B-it, Meta-llama-3-8B-Instruct and Mistral-7B-Instruct-v0.3. We report average win-rates on 23 languages for the best ML-23-230K checkpoints for DPO and RLOO.} 
\label{tab:ml-comp}


\end{wraptable}

\label{sec:against-aya}
\textbf{Win Rates Against Aya-23-8B} Table \ref{tab:main_res} shows win-rate results for open-ended generations. Win-rates are measured against Aya 23 8B. We report win-rates for both DPO and RLOO trained with different preference data mixtures. XLSum summarization results are shown in Appendix (Table \ref{tab:xlsum-overall}). Our best variant across experiments achieves a win-rate of 70.7\% over the base Aya model.



\textbf{Increasing multilinguality in preference data improves winrates} 
For a fixed dataset size of 50K pairwise preferences, we find that increasing the number of languages in training data improves overall performance as shown in Figure \ref{fig:ml-23} and Table \ref{tab:main_res} (right). For DPO, win-rates against the base model on 23 languages increases from 43.3\% to 47.0\%.\footnote{Win-rates that are under 50\% do not correspond lower performance since our evaluation allows for \emph{Tie}. To indicate if there is a performance gain, we also report the difference between win- and loss-rates (\textbf{$\Delta$W-L}) in the results.} For RLOO, this gain is most visible when the number of training languages is 5 where win-rate improves from 46.4\% to 54.0\% (+7.6 increase).
Interestingly, in open-ended generations, using all 23 languages does not improve performance further for RLOO, however, in summarization, the 23-language training mixture increases win-rates compared to the 5-language subset (from 65.1\% to 70.7\%, Table \ref{tab:xlsum-overall}).

\begin{table}[]
    \centering

\begin{tabular}{cc}

\fontsize{9.5}{9.5}\selectfont
\begin{tabular}{ll|rrr|}

\toprule
 &  & \multicolumn{3}{c}{\textbf{English}}    \\
 & &
\multicolumn{1}{c}{\textbf{Win\%}} & \multicolumn{1}{c}{\textbf{Loss\%}} & \multicolumn{1}{c}{$\Delta$\textbf{W-L\%}} \\ 

\midrule

& \textsc{EN-1}    & 
\multicolumn{1}{r}{52.0} & \multicolumn{1}{r}{33.5} & \multicolumn{1}{c}{\cellcolor[HTML]{A0D9BD}18.5} \\

& \textsc{ML-5}  & 
\multicolumn{1}{r}{50.5} & \multicolumn{1}{r}{28.5} & \multicolumn{1}{c}{\cellcolor[HTML]{80CCA7}22.0} \\ 

& \textsc{ML-23}   & 
\multicolumn{1}{r}{44.5} & \multicolumn{1}{r}{36.5} & \multicolumn{1}{c}{\cellcolor[HTML]{FFFFFF}8.0} \\ 

\multirow{-4}{*}{\textsc{DPO}} & \textsc{ML-23*} & 
\multicolumn{1}{r}{57.5} & \multicolumn{1}{r}{31.0} & \multicolumn{1}{c}{\cellcolor[HTML]{57BB8A}26.5} \\  

\midrule

& \textsc{EN-1}    & 
\multicolumn{1}{r}{47.5} & \multicolumn{1}{r}{38.5} & \multicolumn{1}{c}{\cellcolor[HTML]{F6FCF9}9.0} \\

& \textsc{ML-5}  & 
\multicolumn{1}{r}{55.5} & \multicolumn{1}{r}{30.5} & \multicolumn{1}{c}{\cellcolor[HTML]{65C194}25.0} \\ 

& \textsc{ML-23}  & 
\multicolumn{1}{r}{53.0} & \multicolumn{1}{r}{37.0} & \multicolumn{1}{c}{\cellcolor[HTML]{B7E2CD}16.0} \\ 

\multirow{-4}{*}{\textsc{RLOO}} & \textsc{ML-23*}  & 
\multicolumn{1}{r}{53.0} & \multicolumn{1}{r}{35.0} & \multicolumn{1}{c}{\cellcolor[HTML]{A5DBC0}18.0} \\

\bottomrule
\end{tabular}

&

\fontsize{9.5}{9.5}\selectfont
\begin{tabular}{ll|rrr|}
\toprule
 &  & \multicolumn{3}{c}{\textbf{Average 23 Languages}}    \\
 & &
\multicolumn{1}{c}{\textbf{Win\%}} & \multicolumn{1}{c}{\textbf{Loss\%}} & \multicolumn{1}{c}{$\Delta$\textbf{W-L\%}} \\ 

\midrule

& \textsc{EN-1}    & 
\multicolumn{1}{r}{43.3} & \multicolumn{1}{r}{40.6} & \multicolumn{1}{c}{\cellcolor[HTML]{FFFFFF}2.7} \\

& \textsc{ML-5}  & 
\multicolumn{1}{r}{43.8} & \multicolumn{1}{r}{39.1} & \multicolumn{1}{c}{\cellcolor[HTML]{EAF7F1}4.7} \\ 

& \textsc{ML-23}   & 
\multicolumn{1}{r}{47.0} & \multicolumn{1}{r}{37.1} & \multicolumn{1}{c}{\cellcolor[HTML]{B3E1CB}9.9} \\ 

\multirow{-4}{*}{\textsc{DPO}} & \textsc{ML-23*} & 
\multicolumn{1}{r}{50.2} & \multicolumn{1}{r}{39.0} & \multicolumn{1}{c}{\cellcolor[HTML]{A6DBC1}11.2} \\  

\midrule

& \textsc{EN-1}    & 
\multicolumn{1}{r}{46.4} & \multicolumn{1}{r}{38.9} & \multicolumn{1}{c}{\cellcolor[HTML]{CDEBDC}7.5} \\

& \textsc{ML-5}  & 
\multicolumn{1}{r}{54.4} & \multicolumn{1}{r}{35.8} & \multicolumn{1}{c}{\cellcolor[HTML]{57BB8A}18.6} \\ 

& \textsc{ML-23}  & 
\multicolumn{1}{r}{54.0} & \multicolumn{1}{r}{38.0} & \multicolumn{1}{c}{\cellcolor[HTML]{73C79E}16.0} \\ 

\multirow{-4}{*}{\textsc{RLOO}} & \textsc{ML-23*}  & 
\multicolumn{1}{r}{53.4} & \multicolumn{1}{r}{37.0} & \multicolumn{1}{c}{\cellcolor[HTML]{6FC59B}16.4} \\

\bottomrule

\end{tabular}

\end{tabular}

    \caption{Open-ended generation (Dolly) win-rates for DPO/RLOO preference optimized Aya models against the original Aya 23 8B on \textbf{English (left)} and \textbf{averaged over 23 languages (right)}. We report average win-rates on 23 languages for multiple training data mixtures: EN-1 (English Only), ML-5 (5 Languages), and ML-23 (23 Languages). All the data mixtures consist of 50K total training examples with the exception of ML-23*, which includes 230K total training examples. We report results for the best checkpoint across 2 epochs.}
    \label{tab:main_res}
\end{table}




\textbf{English also benefits from multilingual training} Our experiments also show that English can benefit from multilingual preference training and positive transfer, even when there are fewer total English examples in the training data. As shown in Figure \ref{fig:en-only} (and Table \ref{tab:main_res}), despite only having approximately 2K English examples, RLOO improves English win-rates from 47.5\% to 53.0\% with 23-languages training data. However, English win-rates drop for DPO as the number of languages increases when there is a fixed budget of 50K examples, showing that DPO is more prone to negative interference.

\setlength{\columnsep}{15pt}
\begin{wraptable}{R}{0.5\linewidth}
\centering
\vspace{-.5cm}
\fontsize{9.5}{9.5}\selectfont
\setlength{\tabcolsep}{4pt}

\begin{tabular}{ll|rrr}
\toprule
 &  & \multicolumn{3}{c}{\textbf{Avg. Unseen Langs.}}  \\
 &  & \multicolumn{1}{c}{\textbf{Win \%}} & \multicolumn{1}{c}{\textbf{Loss \%}} & \multicolumn{1}{c}{\textbf{$\Delta$W-L\%}} \\ 
\midrule
& \textsc{DPO} & \multicolumn{1}{r}{42.9} & \multicolumn{1}{r}{40.9} & \multicolumn{1}{c}{\cellcolor[HTML]{FFFFFF}2.0}  \\
\multirow{-2}{*}{\textsc{EN-1}}& \textsc{RLOO} & \multicolumn{1}{r}{46.3} & \multicolumn{1}{r}{39.3} & \multicolumn{1}{c}{\cellcolor[HTML]{CCEBDC}7.3}  \\
 
\midrule
& \textsc{DPO} & \multicolumn{1}{r}{43.3} & \multicolumn{1}{r}{39.5} & \multicolumn{1}{c}{\cellcolor[HTML]{EEF8F3}3.8}   \\
\multirow{-2}{*}{\textsc{ML-5}}& \textsc{RLOO} & \multicolumn{1}{r}{54.9} & \multicolumn{1}{r}{35.5} & \multicolumn{1}{c}{\cellcolor[HTML]{57BB8A}19.4}   \\ 
\bottomrule
\end{tabular}%
\caption{Win-rates for the 22 and 18 languages that are not included in the training data (``unseen'') for EN-1 and ML-5 respectively. We observe cross-lingual transfer from preference optimization, with an increased degree of transfer enhanced by multilingual training.
} 
\label{tab:cross-lingual}
\end{wraptable}

\textbf{Cross-lingual transfer to unseen languages} Preference training only with English, achieves performance gains for languages not seen in the training data as shown in Figure \ref{fig:cross-lingual} (and Table \ref{tab:cross-lingual}). This gain (\textbf{$\Delta$W-L}) is 2.0\% for DPO and 7.3\% for RLOO. Furthermore, using a 5-language subset (ML-5) significantly increases these gains to 3.8\% and 19.4\% for DPO and RLOO respectively. These results provide strong evidence of cross-lingual transfer in preference optimization, which is significantly more present after online training, with an increased degree of transfer facilitated by multilingual training data compared with English only.

\textbf{Online optimization vs offline} We find that RLOO (online) outperforms DPO (offline) across the board in multilingual performance. As shown in in Table \ref{tab:main_res}, RLOO achieves higher win-rates with all the preference data mixtures compared to DPO. For the 23-language mixture, RLOO achieves 54.0\% win-rates whereas DPO reaches 47.0\%. 

Furthermore, RLOO achieves higher cross-lingual transfer with English-only and 5-language preference training in comparison to DPO. On languages unseen during training, RLOO achieves 3.4\% higher win-rate compared to DPO (46.3\% vs 42.9\%) when training only on English and an 11.6\% higher win-rate (54.9\% vs 43.3) when training on the 5-language subset as shown in Table \ref{tab:cross-lingual}. This is inline with recent works \citep{tajwar2024preference,tang2024understanding}, which suggest that online preference optimization outperforms fully offline methods. Additionally, the significant improvement in cross-lingual transfer with RLOO compared to DPO suggests that online sampling also enables \textit{generalization} beyond the distribution of the training prompts. This is complementary to findings of \citep{kirk2024understandingRLHF} which show that online RLHF enables better generalization than SFT. 

\textbf{Role of data size and reward over-optimization} 
For DPO, increasing the amount of multilingual data from approximately 2K to 10K per language in the 23-language mixture improves win-rates from 47.0\% to 50.2\% (Table \ref{tab:main_res}). For RLOO, however, the same increase does not lead to an improvement. For all runs, the best checkpoint was the last one (epoch 2), except for the RLOO ML-23 230K run where we observed significant performance degradation after 0.5 epochs, which may be caused by reward model overoptimization \citep{gao2022scaling}. Prior works have shown that low-resource languages can jailbreak LLMs \citep{yong2024lowresource} and it is likely that reward models, which typically are initialized from LLMs, likely share similar vulnerabilities. The degradation for the RLOO ML-23-230K run we observe may be caused by online optimization exploiting more languages and prompts where the reward model may be more prone to reward hacking, as this run includes all 23 languages and more prompts per language than other runs.



\textbf{Is there a multilingual alignment tax?} Post-training stages of LLMs including supervised fine-tuning and preference optimization have increasingly been torn between objectives: improving traditional discriminative benchmarks like 
HellaSwag~\citep{zellers-etal-2019-hellaswag}, MMLU~\citep{hendryckstest2021}  and training LLMs to follow instructions, acquire conversational abilities, and be helpful and harmless~\citep{askell2021general}. Recent work on multilingual instruction finetuning \citep{ustun2024aya} has found that improvements in open-ended generative tasks introduce trade-offs with discriminative tasks. However, this work only studies the tensions introduced by supervised instruction finetuning. 
More recent work \citep{tang2024understanding} on preference training suggests that offline methods impart improved ability for discriminative tasks, whereas on-policy sampling improves generative quality. Here, we explore whether this holds for multilingual and characterize the trade-off between discriminate and generative performance.

We follow \citet{aryabumi2024aya} and evaluate models on the unseen tasks --XWinograd \citep{muennighoff2022crosslingual}, XCOPA \citep{ponti2020xcopa}, XStoryCloze \citep{lin-etal-2022-shot}--, mMMLU \citep{lai-etal-2023-okapi}, and MGSM \citep{shi2023language-mgsm} using the \texttt{eval-harness} framework~\citep{gao2021framework}.


We find that both DPO and RLOO are robust in terms of their results in multilingual benchmarks, matching the performance of the base Aya 23 8B model. More specifically, multilingual preference optimization through DPO leads to a slight gain of 0.3 in mMMLU (48.5 vs 48.2) while RLOO leads to a slight drop of 0.2 (48.0 vs 48.2) as show in Table \ref{tab:m_mmlu}. Performance on MGSM slightly drops by an accuracy of 0.5 (36.6 vs 36.1) and 1.5 (36.6 vs 35.1) for DPO and RLOO respectively as shown in Table \ref{tab:mgsm}. All discriminative benchmark results can be found in Appendix \ref{appendix:disc-results}. 
In contrast to recent work \citep{tang2024understanding} in a monolingual setting, this shows that multilingual preference optimization can substantially improve generative performance as discussed in Section \ref{sec:against-aya}, while incurring a minimal alignment tax, given the right regularization, on common multilingual NLP tasks. 



\section{Related Work}\label{sec:related_work}

\textbf{Reinforcement Learning from Human Feedback (RLHF)}  
RLHF has become  the dominant paradigm for aligning LLMs to human preferences. Canonically, this involves training a reward model and using a reinforcement learning algorithm like PPO \citep{schulman2017trust} or RLOO \citep{ahmadian2024basics} to  optimize the LLM policy to maximize reward of online samples generated throughout training. \citep{ouyang2022instructgpt, LearningToSummarizeHF, christiano2023RLHF}. There has been a plethora of work attempting to take the online inference aspect, and the optimization difficulties and complexities of RL that come with it, out of RLHF. The most prominent of these, is the family of methods base upon Direct Preference Optimization (DPO) \citep{rafailov2023DPO}, such as IPO \citep{azar2024ipo}, KTO \citep{ethayarajh2024kto}, and SRPO \citep{choi2024srpo}. This family of methods directly fine-tunes an LLM to be impicitly consistent with collected preference data, forgoing the need for training a separate reward  model.

\textbf{Synthetic Data} Collecting ground truth completions or feedback data from humans  is often very expensive. Thus, many recent works have explored the use of training LLMs using data generated by LLMs. Distallation is a common technique which uses a stronger (typically larger model) to generate ground truth completions to SFT a weaker (typically smaller model) \citep{taori2023alpaca}. As many recent LLMs (and Reward Models intialized from LLMs) have been shown to be strong evaluators of LLM completions \citep{zheng2023judging, dubois2024alpacafarm}, many works use LLMs or Reward Models to generate synthetic preference data rather than collecting preferences from humans, which is later used in RLHF training\citep{bai2022constitutional, pace2024westofn}.LLMs can also be used to provide additional rankings, ratings, or natural language feedback, which can be used in subsequent RLHF training. Methods which use AI generated feedback are part of a family of Reinforcement Learning from AI Feedback (RLAIF) methods. 

\textbf{Preference Optimization for Multilingual LLMs} 
There have been limited efforts on multilingual preference optimization to-date. \citep{lai-etal-2023-okapi} present a multilingual instruction tuning framework, where they preference train multilingual LLMs such as BLOOMZ \citep{muennighoff2023crosslingual} for 26 non-English languages with RLHF. They synthetically generated a preference dataset by translating an extended version of the Alpaca dataset \citep{taori2023alpaca}, generating responses from their target LLM and ranking back-translated (into English) responses with ChatGPT.\footnote{\url{https://openai.com/blog/chatgpt/}} In contrast to our work, they perform preference optimization for each language separately. However, due to their potentially low-quality dataset which heavily relies on translations, their resulting language-specific models are outperformed by other massively multilingual LLMs without preference optimization \citep{ustun2024aya,aryabumi2024aya}. \citet{wu2024reuse} study cross-lingual transfer in reward model (RM) training where they propose using preference data in one source language to train an RM for target language alignment. They show that RMs based on amultilingual base model exhibit zero-shot cross-lingual transfer consistently across different languages. However, they do not experiment with using multiple source languages in training, which we show is crucial in the preference optimization both for offline optimization such as DPO and online RL methods such as RLOO. A number of existing works also explore methods for preference optimization in a highly related setting where LLMs are aligned to preferences of diverse groups of people around the world \citep{zhao2024group, jiang2023personallm, hwang2023aligning, deshpande2023toxicity}.

\section{Conclusion}

Our work presents a comprehensive study on multilingual preference optimization. We show that the inclusion of multilingual data in preference optimization leads to significant improvements in multilingual performance over English-only preference optimization. This improvement scales both with the number of languages and the total number of examples included in the training data. Additionally, we show that preference training exhibits cross-lingual transfer, leading to significant gains in languages not present in the training data. We also find that using online preference optimization outperforms offline preference optimization, highlighting the importance of online samples during training.

As a result of our study, we expand the frontier of alignment techniques to 23 languages which cover half the world's population, by successfully preference-training an 8-billion parameter Aya 23 model that outperforms both the original Aya 23 8B base and widely used models including Gemma, Mistral, and Llama 3.

\section{Limitations}

A potential risk of relying on synthetic and translated datasets is the presence of particular cultural biases in model behavior. The prompts used in ShareGPT to seed the creation of the synthetic data over-index on contributions from the Global North or Western regions \citep{longpre2023data}. This could introduce a skew towards a narrow selection of cultural viewpoints.

Our preference-trained model covers 23 languages and improves performance relative to the closest open-source model. However, this is still only a tiny fraction of the world's linguistic diversity which encompasses 7000 languages. Furthermore, in this research we don't distinguish between dialects within the languages we cover, which are an important part of how language is used in practice \citep{zampieri2020natural, wolfram1997issues, brown2020languageGPT3, lent-etal-2022-creole, blaschke-etal-2023-survey,falck2012dialects}. Future work, should aim to include more of the world's population and therefore languages.

Due to compute constraints, we are limited in our ability to run preference optimization experiments for larger models. Many of the runs we describe in this work for a single run can take 5 days to complete on a single 8 x H100 80GB GPU instance. Future work should explore the impact of scaling model size and further tune other hyperparameters for multilingunal preference optimization.

\section{Acknowledgements}

We thank Sander Land, Chris Cremer, and Florian Strub for their helpful discussions and support with infrastructure for our experiments.



\bibliography{main}

\begin{thebibliography}{106}
\providecommand{\natexlab}[1]{#1}
\providecommand{\url}[1]{\texttt{#1}}
\expandafter\ifx\csname urlstyle\endcsname\relax
  \providecommand{\doi}[1]{doi: #1}\else
  \providecommand{\doi}{doi: \begingroup \urlstyle{rm}\Url}\fi

\bibitem[Ahmadian et~al.(2024)Ahmadian, Cremer, Gallé, Fadaee, Kreutzer, Pietquin, Üstün, and Hooker]{ahmadian2024basics}
Arash Ahmadian, Chris Cremer, Matthias Gallé, Marzieh Fadaee, Julia Kreutzer, Olivier Pietquin, Ahmet Üstün, and Sara Hooker.
\newblock Back to basics: Revisiting reinforce style optimization for learning from human feedback in llms, 2024.

\bibitem[AI@Meta(2024)]{llama3modelcard}
AI@Meta.
\newblock Llama 3 model card.
\newblock 2024.
\newblock URL \url{https://github.com/meta-llama/llama3/blob/main/MODEL_CARD.md}.

\bibitem[Anthropic(2024)]{anthropic2024claude}
AI~Anthropic.
\newblock The claude 3 model family: Opus, sonnet, haiku.
\newblock \emph{Claude-3 Model Card}, 2024.

\bibitem[Aryabumi et~al.(2024)Aryabumi, Dang, Talupuru, Dash, Cairuz, Lin, Venkitesh, Smith, Marchisio, Ruder, et~al.]{aryabumi2024aya}
Viraat Aryabumi, John Dang, Dwarak Talupuru, Saurabh Dash, David Cairuz, Hangyu Lin, Bharat Venkitesh, Madeline Smith, Kelly Marchisio, Sebastian Ruder, et~al.
\newblock Aya 23: Open weight releases to further multilingual progress.
\newblock \emph{arXiv preprint arXiv:2405.15032}, 2024.

\bibitem[Askell et~al.(2021)Askell, Bai, Chen, Drain, Ganguli, Henighan, Jones, Joseph, Mann, DasSarma, Elhage, Hatfield-Dodds, Hernandez, Kernion, Ndousse, Olsson, Amodei, Brown, Clark, McCandlish, Olah, and Kaplan]{askell2021general}
Amanda Askell, Yuntao Bai, Anna Chen, Dawn Drain, Deep Ganguli, Tom Henighan, Andy Jones, Nicholas Joseph, Ben Mann, Nova DasSarma, Nelson Elhage, Zac Hatfield-Dodds, Danny Hernandez, Jackson Kernion, Kamal Ndousse, Catherine Olsson, Dario Amodei, Tom Brown, Jack Clark, Sam McCandlish, Chris Olah, and Jared Kaplan.
\newblock A general language assistant as a laboratory for alignment, 2021.

\bibitem[Azar et~al.(2024)Azar, Guo, Piot, Munos, Rowland, Valko, and Calandriello]{azar2024ipo}
Mohammad~Gheshlaghi Azar, Zhaohan~Daniel Guo, Bilal Piot, Remi Munos, Mark Rowland, Michal Valko, and Daniele Calandriello.
\newblock A general theoretical paradigm to understand learning from human preferences.
\newblock In \emph{International Conference on Artificial Intelligence and Statistics}, pp.\  4447--4455. PMLR, 2024.

\bibitem[Bai et~al.(2022)Bai, Kadavath, Kundu, Askell, Kernion, Jones, Chen, Goldie, Mirhoseini, McKinnon, Chen, Olsson, Olah, Hernandez, Drain, Ganguli, Li, Tran-Johnson, Perez, Kerr, Mueller, Ladish, Landau, Ndousse, Lukosuite, Lovitt, Sellitto, Elhage, Schiefer, Mercado, DasSarma, Lasenby, Larson, Ringer, Johnston, Kravec, Showk, Fort, Lanham, Telleen-Lawton, Conerly, Henighan, Hume, Bowman, Hatfield-Dodds, Mann, Amodei, Joseph, McCandlish, Brown, and Kaplan]{bai2022constitutional}
Yuntao Bai, Saurav Kadavath, Sandipan Kundu, Amanda Askell, Jackson Kernion, Andy Jones, Anna Chen, Anna Goldie, Azalia Mirhoseini, Cameron McKinnon, Carol Chen, Catherine Olsson, Christopher Olah, Danny Hernandez, Dawn Drain, Deep Ganguli, Dustin Li, Eli Tran-Johnson, Ethan Perez, Jamie Kerr, Jared Mueller, Jeffrey Ladish, Joshua Landau, Kamal Ndousse, Kamile Lukosuite, Liane Lovitt, Michael Sellitto, Nelson Elhage, Nicholas Schiefer, Noemi Mercado, Nova DasSarma, Robert Lasenby, Robin Larson, Sam Ringer, Scott Johnston, Shauna Kravec, Sheer~El Showk, Stanislav Fort, Tamera Lanham, Timothy Telleen-Lawton, Tom Conerly, Tom Henighan, Tristan Hume, Samuel~R. Bowman, Zac Hatfield-Dodds, Ben Mann, Dario Amodei, Nicholas Joseph, Sam McCandlish, Tom Brown, and Jared Kaplan.
\newblock Constitutional ai: Harmlessness from ai feedback, 2022.

\bibitem[Bansal et~al.(2024)Bansal, Dang, and Grover]{bansal2024peering}
Hritik Bansal, John Dang, and Aditya Grover.
\newblock Peering through preferences: Unraveling feedback acquisition for aligning large language models.
\newblock In \emph{The Twelfth International Conference on Learning Representations}, 2024.
\newblock URL \url{https://openreview.net/forum?id=dKl6lMwbCy}.

\bibitem[Bird(2022)]{bird-2022-local}
Steven Bird.
\newblock Local languages, third spaces, and other high-resource scenarios.
\newblock pp.\  7817--7829, 01 2022.
\newblock \doi{10.18653/v1/2022.acl-long.539}.

\bibitem[Bizzoni et~al.(2020)Bizzoni, Juzek, Espa{\~n}a-Bonet, Dutta~Chowdhury, van Genabith, and Teich]{bizzoni-etal-2020-human}
Yuri Bizzoni, Tom~S Juzek, Cristina Espa{\~n}a-Bonet, Koel Dutta~Chowdhury, Josef van Genabith, and Elke Teich.
\newblock How human is machine translationese? comparing human and machine translations of text and speech.
\newblock In Marcello Federico, Alex Waibel, Kevin Knight, Satoshi Nakamura, Hermann Ney, Jan Niehues, Sebastian St{\"u}ker, Dekai Wu, Joseph Mariani, and Francois Yvon (eds.), \emph{Proceedings of the 17th International Conference on Spoken Language Translation}, pp.\  280--290, Online, July 2020. Association for Computational Linguistics.
\newblock \doi{10.18653/v1/2020.iwslt-1.34}.
\newblock URL \url{https://aclanthology.org/2020.iwslt-1.34}.

\bibitem[Blaschke et~al.(2023)Blaschke, Schuetze, and Plank]{blaschke-etal-2023-survey}
Verena Blaschke, Hinrich Schuetze, and Barbara Plank.
\newblock A survey of corpora for {G}ermanic low-resource languages and dialects.
\newblock In \emph{Proceedings of the 24th Nordic Conference on Computational Linguistics (NoDaLiDa)}, pp.\  392--414, T{\'o}rshavn, Faroe Islands, May 2023. University of Tartu Library.
\newblock URL \url{https://aclanthology.org/2023.nodalida-1.41}.

\bibitem[Boubdir et~al.(2023)Boubdir, Kim, Ermis, Fadaee, and Hooker]{boubdir2023prompts}
Meriem Boubdir, Edward Kim, Beyza Ermis, Marzieh Fadaee, and Sara Hooker.
\newblock Which prompts make the difference? data prioritization for efficient human llm evaluation, 2023.

\bibitem[Bradley \& Terry(1952)Bradley and Terry]{Bradley1952RankAO}
Ralph~Allan Bradley and Milton~E. Terry.
\newblock Rank analysis of incomplete block designs: I. the method of paired comparisons.
\newblock \emph{Biometrika}, 39:\penalty0 324, 1952.
\newblock URL \url{https://api.semanticscholar.org/CorpusID:125209808}.

\bibitem[Brown et~al.(2020)Brown, Mann, Ryder, Subbiah, Kaplan, Dhariwal, Neelakantan, Shyam, Sastry, Askell, Agarwal, Herbert-Voss, Krueger, Henighan, Child, Ramesh, Ziegler, Wu, Winter, Hesse, Chen, Sigler, Litwin, Gray, Chess, Clark, Berner, McCandlish, Radford, Sutskever, and Amodei]{brown2020languageGPT3}
Tom~B. Brown, Benjamin Mann, Nick Ryder, Melanie Subbiah, Jared Kaplan, Prafulla Dhariwal, Arvind Neelakantan, Pranav Shyam, Girish Sastry, Amanda Askell, Sandhini Agarwal, Ariel Herbert-Voss, Gretchen Krueger, Tom Henighan, Rewon Child, Aditya Ramesh, Daniel~M. Ziegler, Jeffrey Wu, Clemens Winter, Christopher Hesse, Mark Chen, Eric Sigler, Mateusz Litwin, Scott Gray, Benjamin Chess, Jack Clark, Christopher Berner, Sam McCandlish, Alec Radford, Ilya Sutskever, and Dario Amodei.
\newblock Language models are few-shot learners, 2020.

\bibitem[Casper et~al.(2023)Casper, Davies, Shi, Gilbert, Scheurer, Rando, Freedman, Korbak, Lindner, Freire, Wang, Marks, Segerie, Carroll, Peng, Christoffersen, Damani, Slocum, Anwar, Siththaranjan, Nadeau, Michaud, Pfau, Krasheninnikov, Chen, Langosco, Hase, Bıyık, Dragan, Krueger, Sadigh, and Hadfield-Menell]{casper2023open}
Stephen Casper, Xander Davies, Claudia Shi, Thomas~Krendl Gilbert, Jérémy Scheurer, Javier Rando, Rachel Freedman, Tomasz Korbak, David Lindner, Pedro Freire, Tony Wang, Samuel Marks, Charbel-Raphaël Segerie, Micah Carroll, Andi Peng, Phillip Christoffersen, Mehul Damani, Stewart Slocum, Usman Anwar, Anand Siththaranjan, Max Nadeau, Eric~J. Michaud, Jacob Pfau, Dmitrii Krasheninnikov, Xin Chen, Lauro Langosco, Peter Hase, Erdem Bıyık, Anca Dragan, David Krueger, Dorsa Sadigh, and Dylan Hadfield-Menell.
\newblock Open problems and fundamental limitations of reinforcement learning from human feedback, 2023.

\bibitem[Chaudhari et~al.(2024)Chaudhari, Aggarwal, Murahari, Rajpurohit, Kalyan, Narasimhan, Deshpande, and da~Silva]{chaudhari2024rlhf}
Shreyas Chaudhari, Pranjal Aggarwal, Vishvak Murahari, Tanmay Rajpurohit, Ashwin Kalyan, Karthik Narasimhan, Ameet Deshpande, and Bruno~Castro da~Silva.
\newblock Rlhf deciphered: A critical analysis of reinforcement learning from human feedback for llms.
\newblock \emph{arXiv preprint arXiv:2404.08555}, 2024.

\bibitem[Choi et~al.(2024)Choi, Ahmadian, Geist, Pietquin, and Azar]{choi2024srpo}
Eugene Choi, Arash Ahmadian, Matthieu Geist, Oilvier Pietquin, and Mohammad~Gheshlaghi Azar.
\newblock Self-improving robust preference optimization, 2024.

\bibitem[Christiano et~al.(2017{\natexlab{a}})Christiano, Leike, Brown, Martic, Legg, and Amodei]{christiano2023RLHF}
Paul Christiano, Jan Leike, Tom~B. Brown, Miljan Martic, Shane Legg, and Dario Amodei.
\newblock Deep reinforcement learning from human preferences, 2017{\natexlab{a}}.

\bibitem[Christiano et~al.(2017{\natexlab{b}})Christiano, Leike, Brown, Martic, Legg, and Amodei]{NIPS2017_d5e2c0ad}
Paul~F Christiano, Jan Leike, Tom Brown, Miljan Martic, Shane Legg, and Dario Amodei.
\newblock Deep reinforcement learning from human preferences.
\newblock In I.~Guyon, U.~Von Luxburg, S.~Bengio, H.~Wallach, R.~Fergus, S.~Vishwanathan, and R.~Garnett (eds.), \emph{Advances in Neural Information Processing Systems}, volume~30. Curran Associates, Inc., 2017{\natexlab{b}}.
\newblock URL \url{https://proceedings.neurips.cc/paper_files/paper/2017/file/d5e2c0adad503c91f91df240d0cd4e49-Paper.pdf}.

\bibitem[Chung et~al.(2023)Chung, Kamar, and Amershi]{Chung_2023}
John Chung, Ece Kamar, and Saleema Amershi.
\newblock Increasing diversity while maintaining accuracy: Text data generation with large language models and human interventions.
\newblock In \emph{Proceedings of the 61st Annual Meeting of the Association for Computational Linguistics (Volume 1: Long Papers)}, pp.\  575--593, Toronto, Canada, July 2023. Association for Computational Linguistics.
\newblock \doi{10.18653/v1/2023.acl-long.34}.
\newblock URL \url{http://dx.doi.org/10.18653/v1/2023.acl-long.34}.

\bibitem[Conover et~al.(2023)Conover, Hayes, Mathur, Xie, Wan, Shah, Ghodsi, Wendell, Zaharia, and Xin]{DatabricksBlog2023DollyV2}
Mike Conover, Matt Hayes, Ankit Mathur, Jianwei Xie, Jun Wan, Sam Shah, Ali Ghodsi, Patrick Wendell, Matei Zaharia, and Reynold Xin.
\newblock Free dolly: Introducing the world's first truly open instruction-tuned llm, 2023.
\newblock URL \url{https://www.databricks.com/blog/2023/04/12/dolly-first-open-commercially-viable-instruction-tuned-llm}.

\bibitem[Deng et~al.(2023)Deng, Zhang, Pan, and Bing]{deng2023multilingual}
Yue Deng, Wenxuan Zhang, Sinno~Jialin Pan, and Lidong Bing.
\newblock Multilingual jailbreak challenges in large language models.
\newblock \emph{arXiv preprint arXiv:2310.06474}, 2023.

\bibitem[Deshpande et~al.(2023)Deshpande, Murahari, Rajpurohit, Kalyan, and Narasimhan]{deshpande2023toxicity}
Ameet Deshpande, Vishvak Murahari, Tanmay Rajpurohit, Ashwin Kalyan, and Karthik Narasimhan.
\newblock Toxicity in chatgpt: Analyzing persona-assigned language models.
\newblock \emph{arXiv preprint arXiv:2304.05335}, 2023.

\bibitem[Dodge et~al.(2021)Dodge, Sap, Marasovi{\'c}, Agnew, Ilharco, Groeneveld, Mitchell, and Gardner]{dodge2021documenting}
Jesse Dodge, Maarten Sap, Ana Marasovi{\'c}, William Agnew, Gabriel Ilharco, Dirk Groeneveld, Margaret Mitchell, and Matt Gardner.
\newblock Documenting large webtext corpora: A case study on the colossal clean crawled corpus.
\newblock In Marie-Francine Moens, Xuanjing Huang, Lucia Specia, and Scott Wen-tau Yih (eds.), \emph{Proceedings of the 2021 Conference on Empirical Methods in Natural Language Processing}, pp.\  1286--1305, Online and Punta Cana, Dominican Republic, November 2021. Association for Computational Linguistics.
\newblock \doi{10.18653/v1/2021.emnlp-main.98}.
\newblock URL \url{https://aclanthology.org/2021.emnlp-main.98}.

\bibitem[Dubois et~al.(2024)Dubois, Li, Taori, Zhang, Gulrajani, Ba, Guestrin, Liang, and Hashimoto]{dubois2024alpacafarm}
Yann Dubois, Xuechen Li, Rohan Taori, Tianyi Zhang, Ishaan Gulrajani, Jimmy Ba, Carlos Guestrin, Percy Liang, and Tatsunori~B. Hashimoto.
\newblock Alpacafarm: A simulation framework for methods that learn from human feedback, 2024.

\bibitem[Dutta~Chowdhury et~al.(2022)Dutta~Chowdhury, Jalota, Espa{\~n}a-Bonet, and Genabith]{dutta-chowdhury-etal-2022-towards}
Koel Dutta~Chowdhury, Rricha Jalota, Cristina Espa{\~n}a-Bonet, and Josef Genabith.
\newblock Towards debiasing translation artifacts.
\newblock In Marine Carpuat, Marie-Catherine de~Marneffe, and Ivan~Vladimir Meza~Ruiz (eds.), \emph{Proceedings of the 2022 Conference of the North American Chapter of the Association for Computational Linguistics: Human Language Technologies}, pp.\  3983--3991, Seattle, United States, July 2022. Association for Computational Linguistics.
\newblock \doi{10.18653/v1/2022.naacl-main.292}.
\newblock URL \url{https://aclanthology.org/2022.naacl-main.292}.

\bibitem[Ethayarajh et~al.(2024)Ethayarajh, Xu, Muennighoff, Jurafsky, and Kiela]{ethayarajh2024kto}
Kawin Ethayarajh, Winnie Xu, Niklas Muennighoff, Dan Jurafsky, and Douwe Kiela.
\newblock Kto: Model alignment as prospect theoretic optimization.
\newblock \emph{arXiv preprint arXiv:2402.01306}, 2024.

\bibitem[Falck et~al.(2012)Falck, Heblich, Lameli, and S{\"u}dekum]{falck2012dialects}
Oliver Falck, Stephan Heblich, Alfred Lameli, and Jens S{\"u}dekum.
\newblock Dialects, cultural identity, and economic exchange.
\newblock \emph{Journal of urban economics}, 72\penalty0 (2-3):\penalty0 225--239, 2012.

\bibitem[$\forall$ et~al.(2020)$\forall$, Nekoto, Marivate, Matsila, Fasubaa, Fagbohungbe, Akinola, Muhammad, Kabongo~Kabenamualu, Osei, Sackey, Niyongabo, Macharm, Ogayo, Ahia, Berhe, Adeyemi, Mokgesi-Selinga, Okegbemi, Martinus, Tajudeen, Degila, Ogueji, Siminyu, Kreutzer, Webster, Ali, Abbott, Orife, Ezeani, Dangana, Kamper, Elsahar, Duru, Kioko, Espoir, van Biljon, Whitenack, Onyefuluchi, Emezue, Dossou, Sibanda, Bassey, Olabiyi, Ramkilowan, {\"O}ktem, Akinfaderin, and Bashir]{nekoto-etal-2020-participatory-forall}
{}~$\forall$, Wilhelmina Nekoto, Vukosi Marivate, Tshinondiwa Matsila, Timi Fasubaa, Taiwo Fagbohungbe, Solomon~Oluwole Akinola, Shamsuddeen Muhammad, Salomon Kabongo~Kabenamualu, Salomey Osei, Freshia Sackey, Rubungo~Andre Niyongabo, Ricky Macharm, Perez Ogayo, Orevaoghene Ahia, Musie~Meressa Berhe, Mofetoluwa Adeyemi, Masabata Mokgesi-Selinga, Lawrence Okegbemi, Laura Martinus, Kolawole Tajudeen, Kevin Degila, Kelechi Ogueji, Kathleen Siminyu, Julia Kreutzer, Jason Webster, Jamiil~Toure Ali, Jade Abbott, Iroro Orife, Ignatius Ezeani, Idris~Abdulkadir Dangana, Herman Kamper, Hady Elsahar, Goodness Duru, Ghollah Kioko, Murhabazi Espoir, Elan van Biljon, Daniel Whitenack, Christopher Onyefuluchi, Chris~Chinenye Emezue, Bonaventure F.~P. Dossou, Blessing Sibanda, Blessing Bassey, Ayodele Olabiyi, Arshath Ramkilowan, Alp {\"O}ktem, Adewale Akinfaderin, and Abdallah Bashir.
\newblock Participatory research for low-resourced machine translation: A case study in {A}frican languages.
\newblock In \emph{Findings of the Association for Computational Linguistics: EMNLP 2020}, pp.\  2144--2160, Online, November 2020. Association for Computational Linguistics.
\newblock \doi{10.18653/v1/2020.findings-emnlp.195}.
\newblock URL \url{https://aclanthology.org/2020.findings-emnlp.195}.

\bibitem[Gao et~al.(2021)Gao, Tow, Biderman, Black, DiPofi, Foster, Golding, Hsu, McDonell, Muennighoff, et~al.]{gao2021framework}
Leo Gao, Jonathan Tow, Stella Biderman, Sid Black, Anthony DiPofi, Charles Foster, Laurence Golding, Jeffrey Hsu, Kyle McDonell, Niklas Muennighoff, et~al.
\newblock A framework for few-shot language model evaluation.
\newblock \emph{Version v0. 0.1. Sept}, pp.\ ~8, 2021.

\bibitem[Gao et~al.(2022{\natexlab{a}})Gao, Schulman, and Hilton]{Gao2022ScalingLF}
Leo Gao, John Schulman, and Jacob Hilton.
\newblock Scaling laws for reward model overoptimization.
\newblock In \emph{International Conference on Machine Learning}, 2022{\natexlab{a}}.
\newblock URL \url{https://api.semanticscholar.org/CorpusID:252992904}.

\bibitem[Gao et~al.(2022{\natexlab{b}})Gao, Schulman, and Hilton]{gao2022scaling}
Leo Gao, John Schulman, and Jacob Hilton.
\newblock Scaling laws for reward model overoptimization, 2022{\natexlab{b}}.

\bibitem[Gemini-Team et~al.(2024)Gemini-Team, Anil, Borgeaud, Alayrac, Yu, Soricut, Schalkwyk, Dai, Hauth, Millican, Silver, Johnson, Antonoglou, Schrittwieser, Glaese, Chen, Pitler, Lillicrap, Lazaridou, Firat, Molloy, Isard, Barham, Hennigan, Lee, Viola, Reynolds, Xu, Doherty, Collins, Meyer, Rutherford, Moreira, Ayoub, Goel, Krawczyk, Du, Chi, Cheng, Ni, Shah, Kane, Chan, Faruqui, Severyn, Lin, Li, Cheng, Ittycheriah, Mahdieh, Chen, Sun, Tran, Bagri, Lakshminarayanan, Liu, Orban, Güra, Zhou, Song, Boffy, Ganapathy, Zheng, Choe, Ágoston Weisz, Zhu, Lu, Gopal, Kahn, Kula, Pitman, Shah, Taropa, Merey, Baeuml, Chen, Shafey, Zhang, Sercinoglu, Tucker, Piqueras, Krikun, Barr, Savinov, Danihelka, Roelofs, White, Andreassen, von Glehn, Yagati, Kazemi, Gonzalez, Khalman, Sygnowski, Frechette, Smith, Culp, Proleev, Luan, Chen, Lottes, Schucher, Lebron, Rrustemi, Clay, Crone, Kocisky, Zhao, Perz, Yu, Howard, Bloniarz, Rae, Lu, Sifre, Maggioni, Alcober, Garrette, Barnes, Thakoor, Austin, Barth-Maron, Wong, Joshi,
  Chaabouni, Fatiha, Ahuja, Tomar, Senter, Chadwick, Kornakov, Attaluri, Iturrate, Liu, Li, Cogan, Chen, Jia, Gu, Zhang, Grimstad, Hartman, Garcia, Pillai, Devlin, Laskin, de~Las~Casas, Valter, Tao, Blanco, Badia, Reitter, Chen, Brennan, Rivera, Brin, Iqbal, Surita, Labanowski, Rao, Winkler, Parisotto, Gu, Olszewska, Addanki, Miech, Louis, Teplyashin, Brown, Catt, Balaguer, Xiang, Wang, Ashwood, Briukhov, Webson, Ganapathy, Sanghavi, Kannan, Chang, Stjerngren, Djolonga, Sun, Bapna, Aitchison, Pejman, Michalewski, Yu, Wang, Love, Ahn, Bloxwich, Han, Humphreys, Sellam, Bradbury, Godbole, Samangooei, Damoc, Kaskasoli, Arnold, Vasudevan, Agrawal, Riesa, Lepikhin, Tanburn, Srinivasan, Lim, Hodkinson, Shyam, Ferret, Hand, Garg, Paine, Li, Li, Giang, Neitz, Abbas, York, Reid, Cole, Chowdhery, Das, Rogozińska, Nikolaev, Sprechmann, Nado, Zilka, Prost, He, Monteiro, Mishra, Welty, Newlan, Jia, Allamanis, Hu, de~Liedekerke, Gilmer, Saroufim, Rijhwani, Hou, Shrivastava, Baddepudi, Goldin, Ozturel, Cassirer, Xu, Sohn,
  Sachan, Amplayo, Swanson, Petrova, Narayan, Guez, Brahma, Landon, Patel, Zhao, Villela, Wang, Jia, Rahtz, Giménez, Yeung, Keeling, Georgiev, Mincu, Wu, Haykal, Saputro, Vodrahalli, Qin, Cankara, Sharma, Fernando, Hawkins, Neyshabur, Kim, Hutter, Agrawal, Castro-Ros, van~den Driessche, Wang, Yang, yiin Chang, Komarek, McIlroy, Lučić, Zhang, Farhan, Sharman, Natsev, Michel, Bansal, Qiao, Cao, Shakeri, Butterfield, Chung, Rubenstein, Agrawal, Mensch, Soparkar, Lenc, Chung, Pope, Maggiore, Kay, Jhakra, Wang, Maynez, Phuong, Tobin, Tacchetti, Trebacz, Robinson, Katariya, Riedel, Bailey, Xiao, Ghelani, Aroyo, Slone, Houlsby, Xiong, Yang, Gribovskaya, Adler, Wirth, Lee, Li, Kagohara, Pavagadhi, Bridgers, Bortsova, Ghemawat, Ahmed, Liu, Powell, Bolina, Iinuma, Zablotskaia, Besley, Chung, Dozat, Comanescu, Si, Greer, Su, Polacek, Kaufman, Tokumine, Hu, Buchatskaya, Miao, Elhawaty, Siddhant, Tomasev, Xing, Greer, Miller, Ashraf, Roy, Zhang, Ma, Filos, Besta, Blevins, Klimenko, Yeh, Changpinyo, Mu, Chang,
  Pajarskas, Muir, Cohen, Lan, Haridasan, Marathe, Hansen, Douglas, Samuel, Wang, Austin, Lan, Jiang, Chiu, Lorenzo, Sjösund, Cevey, Gleicher, Avrahami, Boral, Srinivasan, Selo, May, Aisopos, Hussenot, Soares, Baumli, Chang, Recasens, Caine, Pritzel, Pavetic, Pardo, Gergely, Frye, Ramasesh, Horgan, Badola, Kassner, Roy, Dyer, Campos, Tomala, Tang, Badawy, White, Mustafa, Lang, Jindal, Vikram, Gong, Caelles, Hemsley, Thornton, Feng, Stokowiec, Zheng, Thacker, Çağlar Ünlü, Zhang, Saleh, Svensson, Bileschi, Patil, Anand, Ring, Tsihlas, Vezer, Selvi, Shevlane, Rodriguez, Kwiatkowski, Daruki, Rong, Dafoe, FitzGerald, Gu-Lemberg, Khan, Hendricks, Pellat, Feinberg, Cobon-Kerr, Sainath, Rauh, Hashemi, Ives, Hasson, Noland, Cao, Byrd, Hou, Wang, Sottiaux, Paganini, Lespiau, Moufarek, Hassan, Shivakumar, van Amersfoort, Mandhane, Joshi, Goyal, Tung, Brock, Sheahan, Misra, Li, Rakićević, Dehghani, Liu, Mittal, Oh, Noury, Sezener, Huot, Lamm, Cao, Chen, Mudgal, Stella, Brooks, Vasudevan, Liu, Chain, Melinkeri,
  Cohen, Wang, Seymore, Zubkov, Goel, Yue, Krishnakumaran, Albert, Hurley, Sano, Mohananey, Joughin, Filonov, Kępa, Eldawy, Lim, Rishi, Badiezadegan, Bos, Chang, Jain, Padmanabhan, Puttagunta, Krishna, Baker, Kalb, Bedapudi, Kurzrok, Lei, Yu, Litvin, Zhou, Wu, Sobell, Siciliano, Papir, Neale, Bragagnolo, Toor, Chen, Anklin, Wang, Feng, Gholami, Ling, Liu, Walter, Moghaddam, Kishore, Adamek, Mercado, Mallinson, Wandekar, Cagle, Ofek, Garrido, Lombriser, Mukha, Sun, Mohammad, Matak, Qian, Peswani, Janus, Yuan, Schelin, David, Garg, He, Duzhyi, Älgmyr, Lottaz, Li, Yadav, Xu, Chinien, Shivanna, Chuklin, Li, Spadine, Wolfe, Mohamed, Das, Dai, He, von Dincklage, Upadhyay, Maurya, Chi, Krause, Salama, Rabinovitch, M, Selvan, Dektiarev, Ghiasi, Guven, Gupta, Liu, Sharma, Shtacher, Paul, Akerlund, Aubet, Huang, Zhu, Zhu, Teixeira, Fritze, Bertolini, Marinescu, Bölle, Paulus, Gupta, Latkar, Chang, Sanders, Wilson, Wu, Tan, Thiet, Doshi, Lall, Mishra, Chen, Luong, Benjamin, Lee, Andrejczuk, Rabiej, Ranjan, Styrc,
  Yin, Simon, Harriott, Bansal, Robsky, Bacon, Greene, Mirylenka, Zhou, Sarvana, Goyal, Andermatt, Siegler, Horn, Israel, Pongetti, Chen, Selvatici, Silva, Wang, Tolins, Guu, Yogev, Cai, Agostini, Shah, Nguyen, Donnaile, Pereira, Friso, Stambler, Kurzrok, Kuang, Romanikhin, Geller, Yan, Jang, Lee, Fica, Malmi, Tan, Banica, Balle, Pham, Huang, Avram, Shi, Singh, Hidey, Ahuja, Saxena, Dooley, Potharaju, O'Neill, Gokulchandran, Foley, Zhao, Dusenberry, Liu, Mehta, Kotikalapudi, Safranek-Shrader, Goodman, Kessinger, Globen, Kolhar, Gorgolewski, Ibrahim, Song, Eichenbaum, Brovelli, Potluri, Lahoti, Baetu, Ghorbani, Chen, Crawford, Pal, Sridhar, Gurita, Mujika, Petrovski, Cedoz, Li, Chen, Santo, Goyal, Punjabi, Kappaganthu, Kwak, LV, Velury, Choudhury, Hall, Shah, Figueira, Thomas, Lu, Zhou, Kumar, Jurdi, Chikkerur, Ma, Yu, Kwak, Ähdel, Rajayogam, Choma, Liu, Barua, Ji, Park, Hellendoorn, Bailey, Bilal, Zhou, Khatir, Sutton, Rzadkowski, Macintosh, Shagin, Medina, Liang, Zhou, Shah, Bi, Dankovics, Banga, Lehmann,
  Bredesen, Lin, Hoffmann, Lai, Chung, Yang, Balani, Bražinskas, Sozanschi, Hayes, Alcalde, Makarov, Chen, Stella, Snijders, Mandl, Kärrman, Nowak, Wu, Dyck, Vaidyanathan, R, Mallet, Rudominer, Johnston, Mittal, Udathu, Christensen, Verma, Irving, Santucci, Elsayed, Davoodi, Georgiev, Tenney, Hua, Cideron, Leurent, Alnahlawi, Georgescu, Wei, Zheng, Scandinaro, Jiang, Snoek, Sundararajan, Wang, Ontiveros, Karo, Cole, Rajashekhar, Tumeh, Ben-David, Jain, Uesato, Datta, Bunyan, Wu, Zhang, Stanczyk, Zhang, Steiner, Naskar, Azzam, Johnson, Paszke, Chiu, Elias, Mohiuddin, Muhammad, Miao, Lee, Vieillard, Park, Zhang, Stanway, Garmon, Karmarkar, Dong, Lee, Kumar, Zhou, Evens, Isaac, Irving, Loper, Fink, Arkatkar, Chen, Shafran, Petrychenko, Chen, Jia, Levskaya, Zhu, Grabowski, Mao, Magni, Yao, Snaider, Casagrande, Palmer, Suganthan, Castaño, Giannoumis, Kim, Rybiński, Sreevatsa, Prendki, Soergel, Goedeckemeyer, Gierke, Jafari, Gaba, Wiesner, Wright, Wei, Vashisht, Kulizhskaya, Hoover, Le, Li, Iwuanyanwu, Liu,
  Ramirez, Khorlin, Cui, LIN, Wu, Aguilar, Pallo, Chakladar, Perng, Abellan, Zhang, Dasgupta, Kushman, Penchev, Repina, Wu, van~der Weide, Ponnapalli, Kaplan, Simsa, Li, Dousse, Yang, Piper, Ie, Pasumarthi, Lintz, Vijayakumar, Andor, Valenzuela, Lui, Paduraru, Peng, Lee, Zhang, Greene, Nguyen, Kurylowicz, Hardin, Dixon, Janzer, Choo, Feng, Zhang, Singhal, Du, McKinnon, Antropova, Bolukbasi, Keller, Reid, Finchelstein, Raad, Crocker, Hawkins, Dadashi, Gaffney, Franko, Bulanova, Leblond, Chung, Askham, Cobo, Xu, Fischer, Xu, Sorokin, Alberti, Lin, Evans, Dimitriev, Forbes, Banarse, Tung, Omernick, Bishop, Sterneck, Jain, Xia, Amid, Piccinno, Wang, Banzal, Mankowitz, Polozov, Krakovna, Brown, Bateni, Duan, Firoiu, Thotakuri, Natan, Geist, tan Girgin, Li, Ye, Roval, Tojo, Kwong, Lee-Thorp, Yew, Sinopalnikov, Ramos, Mellor, Sharma, Wu, Miller, Sonnerat, Vnukov, Greig, Beattie, Caveness, Bai, Eisenschlos, Korchemniy, Tsai, Jasarevic, Kong, Dao, Zheng, Liu, Yang, Zhu, Teh, Sanmiya, Gladchenko, Trdin, Toyama, Rosen,
  Tavakkol, Xue, Elkind, Woodman, Carpenter, Papamakarios, Kemp, Kafle, Grunina, Sinha, Talbert, Wu, Owusu-Afriyie, Du, Thornton, Pont-Tuset, Narayana, Li, Fatehi, Wieting, Ajmeri, Uria, Ko, Knight, Héliou, Niu, Gu, Pang, Li, Levine, Stolovich, Santamaria-Fernandez, Goenka, Yustalim, Strudel, Elqursh, Deck, Lee, Li, Levin, Hoffmann, Holtmann-Rice, Bachem, Arora, Koh, Yeganeh, Põder, Tariq, Sun, Ionita, Seyedhosseini, Tafti, Liu, Gulati, Liu, Ye, Chrzaszcz, Wang, Sethi, Li, Brown, Singh, Fan, Parisi, Stanton, Koverkathu, Choquette-Choo, Li, Lu, Ittycheriah, Shroff, Varadarajan, Bahargam, Willoughby, Gaddy, Desjardins, Cornero, Robenek, Mittal, Albrecht, Shenoy, Moiseev, Jacobsson, Ghaffarkhah, Rivière, Walton, Crepy, Parrish, Zhou, Farabet, Radebaugh, Srinivasan, van~der Salm, Fidjeland, Scellato, Latorre-Chimoto, Klimczak-Plucińska, Bridson, de~Cesare, Hudson, Mendolicchio, Walker, Morris, Mauger, Guseynov, Reid, Odoom, Loher, Cotruta, Yenugula, Grewe, Petrushkina, Duerig, Sanchez, Yadlowsky, Shen,
  Globerson, Webb, Dua, Li, Bhupatiraju, Hurt, Qureshi, Agarwal, Shani, Eyal, Khare, Belle, Wang, Tekur, Kale, Wei, Sang, Saeta, Liechty, Sun, Zhao, Lee, Nayak, Fritz, Vuyyuru, Aslanides, Vyas, Wicke, Ma, Eltyshev, Martin, Cate, Manyika, Amiri, Kim, Xiong, Kang, Luisier, Tripuraneni, Madras, Guo, Waters, Wang, Ainslie, Baldridge, Zhang, Pruthi, Bauer, Yang, Mansour, Gelman, Xu, Polovets, Liu, Cai, Chen, Sheng, Xue, Ozair, Angermueller, Li, Sinha, Wang, Wiesinger, Koukoumidis, Tian, Iyer, Gurumurthy, Goldenson, Shah, Blake, Yu, Urbanowicz, Palomaki, Fernando, Durden, Mehta, Momchev, Rahimtoroghi, Georgaki, Raul, Ruder, Redshaw, Lee, Zhou, Jalan, Li, Hechtman, Schuh, Nasr, Milan, Mikulik, Franco, Green, Nguyen, Kelley, Mahendru, Hu, Howland, Vargas, Hui, Bansal, Rao, Ghiya, Wang, Ye, Sarr, Preston, Elish, Li, Kaku, Gupta, Pasupat, Juan, Someswar, M., Chen, Amini, Fabrikant, Chu, Dong, Muthal, Buthpitiya, Jauhari, Hua, Khandelwal, Hitron, Ren, Rinaldi, Drath, Dabush, Jiang, Godhia, Sachs, Chen, Fan, Taitelbaum,
  Noga, Dai, Wang, Liang, Hamer, Ferng, Elkind, Atias, Lee, Listík, Carlen, van~de Kerkhof, Pikus, Zaher, Müller, Zykova, Stefanec, Gatsko, Hirnschall, Sethi, Xu, Ahuja, Tsai, Stefanoiu, Feng, Dhandhania, Katyal, Gupta, Parulekar, Pitta, Zhao, Bhatia, Bhavnani, Alhadlaq, Li, Danenberg, Tu, Pine, Filippova, Ghosh, Limonchik, Urala, Lanka, Clive, Sun, Li, Wu, Hongtongsak, Li, Thakkar, Omarov, Majmundar, Alverson, Kucharski, Patel, Jain, Zabelin, Pelagatti, Kohli, Kumar, Kim, Sankar, Shah, Ramachandruni, Zeng, Bariach, Weidinger, Subramanya, Hsiao, Hassabis, Kavukcuoglu, Sadovsky, Le, Strohman, Wu, Petrov, Dean, and Vinyals]{geminiteam2024gemini}
Gemini-Team, Rohan Anil, Sebastian Borgeaud, Jean-Baptiste Alayrac, Jiahui Yu, Radu Soricut, Johan Schalkwyk, Andrew~M. Dai, Anja Hauth, Katie Millican, David Silver, Melvin Johnson, Ioannis Antonoglou, Julian Schrittwieser, Amelia Glaese, Jilin Chen, Emily Pitler, Timothy Lillicrap, Angeliki Lazaridou, Orhan Firat, James Molloy, Michael Isard, Paul~R. Barham, Tom Hennigan, Benjamin Lee, Fabio Viola, Malcolm Reynolds, Yuanzhong Xu, Ryan Doherty, Eli Collins, Clemens Meyer, Eliza Rutherford, Erica Moreira, Kareem Ayoub, Megha Goel, Jack Krawczyk, Cosmo Du, Ed~Chi, Heng-Tze Cheng, Eric Ni, Purvi Shah, Patrick Kane, Betty Chan, Manaal Faruqui, Aliaksei Severyn, Hanzhao Lin, YaGuang Li, Yong Cheng, Abe Ittycheriah, Mahdis Mahdieh, Mia Chen, Pei Sun, Dustin Tran, Sumit Bagri, Balaji Lakshminarayanan, Jeremiah Liu, Andras Orban, Fabian Güra, Hao Zhou, Xinying Song, Aurelien Boffy, Harish Ganapathy, Steven Zheng, HyunJeong Choe, Ágoston Weisz, Tao Zhu, Yifeng Lu, Siddharth Gopal, Jarrod Kahn, Maciej Kula, Jeff
  Pitman, Rushin Shah, Emanuel Taropa, Majd~Al Merey, Martin Baeuml, Zhifeng Chen, Laurent~El Shafey, Yujing Zhang, Olcan Sercinoglu, George Tucker, Enrique Piqueras, Maxim Krikun, Iain Barr, Nikolay Savinov, Ivo Danihelka, Becca Roelofs, Anaïs White, Anders Andreassen, Tamara von Glehn, Lakshman Yagati, Mehran Kazemi, Lucas Gonzalez, Misha Khalman, Jakub Sygnowski, Alexandre Frechette, Charlotte Smith, Laura Culp, Lev Proleev, Yi~Luan, Xi~Chen, James Lottes, Nathan Schucher, Federico Lebron, Alban Rrustemi, Natalie Clay, Phil Crone, Tomas Kocisky, Jeffrey Zhao, Bartek Perz, Dian Yu, Heidi Howard, Adam Bloniarz, Jack~W. Rae, Han Lu, Laurent Sifre, Marcello Maggioni, Fred Alcober, Dan Garrette, Megan Barnes, Shantanu Thakoor, Jacob Austin, Gabriel Barth-Maron, William Wong, Rishabh Joshi, Rahma Chaabouni, Deeni Fatiha, Arun Ahuja, Gaurav~Singh Tomar, Evan Senter, Martin Chadwick, Ilya Kornakov, Nithya Attaluri, Iñaki Iturrate, Ruibo Liu, Yunxuan Li, Sarah Cogan, Jeremy Chen, Chao Jia, Chenjie Gu, Qiao Zhang,
  Jordan Grimstad, Ale~Jakse Hartman, Xavier Garcia, Thanumalayan~Sankaranarayana Pillai, Jacob Devlin, Michael Laskin, Diego de~Las~Casas, Dasha Valter, Connie Tao, Lorenzo Blanco, Adrià~Puigdomènech Badia, David Reitter, Mianna Chen, Jenny Brennan, Clara Rivera, Sergey Brin, Shariq Iqbal, Gabriela Surita, Jane Labanowski, Abhi Rao, Stephanie Winkler, Emilio Parisotto, Yiming Gu, Kate Olszewska, Ravi Addanki, Antoine Miech, Annie Louis, Denis Teplyashin, Geoff Brown, Elliot Catt, Jan Balaguer, Jackie Xiang, Pidong Wang, Zoe Ashwood, Anton Briukhov, Albert Webson, Sanjay Ganapathy, Smit Sanghavi, Ajay Kannan, Ming-Wei Chang, Axel Stjerngren, Josip Djolonga, Yuting Sun, Ankur Bapna, Matthew Aitchison, Pedram Pejman, Henryk Michalewski, Tianhe Yu, Cindy Wang, Juliette Love, Junwhan Ahn, Dawn Bloxwich, Kehang Han, Peter Humphreys, Thibault Sellam, James Bradbury, Varun Godbole, Sina Samangooei, Bogdan Damoc, Alex Kaskasoli, Sébastien M.~R. Arnold, Vijay Vasudevan, Shubham Agrawal, Jason Riesa, Dmitry
  Lepikhin, Richard Tanburn, Srivatsan Srinivasan, Hyeontaek Lim, Sarah Hodkinson, Pranav Shyam, Johan Ferret, Steven Hand, Ankush Garg, Tom~Le Paine, Jian Li, Yujia Li, Minh Giang, Alexander Neitz, Zaheer Abbas, Sarah York, Machel Reid, Elizabeth Cole, Aakanksha Chowdhery, Dipanjan Das, Dominika Rogozińska, Vitaliy Nikolaev, Pablo Sprechmann, Zachary Nado, Lukas Zilka, Flavien Prost, Luheng He, Marianne Monteiro, Gaurav Mishra, Chris Welty, Josh Newlan, Dawei Jia, Miltiadis Allamanis, Clara~Huiyi Hu, Raoul de~Liedekerke, Justin Gilmer, Carl Saroufim, Shruti Rijhwani, Shaobo Hou, Disha Shrivastava, Anirudh Baddepudi, Alex Goldin, Adnan Ozturel, Albin Cassirer, Yunhan Xu, Daniel Sohn, Devendra Sachan, Reinald~Kim Amplayo, Craig Swanson, Dessie Petrova, Shashi Narayan, Arthur Guez, Siddhartha Brahma, Jessica Landon, Miteyan Patel, Ruizhe Zhao, Kevin Villela, Luyu Wang, Wenhao Jia, Matthew Rahtz, Mai Giménez, Legg Yeung, James Keeling, Petko Georgiev, Diana Mincu, Boxi Wu, Salem Haykal, Rachel Saputro, Kiran
  Vodrahalli, James Qin, Zeynep Cankara, Abhanshu Sharma, Nick Fernando, Will Hawkins, Behnam Neyshabur, Solomon Kim, Adrian Hutter, Priyanka Agrawal, Alex Castro-Ros, George van~den Driessche, Tao Wang, Fan Yang, Shuo yiin Chang, Paul Komarek, Ross McIlroy, Mario Lučić, Guodong Zhang, Wael Farhan, Michael Sharman, Paul Natsev, Paul Michel, Yamini Bansal, Siyuan Qiao, Kris Cao, Siamak Shakeri, Christina Butterfield, Justin Chung, Paul~Kishan Rubenstein, Shivani Agrawal, Arthur Mensch, Kedar Soparkar, Karel Lenc, Timothy Chung, Aedan Pope, Loren Maggiore, Jackie Kay, Priya Jhakra, Shibo Wang, Joshua Maynez, Mary Phuong, Taylor Tobin, Andrea Tacchetti, Maja Trebacz, Kevin Robinson, Yash Katariya, Sebastian Riedel, Paige Bailey, Kefan Xiao, Nimesh Ghelani, Lora Aroyo, Ambrose Slone, Neil Houlsby, Xuehan Xiong, Zhen Yang, Elena Gribovskaya, Jonas Adler, Mateo Wirth, Lisa Lee, Music Li, Thais Kagohara, Jay Pavagadhi, Sophie Bridgers, Anna Bortsova, Sanjay Ghemawat, Zafarali Ahmed, Tianqi Liu, Richard Powell,
  Vijay Bolina, Mariko Iinuma, Polina Zablotskaia, James Besley, Da-Woon Chung, Timothy Dozat, Ramona Comanescu, Xiance Si, Jeremy Greer, Guolong Su, Martin Polacek, Raphaël~Lopez Kaufman, Simon Tokumine, Hexiang Hu, Elena Buchatskaya, Yingjie Miao, Mohamed Elhawaty, Aditya Siddhant, Nenad Tomasev, Jinwei Xing, Christina Greer, Helen Miller, Shereen Ashraf, Aurko Roy, Zizhao Zhang, Ada Ma, Angelos Filos, Milos Besta, Rory Blevins, Ted Klimenko, Chih-Kuan Yeh, Soravit Changpinyo, Jiaqi Mu, Oscar Chang, Mantas Pajarskas, Carrie Muir, Vered Cohen, Charline~Le Lan, Krishna Haridasan, Amit Marathe, Steven Hansen, Sholto Douglas, Rajkumar Samuel, Mingqiu Wang, Sophia Austin, Chang Lan, Jiepu Jiang, Justin Chiu, Jaime~Alonso Lorenzo, Lars~Lowe Sjösund, Sébastien Cevey, Zach Gleicher, Thi Avrahami, Anudhyan Boral, Hansa Srinivasan, Vittorio Selo, Rhys May, Konstantinos Aisopos, Léonard Hussenot, Livio~Baldini Soares, Kate Baumli, Michael~B. Chang, Adrià Recasens, Ben Caine, Alexander Pritzel, Filip Pavetic,
  Fabio Pardo, Anita Gergely, Justin Frye, Vinay Ramasesh, Dan Horgan, Kartikeya Badola, Nora Kassner, Subhrajit Roy, Ethan Dyer, Víctor~Campos Campos, Alex Tomala, Yunhao Tang, Dalia~El Badawy, Elspeth White, Basil Mustafa, Oran Lang, Abhishek Jindal, Sharad Vikram, Zhitao Gong, Sergi Caelles, Ross Hemsley, Gregory Thornton, Fangxiaoyu Feng, Wojciech Stokowiec, Ce~Zheng, Phoebe Thacker, Çağlar Ünlü, Zhishuai Zhang, Mohammad Saleh, James Svensson, Max Bileschi, Piyush Patil, Ankesh Anand, Roman Ring, Katerina Tsihlas, Arpi Vezer, Marco Selvi, Toby Shevlane, Mikel Rodriguez, Tom Kwiatkowski, Samira Daruki, Keran Rong, Allan Dafoe, Nicholas FitzGerald, Keren Gu-Lemberg, Mina Khan, Lisa~Anne Hendricks, Marie Pellat, Vladimir Feinberg, James Cobon-Kerr, Tara Sainath, Maribeth Rauh, Sayed~Hadi Hashemi, Richard Ives, Yana Hasson, Eric Noland, Yuan Cao, Nathan Byrd, Le~Hou, Qingze Wang, Thibault Sottiaux, Michela Paganini, Jean-Baptiste Lespiau, Alexandre Moufarek, Samer Hassan, Kaushik Shivakumar, Joost van
  Amersfoort, Amol Mandhane, Pratik Joshi, Anirudh Goyal, Matthew Tung, Andrew Brock, Hannah Sheahan, Vedant Misra, Cheng Li, Nemanja Rakićević, Mostafa Dehghani, Fangyu Liu, Sid Mittal, Junhyuk Oh, Seb Noury, Eren Sezener, Fantine Huot, Matthew Lamm, Nicola~De Cao, Charlie Chen, Sidharth Mudgal, Romina Stella, Kevin Brooks, Gautam Vasudevan, Chenxi Liu, Mainak Chain, Nivedita Melinkeri, Aaron Cohen, Venus Wang, Kristie Seymore, Sergey Zubkov, Rahul Goel, Summer Yue, Sai Krishnakumaran, Brian Albert, Nate Hurley, Motoki Sano, Anhad Mohananey, Jonah Joughin, Egor Filonov, Tomasz Kępa, Yomna Eldawy, Jiawern Lim, Rahul Rishi, Shirin Badiezadegan, Taylor Bos, Jerry Chang, Sanil Jain, Sri Gayatri~Sundara Padmanabhan, Subha Puttagunta, Kalpesh Krishna, Leslie Baker, Norbert Kalb, Vamsi Bedapudi, Adam Kurzrok, Shuntong Lei, Anthony Yu, Oren Litvin, Xiang Zhou, Zhichun Wu, Sam Sobell, Andrea Siciliano, Alan Papir, Robby Neale, Jonas Bragagnolo, Tej Toor, Tina Chen, Valentin Anklin, Feiran Wang, Richie Feng, Milad
  Gholami, Kevin Ling, Lijuan Liu, Jules Walter, Hamid Moghaddam, Arun Kishore, Jakub Adamek, Tyler Mercado, Jonathan Mallinson, Siddhinita Wandekar, Stephen Cagle, Eran Ofek, Guillermo Garrido, Clemens Lombriser, Maksim Mukha, Botu Sun, Hafeezul~Rahman Mohammad, Josip Matak, Yadi Qian, Vikas Peswani, Pawel Janus, Quan Yuan, Leif Schelin, Oana David, Ankur Garg, Yifan He, Oleksii Duzhyi, Anton Älgmyr, Timothée Lottaz, Qi~Li, Vikas Yadav, Luyao Xu, Alex Chinien, Rakesh Shivanna, Aleksandr Chuklin, Josie Li, Carrie Spadine, Travis Wolfe, Kareem Mohamed, Subhabrata Das, Zihang Dai, Kyle He, Daniel von Dincklage, Shyam Upadhyay, Akanksha Maurya, Luyan Chi, Sebastian Krause, Khalid Salama, Pam~G Rabinovitch, Pavan Kumar~Reddy M, Aarush Selvan, Mikhail Dektiarev, Golnaz Ghiasi, Erdem Guven, Himanshu Gupta, Boyi Liu, Deepak Sharma, Idan~Heimlich Shtacher, Shachi Paul, Oscar Akerlund, François-Xavier Aubet, Terry Huang, Chen Zhu, Eric Zhu, Elico Teixeira, Matthew Fritze, Francesco Bertolini, Liana-Eleonora
  Marinescu, Martin Bölle, Dominik Paulus, Khyatti Gupta, Tejasi Latkar, Max Chang, Jason Sanders, Roopa Wilson, Xuewei Wu, Yi-Xuan Tan, Lam~Nguyen Thiet, Tulsee Doshi, Sid Lall, Swaroop Mishra, Wanming Chen, Thang Luong, Seth Benjamin, Jasmine Lee, Ewa Andrejczuk, Dominik Rabiej, Vipul Ranjan, Krzysztof Styrc, Pengcheng Yin, Jon Simon, Malcolm~Rose Harriott, Mudit Bansal, Alexei Robsky, Geoff Bacon, David Greene, Daniil Mirylenka, Chen Zhou, Obaid Sarvana, Abhimanyu Goyal, Samuel Andermatt, Patrick Siegler, Ben Horn, Assaf Israel, Francesco Pongetti, Chih-Wei~"Louis" Chen, Marco Selvatici, Pedro Silva, Kathie Wang, Jackson Tolins, Kelvin Guu, Roey Yogev, Xiaochen Cai, Alessandro Agostini, Maulik Shah, Hung Nguyen, Noah~Ó Donnaile, Sébastien Pereira, Linda Friso, Adam Stambler, Adam Kurzrok, Chenkai Kuang, Yan Romanikhin, Mark Geller, ZJ~Yan, Kane Jang, Cheng-Chun Lee, Wojciech Fica, Eric Malmi, Qijun Tan, Dan Banica, Daniel Balle, Ryan Pham, Yanping Huang, Diana Avram, Hongzhi Shi, Jasjot Singh, Chris
  Hidey, Niharika Ahuja, Pranab Saxena, Dan Dooley, Srividya~Pranavi Potharaju, Eileen O'Neill, Anand Gokulchandran, Ryan Foley, Kai Zhao, Mike Dusenberry, Yuan Liu, Pulkit Mehta, Ragha Kotikalapudi, Chalence Safranek-Shrader, Andrew Goodman, Joshua Kessinger, Eran Globen, Prateek Kolhar, Chris Gorgolewski, Ali Ibrahim, Yang Song, Ali Eichenbaum, Thomas Brovelli, Sahitya Potluri, Preethi Lahoti, Cip Baetu, Ali Ghorbani, Charles Chen, Andy Crawford, Shalini Pal, Mukund Sridhar, Petru Gurita, Asier Mujika, Igor Petrovski, Pierre-Louis Cedoz, Chenmei Li, Shiyuan Chen, Niccolò~Dal Santo, Siddharth Goyal, Jitesh Punjabi, Karthik Kappaganthu, Chester Kwak, Pallavi LV, Sarmishta Velury, Himadri Choudhury, Jamie Hall, Premal Shah, Ricardo Figueira, Matt Thomas, Minjie Lu, Ting Zhou, Chintu Kumar, Thomas Jurdi, Sharat Chikkerur, Yenai Ma, Adams Yu, Soo Kwak, Victor Ähdel, Sujeevan Rajayogam, Travis Choma, Fei Liu, Aditya Barua, Colin Ji, Ji~Ho Park, Vincent Hellendoorn, Alex Bailey, Taylan Bilal, Huanjie Zhou,
  Mehrdad Khatir, Charles Sutton, Wojciech Rzadkowski, Fiona Macintosh, Konstantin Shagin, Paul Medina, Chen Liang, Jinjing Zhou, Pararth Shah, Yingying Bi, Attila Dankovics, Shipra Banga, Sabine Lehmann, Marissa Bredesen, Zifan Lin, John~Eric Hoffmann, Jonathan Lai, Raynald Chung, Kai Yang, Nihal Balani, Arthur Bražinskas, Andrei Sozanschi, Matthew Hayes, Héctor~Fernández Alcalde, Peter Makarov, Will Chen, Antonio Stella, Liselotte Snijders, Michael Mandl, Ante Kärrman, Paweł Nowak, Xinyi Wu, Alex Dyck, Krishnan Vaidyanathan, Raghavender R, Jessica Mallet, Mitch Rudominer, Eric Johnston, Sushil Mittal, Akhil Udathu, Janara Christensen, Vishal Verma, Zach Irving, Andreas Santucci, Gamaleldin Elsayed, Elnaz Davoodi, Marin Georgiev, Ian Tenney, Nan Hua, Geoffrey Cideron, Edouard Leurent, Mahmoud Alnahlawi, Ionut Georgescu, Nan Wei, Ivy Zheng, Dylan Scandinaro, Heinrich Jiang, Jasper Snoek, Mukund Sundararajan, Xuezhi Wang, Zack Ontiveros, Itay Karo, Jeremy Cole, Vinu Rajashekhar, Lara Tumeh, Eyal
  Ben-David, Rishub Jain, Jonathan Uesato, Romina Datta, Oskar Bunyan, Shimu Wu, John Zhang, Piotr Stanczyk, Ye~Zhang, David Steiner, Subhajit Naskar, Michael Azzam, Matthew Johnson, Adam Paszke, Chung-Cheng Chiu, Jaume~Sanchez Elias, Afroz Mohiuddin, Faizan Muhammad, Jin Miao, Andrew Lee, Nino Vieillard, Jane Park, Jiageng Zhang, Jeff Stanway, Drew Garmon, Abhijit Karmarkar, Zhe Dong, Jong Lee, Aviral Kumar, Luowei Zhou, Jonathan Evens, William Isaac, Geoffrey Irving, Edward Loper, Michael Fink, Isha Arkatkar, Nanxin Chen, Izhak Shafran, Ivan Petrychenko, Zhe Chen, Johnson Jia, Anselm Levskaya, Zhenkai Zhu, Peter Grabowski, Yu~Mao, Alberto Magni, Kaisheng Yao, Javier Snaider, Norman Casagrande, Evan Palmer, Paul Suganthan, Alfonso Castaño, Irene Giannoumis, Wooyeol Kim, Mikołaj Rybiński, Ashwin Sreevatsa, Jennifer Prendki, David Soergel, Adrian Goedeckemeyer, Willi Gierke, Mohsen Jafari, Meenu Gaba, Jeremy Wiesner, Diana~Gage Wright, Yawen Wei, Harsha Vashisht, Yana Kulizhskaya, Jay Hoover, Maigo Le,
  Lu~Li, Chimezie Iwuanyanwu, Lu~Liu, Kevin Ramirez, Andrey Khorlin, Albert Cui, Tian LIN, Marcus Wu, Ricardo Aguilar, Keith Pallo, Abhishek Chakladar, Ginger Perng, Elena~Allica Abellan, Mingyang Zhang, Ishita Dasgupta, Nate Kushman, Ivo Penchev, Alena Repina, Xihui Wu, Tom van~der Weide, Priya Ponnapalli, Caroline Kaplan, Jiri Simsa, Shuangfeng Li, Olivier Dousse, Fan Yang, Jeff Piper, Nathan Ie, Rama Pasumarthi, Nathan Lintz, Anitha Vijayakumar, Daniel Andor, Pedro Valenzuela, Minnie Lui, Cosmin Paduraru, Daiyi Peng, Katherine Lee, Shuyuan Zhang, Somer Greene, Duc~Dung Nguyen, Paula Kurylowicz, Cassidy Hardin, Lucas Dixon, Lili Janzer, Kiam Choo, Ziqiang Feng, Biao Zhang, Achintya Singhal, Dayou Du, Dan McKinnon, Natasha Antropova, Tolga Bolukbasi, Orgad Keller, David Reid, Daniel Finchelstein, Maria~Abi Raad, Remi Crocker, Peter Hawkins, Robert Dadashi, Colin Gaffney, Ken Franko, Anna Bulanova, Rémi Leblond, Shirley Chung, Harry Askham, Luis~C. Cobo, Kelvin Xu, Felix Fischer, Jun Xu, Christina Sorokin,
  Chris Alberti, Chu-Cheng Lin, Colin Evans, Alek Dimitriev, Hannah Forbes, Dylan Banarse, Zora Tung, Mark Omernick, Colton Bishop, Rachel Sterneck, Rohan Jain, Jiawei Xia, Ehsan Amid, Francesco Piccinno, Xingyu Wang, Praseem Banzal, Daniel~J. Mankowitz, Alex Polozov, Victoria Krakovna, Sasha Brown, MohammadHossein Bateni, Dennis Duan, Vlad Firoiu, Meghana Thotakuri, Tom Natan, Matthieu Geist, Ser tan Girgin, Hui Li, Jiayu Ye, Ofir Roval, Reiko Tojo, Michael Kwong, James Lee-Thorp, Christopher Yew, Danila Sinopalnikov, Sabela Ramos, John Mellor, Abhishek Sharma, Kathy Wu, David Miller, Nicolas Sonnerat, Denis Vnukov, Rory Greig, Jennifer Beattie, Emily Caveness, Libin Bai, Julian Eisenschlos, Alex Korchemniy, Tomy Tsai, Mimi Jasarevic, Weize Kong, Phuong Dao, Zeyu Zheng, Frederick Liu, Fan Yang, Rui Zhu, Tian~Huey Teh, Jason Sanmiya, Evgeny Gladchenko, Nejc Trdin, Daniel Toyama, Evan Rosen, Sasan Tavakkol, Linting Xue, Chen Elkind, Oliver Woodman, John Carpenter, George Papamakarios, Rupert Kemp, Sushant
  Kafle, Tanya Grunina, Rishika Sinha, Alice Talbert, Diane Wu, Denese Owusu-Afriyie, Cosmo Du, Chloe Thornton, Jordi Pont-Tuset, Pradyumna Narayana, Jing Li, Saaber Fatehi, John Wieting, Omar Ajmeri, Benigno Uria, Yeongil Ko, Laura Knight, Amélie Héliou, Ning Niu, Shane Gu, Chenxi Pang, Yeqing Li, Nir Levine, Ariel Stolovich, Rebeca Santamaria-Fernandez, Sonam Goenka, Wenny Yustalim, Robin Strudel, Ali Elqursh, Charlie Deck, Hyo Lee, Zonglin Li, Kyle Levin, Raphael Hoffmann, Dan Holtmann-Rice, Olivier Bachem, Sho Arora, Christy Koh, Soheil~Hassas Yeganeh, Siim Põder, Mukarram Tariq, Yanhua Sun, Lucian Ionita, Mojtaba Seyedhosseini, Pouya Tafti, Zhiyu Liu, Anmol Gulati, Jasmine Liu, Xinyu Ye, Bart Chrzaszcz, Lily Wang, Nikhil Sethi, Tianrun Li, Ben Brown, Shreya Singh, Wei Fan, Aaron Parisi, Joe Stanton, Vinod Koverkathu, Christopher~A. Choquette-Choo, Yunjie Li, TJ~Lu, Abe Ittycheriah, Prakash Shroff, Mani Varadarajan, Sanaz Bahargam, Rob Willoughby, David Gaddy, Guillaume Desjardins, Marco Cornero, Brona
  Robenek, Bhavishya Mittal, Ben Albrecht, Ashish Shenoy, Fedor Moiseev, Henrik Jacobsson, Alireza Ghaffarkhah, Morgane Rivière, Alanna Walton, Clément Crepy, Alicia Parrish, Zongwei Zhou, Clement Farabet, Carey Radebaugh, Praveen Srinivasan, Claudia van~der Salm, Andreas Fidjeland, Salvatore Scellato, Eri Latorre-Chimoto, Hanna Klimczak-Plucińska, David Bridson, Dario de~Cesare, Tom Hudson, Piermaria Mendolicchio, Lexi Walker, Alex Morris, Matthew Mauger, Alexey Guseynov, Alison Reid, Seth Odoom, Lucia Loher, Victor Cotruta, Madhavi Yenugula, Dominik Grewe, Anastasia Petrushkina, Tom Duerig, Antonio Sanchez, Steve Yadlowsky, Amy Shen, Amir Globerson, Lynette Webb, Sahil Dua, Dong Li, Surya Bhupatiraju, Dan Hurt, Haroon Qureshi, Ananth Agarwal, Tomer Shani, Matan Eyal, Anuj Khare, Shreyas~Rammohan Belle, Lei Wang, Chetan Tekur, Mihir~Sanjay Kale, Jinliang Wei, Ruoxin Sang, Brennan Saeta, Tyler Liechty, Yi~Sun, Yao Zhao, Stephan Lee, Pandu Nayak, Doug Fritz, Manish~Reddy Vuyyuru, John Aslanides, Nidhi Vyas,
  Martin Wicke, Xiao Ma, Evgenii Eltyshev, Nina Martin, Hardie Cate, James Manyika, Keyvan Amiri, Yelin Kim, Xi~Xiong, Kai Kang, Florian Luisier, Nilesh Tripuraneni, David Madras, Mandy Guo, Austin Waters, Oliver Wang, Joshua Ainslie, Jason Baldridge, Han Zhang, Garima Pruthi, Jakob Bauer, Feng Yang, Riham Mansour, Jason Gelman, Yang Xu, George Polovets, Ji~Liu, Honglong Cai, Warren Chen, XiangHai Sheng, Emily Xue, Sherjil Ozair, Christof Angermueller, Xiaowei Li, Anoop Sinha, Weiren Wang, Julia Wiesinger, Emmanouil Koukoumidis, Yuan Tian, Anand Iyer, Madhu Gurumurthy, Mark Goldenson, Parashar Shah, MK~Blake, Hongkun Yu, Anthony Urbanowicz, Jennimaria Palomaki, Chrisantha Fernando, Ken Durden, Harsh Mehta, Nikola Momchev, Elahe Rahimtoroghi, Maria Georgaki, Amit Raul, Sebastian Ruder, Morgan Redshaw, Jinhyuk Lee, Denny Zhou, Komal Jalan, Dinghua Li, Blake Hechtman, Parker Schuh, Milad Nasr, Kieran Milan, Vladimir Mikulik, Juliana Franco, Tim Green, Nam Nguyen, Joe Kelley, Aroma Mahendru, Andrea Hu, Joshua
  Howland, Ben Vargas, Jeffrey Hui, Kshitij Bansal, Vikram Rao, Rakesh Ghiya, Emma Wang, Ke~Ye, Jean~Michel Sarr, Melanie~Moranski Preston, Madeleine Elish, Steve Li, Aakash Kaku, Jigar Gupta, Ice Pasupat, Da-Cheng Juan, Milan Someswar, Tejvi M., Xinyun Chen, Aida Amini, Alex Fabrikant, Eric Chu, Xuanyi Dong, Amruta Muthal, Senaka Buthpitiya, Sarthak Jauhari, Nan Hua, Urvashi Khandelwal, Ayal Hitron, Jie Ren, Larissa Rinaldi, Shahar Drath, Avigail Dabush, Nan-Jiang Jiang, Harshal Godhia, Uli Sachs, Anthony Chen, Yicheng Fan, Hagai Taitelbaum, Hila Noga, Zhuyun Dai, James Wang, Chen Liang, Jenny Hamer, Chun-Sung Ferng, Chenel Elkind, Aviel Atias, Paulina Lee, Vít Listík, Mathias Carlen, Jan van~de Kerkhof, Marcin Pikus, Krunoslav Zaher, Paul Müller, Sasha Zykova, Richard Stefanec, Vitaly Gatsko, Christoph Hirnschall, Ashwin Sethi, Xingyu~Federico Xu, Chetan Ahuja, Beth Tsai, Anca Stefanoiu, Bo~Feng, Keshav Dhandhania, Manish Katyal, Akshay Gupta, Atharva Parulekar, Divya Pitta, Jing Zhao, Vivaan Bhatia,
  Yashodha Bhavnani, Omar Alhadlaq, Xiaolin Li, Peter Danenberg, Dennis Tu, Alex Pine, Vera Filippova, Abhipso Ghosh, Ben Limonchik, Bhargava Urala, Chaitanya~Krishna Lanka, Derik Clive, Yi~Sun, Edward Li, Hao Wu, Kevin Hongtongsak, Ianna Li, Kalind Thakkar, Kuanysh Omarov, Kushal Majmundar, Michael Alverson, Michael Kucharski, Mohak Patel, Mudit Jain, Maksim Zabelin, Paolo Pelagatti, Rohan Kohli, Saurabh Kumar, Joseph Kim, Swetha Sankar, Vineet Shah, Lakshmi Ramachandruni, Xiangkai Zeng, Ben Bariach, Laura Weidinger, Amar Subramanya, Sissie Hsiao, Demis Hassabis, Koray Kavukcuoglu, Adam Sadovsky, Quoc Le, Trevor Strohman, Yonghui Wu, Slav Petrov, Jeffrey Dean, and Oriol Vinyals.
\newblock Gemini: A family of highly capable multimodal models, 2024.

\bibitem[Gemma-Team(2024)]{gemmareport}
Gemma-Team.
\newblock Gemma: Open models based on gemini research and technology, 2024.

\bibitem[Hasan et~al.(2021)Hasan, Bhattacharjee, Islam, Mubasshir, Li, Kang, Rahman, and Shahriyar]{hasan-etal-2021-xl}
Tahmid Hasan, Abhik Bhattacharjee, Md.~Saiful Islam, Kazi Mubasshir, Yuan-Fang Li, Yong-Bin Kang, M.~Sohel Rahman, and Rifat Shahriyar.
\newblock {XL}-sum: Large-scale multilingual abstractive summarization for 44 languages.
\newblock In \emph{Findings of the Association for Computational Linguistics: ACL-IJCNLP 2021}, pp.\  4693--4703, Online, August 2021. Association for Computational Linguistics.
\newblock URL \url{https://aclanthology.org/2021.findings-acl.413}.

\bibitem[Hendrycks et~al.(2021)Hendrycks, Burns, Basart, Zou, Mazeika, Song, and Steinhardt]{hendryckstest2021}
Dan Hendrycks, Collin Burns, Steven Basart, Andy Zou, Mantas Mazeika, Dawn Song, and Jacob Steinhardt.
\newblock Measuring massive multitask language understanding.
\newblock \emph{Proceedings of the International Conference on Learning Representations (ICLR)}, 2021.

\bibitem[Hwang et~al.(2023)Hwang, Prasad~Majumder, and Tandon]{hwang2023aligning}
EunJeong Hwang, Bodhisattwa Prasad~Majumder, and Niket Tandon.
\newblock Aligning language models to user opinions.
\newblock \emph{arXiv e-prints}, pp.\  arXiv--2305, 2023.

\bibitem[Ivison et~al.(2023)Ivison, Wang, Pyatkin, Lambert, Peters, Dasigi, Jang, Wadden, Smith, Beltagy, and Hajishirzi]{ivison2023camels}
Hamish Ivison, Yizhong Wang, Valentina Pyatkin, Nathan Lambert, Matthew Peters, Pradeep Dasigi, Joel Jang, David Wadden, Noah~A. Smith, Iz~Beltagy, and Hannaneh Hajishirzi.
\newblock Camels in a changing climate: Enhancing lm adaptation with tulu 2, 2023.

\bibitem[Jiang et~al.(2023{\natexlab{a}})Jiang, Sablayrolles, Mensch, Bamford, Chaplot, de~las Casas, Bressand, Lengyel, Lample, Saulnier, Lavaud, Lachaux, Stock, Scao, Lavril, Wang, Lacroix, and Sayed]{jiang2023mistral}
Albert~Q. Jiang, Alexandre Sablayrolles, Arthur Mensch, Chris Bamford, Devendra~Singh Chaplot, Diego de~las Casas, Florian Bressand, Gianna Lengyel, Guillaume Lample, Lucile Saulnier, Lélio~Renard Lavaud, Marie-Anne Lachaux, Pierre Stock, Teven~Le Scao, Thibaut Lavril, Thomas Wang, Timothée Lacroix, and William~El Sayed.
\newblock Mistral 7b, 2023{\natexlab{a}}.

\bibitem[Jiang et~al.(2023{\natexlab{b}})Jiang, Zhang, Cao, Kabbara, and Roy]{jiang2023personallm}
Hang Jiang, Xiajie Zhang, Xubo Cao, Jad Kabbara, and Deb Roy.
\newblock Personallm: Investigating the ability of gpt-3.5 to express personality traits and gender differences.
\newblock \emph{arXiv preprint arXiv:2305.02547}, 2023{\natexlab{b}}.

\bibitem[Khandelwal et~al.(2023)Khandelwal, Tonneau, Bean, Kirk, and Hale]{Khandelwal2023CasteistBN}
Khyati Khandelwal, Manuel Tonneau, Andrew~M. Bean, Hannah~Rose Kirk, and Scott~A. Hale.
\newblock Casteist but not racist? quantifying disparities in large language model bias between india and the west.
\newblock \emph{ArXiv}, abs/2309.08573, 2023.
\newblock URL \url{https://api.semanticscholar.org/CorpusID:262013517}.

\bibitem[Khondaker et~al.(2023)Khondaker, Waheed, Nagoudi, and Abdul-Mageed]{khondaker2023gptaraeval}
Md~Tawkat~Islam Khondaker, Abdul Waheed, El~Moatez~Billah Nagoudi, and Muhammad Abdul-Mageed.
\newblock Gptaraeval: A comprehensive evaluation of chatgpt on arabic nlp.
\newblock \emph{arXiv}, abs/2305.14976, 2023.

\bibitem[Kingma \& Ba(2014)Kingma and Ba]{kingma2014adam}
Diederik~P Kingma and Jimmy Ba.
\newblock Adam: A method for stochastic optimization.
\newblock \emph{arXiv preprint arXiv:1412.6980}, 2014.

\bibitem[Kirk et~al.(2024{\natexlab{a}})Kirk, Mediratta, Nalmpantis, Luketina, Hambro, Grefenstette, and Raileanu]{kirk2024understanding}
Robert Kirk, Ishita Mediratta, Christoforos Nalmpantis, Jelena Luketina, Eric Hambro, Edward Grefenstette, and Roberta Raileanu.
\newblock Understanding the effects of {RLHF} on {LLM} generalisation and diversity.
\newblock In \emph{The Twelfth International Conference on Learning Representations}, 2024{\natexlab{a}}.
\newblock URL \url{https://openreview.net/forum?id=PXD3FAVHJT}.

\bibitem[Kirk et~al.(2024{\natexlab{b}})Kirk, Mediratta, Nalmpantis, Luketina, Hambro, Grefenstette, and Raileanu]{kirk2024understandingRLHF}
Robert Kirk, Ishita Mediratta, Christoforos Nalmpantis, Jelena Luketina, Eric Hambro, Edward Grefenstette, and Roberta Raileanu.
\newblock Understanding the effects of rlhf on llm generalisation and diversity, 2024{\natexlab{b}}.

\bibitem[Kool et~al.(2019)Kool, van Hoof, and Welling]{Kool2019Buy4R}
Wouter Kool, Herke van Hoof, and Max Welling.
\newblock Buy 4 reinforce samples, get a baseline for free!
\newblock In \emph{DeepRLStructPred@ICLR}, 2019.
\newblock URL \url{https://api.semanticscholar.org/CorpusID:198489118}.

\bibitem[Kotek et~al.(2023)Kotek, Dockum, and Sun]{Kotek2023GenderBA}
Hadas Kotek, Rikker Dockum, and David~Q. Sun.
\newblock Gender bias and stereotypes in large language models.
\newblock \emph{Proceedings of The ACM Collective Intelligence Conference}, 2023.
\newblock URL \url{https://api.semanticscholar.org/CorpusID:261276445}.

\bibitem[Kreutzer et~al.(2022)Kreutzer, Caswell, Wang, Wahab, van Esch, Ulzii-Orshikh, Tapo, Subramani, Sokolov, Sikasote, Setyawan, Sarin, Samb, Sagot, Rivera, Rios, Papadimitriou, Osei, Suarez, Orife, Ogueji, Rubungo, Nguyen, M{\"u}ller, M{\"u}ller, Muhammad, Muhammad, Mnyakeni, Mirzakhalov, Matangira, Leong, Lawson, Kudugunta, Jernite, Jenny, Firat, Dossou, Dlamini, de~Silva, {\c{C}}abuk~Ball{\i}, Biderman, Battisti, Baruwa, Bapna, Baljekar, Azime, Awokoya, Ataman, Ahia, Ahia, Agrawal, and Adeyemi]{kreutzer-etal-2022-quality}
Julia Kreutzer, Isaac Caswell, Lisa Wang, Ahsan Wahab, Daan van Esch, Nasanbayar Ulzii-Orshikh, Allahsera Tapo, Nishant Subramani, Artem Sokolov, Claytone Sikasote, Monang Setyawan, Supheakmungkol Sarin, Sokhar Samb, Beno{\^\i}t Sagot, Clara Rivera, Annette Rios, Isabel Papadimitriou, Salomey Osei, Pedro~Ortiz Suarez, Iroro Orife, Kelechi Ogueji, Andre~Niyongabo Rubungo, Toan~Q. Nguyen, Mathias M{\"u}ller, Andr{\'e} M{\"u}ller, Shamsuddeen~Hassan Muhammad, Nanda Muhammad, Ayanda Mnyakeni, Jamshidbek Mirzakhalov, Tapiwanashe Matangira, Colin Leong, Nze Lawson, Sneha Kudugunta, Yacine Jernite, Mathias Jenny, Orhan Firat, Bonaventure F.~P. Dossou, Sakhile Dlamini, Nisansa de~Silva, Sakine {\c{C}}abuk~Ball{\i}, Stella Biderman, Alessia Battisti, Ahmed Baruwa, Ankur Bapna, Pallavi Baljekar, Israel~Abebe Azime, Ayodele Awokoya, Duygu Ataman, Orevaoghene Ahia, Oghenefego Ahia, Sweta Agrawal, and Mofetoluwa Adeyemi.
\newblock Quality at a glance: An audit of web-crawled multilingual datasets.
\newblock \emph{Transactions of the Association for Computational Linguistics}, 10:\penalty0 50--72, 2022.
\newblock \doi{10.1162/tacl_a_00447}.
\newblock URL \url{https://aclanthology.org/2022.tacl-1.4}.

\bibitem[Lahoti et~al.(2023)Lahoti, Blumm, Ma, Kotikalapudi, Potluri, Tan, Srinivasan, Packer, Beirami, Beutel, and Chen]{lahoti2023improving}
Preethi Lahoti, Nicholas Blumm, Xiao Ma, Raghavendra Kotikalapudi, Sahitya Potluri, Qijun Tan, Hansa Srinivasan, Ben Packer, Ahmad Beirami, Alex Beutel, and Jilin Chen.
\newblock Improving diversity of demographic representation in large language models via collective-critiques and self-voting.
\newblock \emph{arXiv}, abs/2310.16523, 2023.

\bibitem[Lai et~al.(2023)Lai, Nguyen, Ngo, Nguyen, Dernoncourt, Rossi, and Nguyen]{lai-etal-2023-okapi}
Viet Lai, Chien Nguyen, Nghia Ngo, Thuat Nguyen, Franck Dernoncourt, Ryan Rossi, and Thien Nguyen.
\newblock Okapi: Instruction-tuned large language models in multiple languages with reinforcement learning from human feedback.
\newblock In Yansong Feng and Els Lefever (eds.), \emph{Proceedings of the 2023 Conference on Empirical Methods in Natural Language Processing: System Demonstrations}, pp.\  318--327, Singapore, December 2023. Association for Computational Linguistics.
\newblock \doi{10.18653/v1/2023.emnlp-demo.28}.
\newblock URL \url{https://aclanthology.org/2023.emnlp-demo.28}.

\bibitem[Lai et~al.(2024)Lai, Mesgar, and Fraser]{lai2024llms}
Wen Lai, Mohsen Mesgar, and Alexander Fraser.
\newblock Llms beyond english: Scaling the multilingual capability of llms with cross-lingual feedback, 2024.

\bibitem[Lambert et~al.(2024)Lambert, Pyatkin, Morrison, Miranda, Lin, Chandu, Dziri, Kumar, Zick, Choi, Smith, and Hajishirzi]{lambert2024rewardbench}
Nathan Lambert, Valentina Pyatkin, Jacob Morrison, LJ~Miranda, Bill~Yuchen Lin, Khyathi Chandu, Nouha Dziri, Sachin Kumar, Tom Zick, Yejin Choi, Noah~A. Smith, and Hannaneh Hajishirzi.
\newblock Rewardbench: Evaluating reward models for language modeling, 2024.

\bibitem[Lent et~al.(2022)Lent, Ogueji, de~Lhoneux, Ahia, and S{\o}gaard]{lent-etal-2022-creole}
Heather Lent, Kelechi Ogueji, Miryam de~Lhoneux, Orevaoghene Ahia, and Anders S{\o}gaard.
\newblock What a creole wants, what a creole needs.
\newblock In Nicoletta Calzolari, Fr{\'e}d{\'e}ric B{\'e}chet, Philippe Blache, Khalid Choukri, Christopher Cieri, Thierry Declerck, Sara Goggi, Hitoshi Isahara, Bente Maegaard, Joseph Mariani, H{\'e}l{\`e}ne Mazo, Jan Odijk, and Stelios Piperidis (eds.), \emph{Proceedings of the Thirteenth Language Resources and Evaluation Conference}, pp.\  6439--6449, Marseille, France, June 2022. European Language Resources Association.
\newblock URL \url{https://aclanthology.org/2022.lrec-1.691}.

\bibitem[Li et~al.(2023{\natexlab{a}})Li, Chen, Luo, Kang, Zhang, Hu, Chan, and Song]{Li2023PrivacyIL}
Haoran Li, Yulin Chen, Jinglong Luo, Yan Kang, Xiaojin Zhang, Qi~Hu, Chunkit Chan, and Yangqiu Song.
\newblock Privacy in large language models: Attacks, defenses and future directions.
\newblock \emph{ArXiv}, abs/2310.10383, 2023{\natexlab{a}}.
\newblock URL \url{https://api.semanticscholar.org/CorpusID:264145758}.

\bibitem[Li et~al.(2023{\natexlab{b}})Li, Lin, Zhang, Fu, Chen, Lou, and Chen]{li2023making}
Yifei Li, Zeqi Lin, Shizhuo Zhang, Qiang Fu, Bei Chen, Jian-Guang Lou, and Weizhu Chen.
\newblock Making large language models better reasoners with step-aware verifier.
\newblock \emph{arXiv}, abs/2206.02336, 2023{\natexlab{b}}.

\bibitem[Lightman et~al.(2024)Lightman, Kosaraju, Burda, Edwards, Baker, Lee, Leike, Schulman, Sutskever, and Cobbe]{lightman2024lets}
Hunter Lightman, Vineet Kosaraju, Yuri Burda, Harrison Edwards, Bowen Baker, Teddy Lee, Jan Leike, John Schulman, Ilya Sutskever, and Karl Cobbe.
\newblock Let's verify step by step.
\newblock In \emph{The Twelfth International Conference on Learning Representations}, 2024.
\newblock URL \url{https://openreview.net/forum?id=v8L0pN6EOi}.

\bibitem[Lin et~al.(2022)Lin, Mihaylov, Artetxe, Wang, Chen, Simig, Ott, Goyal, Bhosale, Du, Pasunuru, Shleifer, Koura, Chaudhary, O{'}Horo, Wang, Zettlemoyer, Kozareva, Diab, Stoyanov, and Li]{lin-etal-2022-shot}
Xi~Victoria Lin, Todor Mihaylov, Mikel Artetxe, Tianlu Wang, Shuohui Chen, Daniel Simig, Myle Ott, Naman Goyal, Shruti Bhosale, Jingfei Du, Ramakanth Pasunuru, Sam Shleifer, Punit~Singh Koura, Vishrav Chaudhary, Brian O{'}Horo, Jeff Wang, Luke Zettlemoyer, Zornitsa Kozareva, Mona Diab, Veselin Stoyanov, and Xian Li.
\newblock Few-shot learning with multilingual generative language models.
\newblock In Yoav Goldberg, Zornitsa Kozareva, and Yue Zhang (eds.), \emph{Proceedings of the 2022 Conference on Empirical Methods in Natural Language Processing}, pp.\  9019--9052, Abu Dhabi, United Arab Emirates, December 2022. Association for Computational Linguistics.
\newblock \doi{10.18653/v1/2022.emnlp-main.616}.
\newblock URL \url{https://aclanthology.org/2022.emnlp-main.616}.

\bibitem[Longpre et~al.(2023)Longpre, Mahari, Chen, Obeng-Marnu, Sileo, Brannon, Muennighoff, Khazam, Kabbara, Perisetla, Wu, Shippole, Bollacker, Wu, Villa, Pentland, and Hooker]{longpre2023data}
Shayne Longpre, Robert Mahari, Anthony Chen, Naana Obeng-Marnu, Damien Sileo, William Brannon, Niklas Muennighoff, Nathan Khazam, Jad Kabbara, Kartik Perisetla, Xinyi Wu, Enrico Shippole, Kurt Bollacker, Tongshuang Wu, Luis Villa, Sandy Pentland, and Sara Hooker.
\newblock The data provenance initiative: A large scale audit of dataset licensing \& attribution in ai, 2023.

\bibitem[Luccioni \& Viviano(2021)Luccioni and Viviano]{luccioni-viviano-2021-whats}
Alexandra Luccioni and Joseph Viviano.
\newblock What{'}s in the box? an analysis of undesirable content in the {C}ommon {C}rawl corpus.
\newblock In Chengqing Zong, Fei Xia, Wenjie Li, and Roberto Navigli (eds.), \emph{Proceedings of the 59th Annual Meeting of the Association for Computational Linguistics and the 11th International Joint Conference on Natural Language Processing (Volume 2: Short Papers)}, pp.\  182--189, Online, August 2021. Association for Computational Linguistics.
\newblock \doi{10.18653/v1/2021.acl-short.24}.
\newblock URL \url{https://aclanthology.org/2021.acl-short.24}.

\bibitem[Lukas et~al.(2023)Lukas, Salem, Sim, Tople, Wutschitz, and Zanella-B'eguelin]{Lukas2023AnalyzingLO}
Nils Lukas, A.~Salem, Robert Sim, Shruti Tople, Lukas Wutschitz, and Santiago Zanella-B'eguelin.
\newblock Analyzing leakage of personally identifiable information in language models.
\newblock \emph{2023 IEEE Symposium on Security and Privacy (SP)}, pp.\  346--363, 2023.
\newblock URL \url{https://api.semanticscholar.org/CorpusID:256459554}.

\bibitem[Muennighoff et~al.(2023{\natexlab{a}})Muennighoff, Wang, Sutawika, Roberts, Biderman, Le~Scao, Bari, Shen, Yong, Schoelkopf, et~al.]{muennighoff2022crosslingual}
Niklas Muennighoff, Thomas Wang, Lintang Sutawika, Adam Roberts, Stella Biderman, Teven Le~Scao, M~Saiful Bari, Sheng Shen, Zheng~Xin Yong, Hailey Schoelkopf, et~al.
\newblock Crosslingual generalization through multitask finetuning.
\newblock In \emph{The 61st Annual Meeting Of The Association For Computational Linguistics}, 2023{\natexlab{a}}.

\bibitem[Muennighoff et~al.(2023{\natexlab{b}})Muennighoff, Wang, Sutawika, Roberts, Biderman, Le~Scao, Bari, Shen, Yong, Schoelkopf, et~al.]{muennighoff2023crosslingual}
Niklas Muennighoff, Thomas Wang, Lintang Sutawika, Adam Roberts, Stella Biderman, Teven Le~Scao, M~Saiful Bari, Sheng Shen, Zheng~Xin Yong, Hailey Schoelkopf, et~al.
\newblock Crosslingual generalization through multitask finetuning.
\newblock In \emph{The 61st Annual Meeting Of The Association For Computational Linguistics}, 2023{\natexlab{b}}.

\bibitem[Naik et~al.(2023)Naik, Chandrasekaran, Yuksekgonul, Palangi, and Nushi]{naik2023diversity}
Ranjita Naik, Varun Chandrasekaran, Mert Yuksekgonul, Hamid Palangi, and Besmira Nushi.
\newblock Diversity of thought improves reasoning abilities of large language models.
\newblock \emph{arXiv}, abs/2310.07088, 2023.

\bibitem[Nakano et~al.(2021)Nakano, Hilton, Balaji, Wu, Ouyang, Kim, Hesse, Jain, Kosaraju, Saunders, et~al.]{nakano2021webgpt}
Reiichiro Nakano, Jacob Hilton, Suchir Balaji, Jeff Wu, Long Ouyang, Christina Kim, Christopher Hesse, Shantanu Jain, Vineet Kosaraju, William Saunders, et~al.
\newblock Webgpt: Browser-assisted question-answering with human feedback.
\newblock \emph{arXiv preprint arXiv:2112.09332}, 2021.

\bibitem[Nasr et~al.(2023)Nasr, Carlini, Hayase, Jagielski, Cooper, Ippolito, Choquette-Choo, Wallace, Tramèr, and Lee]{nasr2023scalable}
Milad Nasr, Nicholas Carlini, Jonathan Hayase, Matthew Jagielski, A.~Feder Cooper, Daphne Ippolito, Christopher~A. Choquette-Choo, Eric Wallace, Florian Tramèr, and Katherine Lee.
\newblock Scalable extraction of training data from (production) language models.
\newblock \emph{arXiv}, abs/2311.17035, 2023.

\bibitem[{NLLB Team} et~al.(2022){NLLB Team}, Costa-jussà, Cross, Çelebi, Elbayad, Heafield, Heffernan, Kalbassi, Lam, Licht, Maillard, Sun, Wang, Wenzek, Youngblood, Akula, Barrault, Mejia-Gonzalez, Hansanti, Hoffman, Jarrett, Sadagopan, Rowe, Spruit, Tran, Andrews, Ayan, Bhosale, Edunov, Fan, Gao, Goswami, Guzmán, Koehn, Mourachko, Ropers, Saleem, Schwenk, and Wang]{nllb2022}
{NLLB Team}, Marta~R. Costa-jussà, James Cross, Onur Çelebi, Maha Elbayad, Kenneth Heafield, Kevin Heffernan, Elahe Kalbassi, Janice Lam, Daniel Licht, Jean Maillard, Anna Sun, Skyler Wang, Guillaume Wenzek, Al~Youngblood, Bapi Akula, Loic Barrault, Gabriel Mejia-Gonzalez, Prangthip Hansanti, John Hoffman, Semarley Jarrett, Kaushik~Ram Sadagopan, Dirk Rowe, Shannon Spruit, Chau Tran, Pierre Andrews, Necip~Fazil Ayan, Shruti Bhosale, Sergey Edunov, Angela Fan, Cynthia Gao, Vedanuj Goswami, Francisco Guzmán, Philipp Koehn, Alexandre Mourachko, Christophe Ropers, Safiyyah Saleem, Holger Schwenk, and Jeff Wang.
\newblock No language left behind: Scaling human-centered machine translation.
\newblock 2022.

\bibitem[OpenAI et~al.(2023)OpenAI, :, Achiam, Adler, Agarwal, Ahmad, Akkaya, Aleman, Almeida, Altenschmidt, Altman, Anadkat, Avila, Babuschkin, Balaji, Balcom, Baltescu, Bao, Bavarian, Belgum, Bello, Berdine, Bernadett-Shapiro, Berner, Bogdonoff, Boiko, Boyd, Brakman, Brockman, Brooks, Brundage, Button, Cai, Campbell, Cann, Carey, Carlson, Carmichael, Chan, Chang, Chantzis, Chen, Chen, Chen, Chen, Chen, Chess, Cho, Chu, Chung, Cummings, Currier, Dai, Decareaux, Degry, Deutsch, Deville, Dhar, Dohan, Dowling, Dunning, Ecoffet, Eleti, Eloundou, Farhi, Fedus, Felix, Fishman, Forte, Fulford, Gao, Georges, Gibson, Goel, Gogineni, Goh, Gontijo-Lopes, Gordon, Grafstein, Gray, Greene, Gross, Gu, Guo, Hallacy, Han, Harris, He, Heaton, Heidecke, Hesse, Hickey, Hickey, Hoeschele, Houghton, Hsu, Hu, Hu, Huizinga, Jain, Jain, Jang, Jiang, Jiang, Jin, Jin, Jomoto, Jonn, Jun, Kaftan, Łukasz Kaiser, Kamali, Kanitscheider, Keskar, Khan, Kilpatrick, Kim, Kim, Kim, Kirchner, Kiros, Knight, Kokotajlo, Łukasz Kondraciuk,
  Kondrich, Konstantinidis, Kosic, Krueger, Kuo, Lampe, Lan, Lee, Leike, Leung, Levy, Li, Lim, Lin, Lin, Litwin, Lopez, Lowe, Lue, Makanju, Malfacini, Manning, Markov, Markovski, Martin, Mayer, Mayne, McGrew, McKinney, McLeavey, McMillan, McNeil, Medina, Mehta, Menick, Metz, Mishchenko, Mishkin, Monaco, Morikawa, Mossing, Mu, Murati, Murk, Mély, Nair, Nakano, Nayak, Neelakantan, Ngo, Noh, Ouyang, O'Keefe, Pachocki, Paino, Palermo, Pantuliano, Parascandolo, Parish, Parparita, Passos, Pavlov, Peng, Perelman, de~Avila Belbute~Peres, Petrov, de~Oliveira~Pinto, Michael, Pokorny, Pokrass, Pong, Powell, Power, Power, Proehl, Puri, Radford, Rae, Ramesh, Raymond, Real, Rimbach, Ross, Rotsted, Roussez, Ryder, Saltarelli, Sanders, Santurkar, Sastry, Schmidt, Schnurr, Schulman, Selsam, Sheppard, Sherbakov, Shieh, Shoker, Shyam, Sidor, Sigler, Simens, Sitkin, Slama, Sohl, Sokolowsky, Song, Staudacher, Such, Summers, Sutskever, Tang, Tezak, Thompson, Tillet, Tootoonchian, Tseng, Tuggle, Turley, Tworek, Uribe, Vallone,
  Vijayvergiya, Voss, Wainwright, Wang, Wang, Wang, Ward, Wei, Weinmann, Welihinda, Welinder, Weng, Weng, Wiethoff, Willner, Winter, Wolrich, Wong, Workman, Wu, Wu, Wu, Xiao, Xu, Yoo, Yu, Yuan, Zaremba, Zellers, Zhang, Zhang, Zhao, Zheng, Zhuang, Zhuk, and Zoph]{openai2023GPT4}
OpenAI, :, Josh Achiam, Steven Adler, Sandhini Agarwal, Lama Ahmad, Ilge Akkaya, Florencia~Leoni Aleman, Diogo Almeida, Janko Altenschmidt, Sam Altman, Shyamal Anadkat, Red Avila, Igor Babuschkin, Suchir Balaji, Valerie Balcom, Paul Baltescu, Haiming Bao, Mo~Bavarian, Jeff Belgum, Irwan Bello, Jake Berdine, Gabriel Bernadett-Shapiro, Christopher Berner, Lenny Bogdonoff, Oleg Boiko, Madelaine Boyd, Anna-Luisa Brakman, Greg Brockman, Tim Brooks, Miles Brundage, Kevin Button, Trevor Cai, Rosie Campbell, Andrew Cann, Brittany Carey, Chelsea Carlson, Rory Carmichael, Brooke Chan, Che Chang, Fotis Chantzis, Derek Chen, Sully Chen, Ruby Chen, Jason Chen, Mark Chen, Ben Chess, Chester Cho, Casey Chu, Hyung~Won Chung, Dave Cummings, Jeremiah Currier, Yunxing Dai, Cory Decareaux, Thomas Degry, Noah Deutsch, Damien Deville, Arka Dhar, David Dohan, Steve Dowling, Sheila Dunning, Adrien Ecoffet, Atty Eleti, Tyna Eloundou, David Farhi, Liam Fedus, Niko Felix, Simón~Posada Fishman, Juston Forte, Isabella Fulford, Leo Gao,
  Elie Georges, Christian Gibson, Vik Goel, Tarun Gogineni, Gabriel Goh, Rapha Gontijo-Lopes, Jonathan Gordon, Morgan Grafstein, Scott Gray, Ryan Greene, Joshua Gross, Shixiang~Shane Gu, Yufei Guo, Chris Hallacy, Jesse Han, Jeff Harris, Yuchen He, Mike Heaton, Johannes Heidecke, Chris Hesse, Alan Hickey, Wade Hickey, Peter Hoeschele, Brandon Houghton, Kenny Hsu, Shengli Hu, Xin Hu, Joost Huizinga, Shantanu Jain, Shawn Jain, Joanne Jang, Angela Jiang, Roger Jiang, Haozhun Jin, Denny Jin, Shino Jomoto, Billie Jonn, Heewoo Jun, Tomer Kaftan, Łukasz Kaiser, Ali Kamali, Ingmar Kanitscheider, Nitish~Shirish Keskar, Tabarak Khan, Logan Kilpatrick, Jong~Wook Kim, Christina Kim, Yongjik Kim, Hendrik Kirchner, Jamie Kiros, Matt Knight, Daniel Kokotajlo, Łukasz Kondraciuk, Andrew Kondrich, Aris Konstantinidis, Kyle Kosic, Gretchen Krueger, Vishal Kuo, Michael Lampe, Ikai Lan, Teddy Lee, Jan Leike, Jade Leung, Daniel Levy, Chak~Ming Li, Rachel Lim, Molly Lin, Stephanie Lin, Mateusz Litwin, Theresa Lopez, Ryan Lowe,
  Patricia Lue, Anna Makanju, Kim Malfacini, Sam Manning, Todor Markov, Yaniv Markovski, Bianca Martin, Katie Mayer, Andrew Mayne, Bob McGrew, Scott~Mayer McKinney, Christine McLeavey, Paul McMillan, Jake McNeil, David Medina, Aalok Mehta, Jacob Menick, Luke Metz, Andrey Mishchenko, Pamela Mishkin, Vinnie Monaco, Evan Morikawa, Daniel Mossing, Tong Mu, Mira Murati, Oleg Murk, David Mély, Ashvin Nair, Reiichiro Nakano, Rajeev Nayak, Arvind Neelakantan, Richard Ngo, Hyeonwoo Noh, Long Ouyang, Cullen O'Keefe, Jakub Pachocki, Alex Paino, Joe Palermo, Ashley Pantuliano, Giambattista Parascandolo, Joel Parish, Emy Parparita, Alex Passos, Mikhail Pavlov, Andrew Peng, Adam Perelman, Filipe de~Avila Belbute~Peres, Michael Petrov, Henrique~Ponde de~Oliveira~Pinto, Michael, Pokorny, Michelle Pokrass, Vitchyr Pong, Tolly Powell, Alethea Power, Boris Power, Elizabeth Proehl, Raul Puri, Alec Radford, Jack Rae, Aditya Ramesh, Cameron Raymond, Francis Real, Kendra Rimbach, Carl Ross, Bob Rotsted, Henri Roussez, Nick Ryder,
  Mario Saltarelli, Ted Sanders, Shibani Santurkar, Girish Sastry, Heather Schmidt, David Schnurr, John Schulman, Daniel Selsam, Kyla Sheppard, Toki Sherbakov, Jessica Shieh, Sarah Shoker, Pranav Shyam, Szymon Sidor, Eric Sigler, Maddie Simens, Jordan Sitkin, Katarina Slama, Ian Sohl, Benjamin Sokolowsky, Yang Song, Natalie Staudacher, Felipe~Petroski Such, Natalie Summers, Ilya Sutskever, Jie Tang, Nikolas Tezak, Madeleine Thompson, Phil Tillet, Amin Tootoonchian, Elizabeth Tseng, Preston Tuggle, Nick Turley, Jerry Tworek, Juan Felipe~Cerón Uribe, Andrea Vallone, Arun Vijayvergiya, Chelsea Voss, Carroll Wainwright, Justin~Jay Wang, Alvin Wang, Ben Wang, Jonathan Ward, Jason Wei, CJ~Weinmann, Akila Welihinda, Peter Welinder, Jiayi Weng, Lilian Weng, Matt Wiethoff, Dave Willner, Clemens Winter, Samuel Wolrich, Hannah Wong, Lauren Workman, Sherwin Wu, Jeff Wu, Michael Wu, Kai Xiao, Tao Xu, Sarah Yoo, Kevin Yu, Qiming Yuan, Wojciech Zaremba, Rowan Zellers, Chong Zhang, Marvin Zhang, Shengjia Zhao, Tianhao
  Zheng, Juntang Zhuang, William Zhuk, and Barret Zoph.
\newblock Gpt-4 technical report, 2023.

\bibitem[Ouyang et~al.(2022{\natexlab{a}})Ouyang, Wu, Jiang, Almeida, Wainwright, Mishkin, Zhang, Agarwal, Slama, Ray, Schulman, Hilton, Kelton, Miller, Simens, Askell, Welinder, Christiano, Leike, and Lowe]{ouyang2022LLMRLHF}
Long Ouyang, Jeff Wu, Xu~Jiang, Diogo Almeida, Carroll~L. Wainwright, Pamela Mishkin, Chong Zhang, Sandhini Agarwal, Katarina Slama, Alex Ray, John Schulman, Jacob Hilton, Fraser Kelton, Luke Miller, Maddie Simens, Amanda Askell, Peter Welinder, Paul Christiano, Jan Leike, and Ryan Lowe.
\newblock Training language models to follow instructions with human feedback, 2022{\natexlab{a}}.

\bibitem[Ouyang et~al.(2022{\natexlab{b}})Ouyang, Wu, Jiang, Almeida, Wainwright, Mishkin, Zhang, Agarwal, Slama, Ray, et~al.]{ouyang2022instructgpt}
Long Ouyang, Jeffrey Wu, Xu~Jiang, Diogo Almeida, Carroll Wainwright, Pamela Mishkin, Chong Zhang, Sandhini Agarwal, Katarina Slama, Alex Ray, et~al.
\newblock Training language models to follow instructions with human feedback.
\newblock \emph{Advances in Neural Information Processing Systems}, 35:\penalty0 27730--27744, 2022{\natexlab{b}}.

\bibitem[Pace et~al.(2024)Pace, Mallinson, Malmi, Krause, and Severyn]{pace2024westofn}
Alizée Pace, Jonathan Mallinson, Eric Malmi, Sebastian Krause, and Aliaksei Severyn.
\newblock West-of-n: Synthetic preference generation for improved reward modeling, 2024.

\bibitem[Park et~al.(2024)Park, Rafailov, Ermon, and Finn]{park2024dpolength}
Ryan Park, Rafael Rafailov, Stefano Ermon, and Chelsea Finn.
\newblock Disentangling length from quality in direct preference optimization, 2024.

\bibitem[Ponti et~al.(2020)Ponti, Glava{\v{s}}, Majewska, Liu, Vuli{\'c}, and Korhonen]{ponti2020xcopa}
Edoardo~Maria Ponti, Goran Glava{\v{s}}, Olga Majewska, Qianchu Liu, Ivan Vuli{\'c}, and Anna Korhonen.
\newblock Xcopa: A multilingual dataset for causal commonsense reasoning.
\newblock pp.\  2362--2376, November 2020.
\newblock \doi{10.18653/v1/2020.emnlp-main.185}.
\newblock URL \url{https://aclanthology.org/2020.emnlp-main.185}.

\bibitem[Rafailov et~al.(2023)Rafailov, Sharma, Mitchell, Ermon, Manning, and Finn]{rafailov2023DPO}
Rafael Rafailov, Archit Sharma, Eric Mitchell, Stefano Ermon, Christopher~D. Manning, and Chelsea Finn.
\newblock Direct preference optimization: Your language model is secretly a reward model, 2023.

\bibitem[Rajbhandari et~al.(2020)Rajbhandari, Rasley, Ruwase, and He]{rajbhandari2020zero}
Samyam Rajbhandari, Jeff Rasley, Olatunji Ruwase, and Yuxiong He.
\newblock Zero: Memory optimizations toward training trillion parameter models, 2020.

\bibitem[Ranaldi \& Pucci(2023)Ranaldi and Pucci]{ranaldi-pucci-2023-english}
Leonardo Ranaldi and Giulia Pucci.
\newblock Does the {E}nglish matter? elicit cross-lingual abilities of large language models.
\newblock In Duygu Ataman (ed.), \emph{Proceedings of the 3rd Workshop on Multi-lingual Representation Learning (MRL)}, pp.\  173--183, Singapore, December 2023. Association for Computational Linguistics.
\newblock \doi{10.18653/v1/2023.mrl-1.14}.
\newblock URL \url{https://aclanthology.org/2023.mrl-1.14}.

\bibitem[Reid et~al.(2024)Reid, Savinov, Teplyashin, Lepikhin, Lillicrap, Alayrac, Soricut, Lazaridou, Firat, Schrittwieser, et~al.]{reid2024gemini}
Machel Reid, Nikolay Savinov, Denis Teplyashin, Dmitry Lepikhin, Timothy Lillicrap, Jean-baptiste Alayrac, Radu Soricut, Angeliki Lazaridou, Orhan Firat, Julian Schrittwieser, et~al.
\newblock Gemini 1.5: Unlocking multimodal understanding across millions of tokens of context.
\newblock \emph{arXiv preprint arXiv:2403.05530}, 2024.

\bibitem[Saito et~al.(2023)Saito, Wachi, Wataoka, and Akimoto]{saito2023verbosity}
Keita Saito, Akifumi Wachi, Koki Wataoka, and Youhei Akimoto.
\newblock Verbosity bias in preference labeling by large language models, 2023.

\bibitem[Schulman et~al.(2017{\natexlab{a}})Schulman, Levine, Moritz, Jordan, and Abbeel]{schulman2017trust}
John Schulman, Sergey Levine, Philipp Moritz, Michael~I. Jordan, and Pieter Abbeel.
\newblock Trust region policy optimization, 2017{\natexlab{a}}.

\bibitem[Schulman et~al.(2017{\natexlab{b}})Schulman, Wolski, Dhariwal, Radford, and Klimov]{schulman2017proximal}
John Schulman, Filip Wolski, Prafulla Dhariwal, Alec Radford, and Oleg Klimov.
\newblock Proximal policy optimization algorithms, 2017{\natexlab{b}}.

\bibitem[Schwartz et~al.(2022)Schwartz, Vassilev, Greene, Perine, Burt, Hall, et~al.]{schwartz2022towards}
Reva Schwartz, Apostol Vassilev, Kristen Greene, Lori Perine, Andrew Burt, Patrick Hall, et~al.
\newblock Towards a standard for identifying and managing bias in artificial intelligence.
\newblock \emph{NIST special publication}, 1270\penalty0 (10.6028), 2022.

\bibitem[Shi et~al.(2023)Shi, Suzgun, Freitag, Wang, Srivats, Vosoughi, Chung, Tay, Ruder, Zhou, Das, and Wei]{shi2023language-mgsm}
Freda Shi, Mirac Suzgun, Markus Freitag, Xuezhi Wang, Suraj Srivats, Soroush Vosoughi, Hyung~Won Chung, Yi~Tay, Sebastian Ruder, Denny Zhou, Dipanjan Das, and Jason Wei.
\newblock Language models are multilingual chain-of-thought reasoners.
\newblock In \emph{The Eleventh International Conference on Learning Representations}, 2023.
\newblock URL \url{https://openreview.net/forum?id=fR3wGCk-IXp}.

\bibitem[Singh et~al.(2024)Singh, Vargus, Dsouza, Karlsson, Mahendiran, Ko, Shandilya, Patel, Mataciunas, OMahony, Zhang, Hettiarachchi, Wilson, Machado, Moura, Krzemiński, Fadaei, Ergün, Okoh, Alaagib, Mudannayake, Alyafeai, Chien, Ruder, Guthikonda, Alghamdi, Gehrmann, Muennighoff, Bartolo, Kreutzer, Üstün, Fadaee, and Hooker]{ayadata2024}
Shivalika Singh, Freddie Vargus, Daniel Dsouza, Börje~F. Karlsson, Abinaya Mahendiran, Wei-Yin Ko, Herumb Shandilya, Jay Patel, Deividas Mataciunas, Laura OMahony, Mike Zhang, Ramith Hettiarachchi, Joseph Wilson, Marina Machado, Luisa~Souza Moura, Dominik Krzemiński, Hakimeh Fadaei, Irem Ergün, Ifeoma Okoh, Aisha Alaagib, Oshan Mudannayake, Zaid Alyafeai, Vu~Minh Chien, Sebastian Ruder, Surya Guthikonda, Emad~A. Alghamdi, Sebastian Gehrmann, Niklas Muennighoff, Max Bartolo, Julia Kreutzer, Ahmet Üstün, Marzieh Fadaee, and Sara Hooker.
\newblock Aya dataset: An open-access collection for multilingual instruction tuning.
\newblock \emph{arXiv preprint arXiv:2402.06619}, 2024.

\bibitem[Stiennon et~al.(2020)Stiennon, Ouyang, Wu, Ziegler, Lowe, Voss, Radford, Amodei, and Christiano]{LearningToSummarizeHF}
Nisan Stiennon, Long Ouyang, Jeff Wu, Daniel~M. Ziegler, Ryan Lowe, Chelsea Voss, Alec Radford, Dario Amodei, and Paul Christiano.
\newblock Learning to summarize from human feedback, 2020.

\bibitem[Stiennon et~al.(2022)Stiennon, Ouyang, Wu, Ziegler, Lowe, Voss, Radford, Amodei, and Christiano]{stiennon2022learningRLHF}
Nisan Stiennon, Long Ouyang, Jeff Wu, Daniel~M. Ziegler, Ryan Lowe, Chelsea Voss, Alec Radford, Dario Amodei, and Paul Christiano.
\newblock Learning to summarize from human feedback, 2022.

\bibitem[Tajwar et~al.(2024)Tajwar, Singh, Sharma, Rafailov, Schneider, Xie, Ermon, Finn, and Kumar]{tajwar2024preference}
Fahim Tajwar, Anikait Singh, Archit Sharma, Rafael Rafailov, Jeff Schneider, Tengyang Xie, Stefano Ermon, Chelsea Finn, and Aviral Kumar.
\newblock Preference fine-tuning of llms should leverage suboptimal, on-policy data, 2024.

\bibitem[Tang et~al.(2024)Tang, Guo, Zheng, Calandriello, Cao, Tarassov, Munos, Ávila Pires, Valko, Cheng, and Dabney]{tang2024understanding}
Yunhao Tang, Daniel~Zhaohan Guo, Zeyu Zheng, Daniele Calandriello, Yuan Cao, Eugene Tarassov, Rémi Munos, Bernardo Ávila Pires, Michal Valko, Yong Cheng, and Will Dabney.
\newblock Understanding the performance gap between online and offline alignment algorithms, 2024.

\bibitem[Taori et~al.(2023)Taori, Gulrajani, Zhang, Dubois, Li, Guestrin, Liang, and Hashimoto]{taori2023alpaca}
Rohan Taori, Ishaan Gulrajani, Tianyi Zhang, Yann Dubois, Xuechen Li, Carlos Guestrin, Percy Liang, and Tatsunori~B Hashimoto.
\newblock Alpaca: A strong, replicable instruction-following model.
\newblock \emph{Stanford Center for Research on Foundation Models. https://crfm. stanford. edu/2023/03/13/alpaca. html}, 3\penalty0 (6):\penalty0 7, 2023.

\bibitem[Touvron et~al.(2023)Touvron, Martin, Stone, Albert, Almahairi, Babaei, Bashlykov, Batra, Bhargava, Bhosale, Bikel, Blecher, Ferrer, Chen, Cucurull, Esiobu, Fernandes, Fu, Fu, Fuller, Gao, Goswami, Goyal, Hartshorn, Hosseini, Hou, Inan, Kardas, Kerkez, Khabsa, Kloumann, Korenev, Koura, Lachaux, Lavril, Lee, Liskovich, Lu, Mao, Martinet, Mihaylov, Mishra, Molybog, Nie, Poulton, Reizenstein, Rungta, Saladi, Schelten, Silva, Smith, Subramanian, Tan, Tang, Taylor, Williams, Kuan, Xu, Yan, Zarov, Zhang, Fan, Kambadur, Narang, Rodriguez, Stojnic, Edunov, and Scialom]{touvron2023llama2}
Hugo Touvron, Louis Martin, Kevin Stone, Peter Albert, Amjad Almahairi, Yasmine Babaei, Nikolay Bashlykov, Soumya Batra, Prajjwal Bhargava, Shruti Bhosale, Dan Bikel, Lukas Blecher, Cristian~Canton Ferrer, Moya Chen, Guillem Cucurull, David Esiobu, Jude Fernandes, Jeremy Fu, Wenyin Fu, Brian Fuller, Cynthia Gao, Vedanuj Goswami, Naman Goyal, Anthony Hartshorn, Saghar Hosseini, Rui Hou, Hakan Inan, Marcin Kardas, Viktor Kerkez, Madian Khabsa, Isabel Kloumann, Artem Korenev, Punit~Singh Koura, Marie-Anne Lachaux, Thibaut Lavril, Jenya Lee, Diana Liskovich, Yinghai Lu, Yuning Mao, Xavier Martinet, Todor Mihaylov, Pushkar Mishra, Igor Molybog, Yixin Nie, Andrew Poulton, Jeremy Reizenstein, Rashi Rungta, Kalyan Saladi, Alan Schelten, Ruan Silva, Eric~Michael Smith, Ranjan Subramanian, Xiaoqing~Ellen Tan, Binh Tang, Ross Taylor, Adina Williams, Jian~Xiang Kuan, Puxin Xu, Zheng Yan, Iliyan Zarov, Yuchen Zhang, Angela Fan, Melanie Kambadur, Sharan Narang, Aurelien Rodriguez, Robert Stojnic, Sergey Edunov, and Thomas
  Scialom.
\newblock Llama 2: Open foundation and fine-tuned chat models, 2023.

\bibitem[{\"U}st{\"u}n et~al.(2024){\"U}st{\"u}n, Aryabumi, Yong, Ko, D'souza, Onilude, Bhandari, Singh, Ooi, Kayid, Vargus, Blunsom, Longpre, Muennighoff, Fadaee, Kreutzer, and Hooker]{ustun2024aya}
Ahmet {\"U}st{\"u}n, Viraat Aryabumi, Zheng-Xin Yong, Wei-Yin Ko, Daniel D'souza, Gbemileke Onilude, Neel Bhandari, Shivalika Singh, Hui-Lee Ooi, Amr Kayid, Freddie Vargus, Phil Blunsom, Shayne Longpre, Niklas Muennighoff, Marzieh Fadaee, Julia Kreutzer, and Sara Hooker.
\newblock Aya model: An instruction finetuned open-access multilingual language model, 2024.

\bibitem[Vanmassenhove et~al.(2021)Vanmassenhove, Shterionov, and Gwilliam]{vanmassenhove-etal-2021-machine}
Eva Vanmassenhove, Dimitar Shterionov, and Matthew Gwilliam.
\newblock Machine translationese: Effects of algorithmic bias on linguistic complexity in machine translation.
\newblock In Paola Merlo, Jorg Tiedemann, and Reut Tsarfaty (eds.), \emph{Proceedings of the 16th Conference of the European Chapter of the Association for Computational Linguistics: Main Volume}, pp.\  2203--2213, Online, April 2021. Association for Computational Linguistics.
\newblock \doi{10.18653/v1/2021.eacl-main.188}.
\newblock URL \url{https://aclanthology.org/2021.eacl-main.188}.

\bibitem[Vashishtha et~al.(2023)Vashishtha, Ahuja, and Sitaram]{vashishtha2023evaluating}
Aniket Vashishtha, Kabir Ahuja, and Sunayana Sitaram.
\newblock On evaluating and mitigating gender biases in multilingual settings.
\newblock \emph{arXiv}, abs/2307.01503, 2023.

\bibitem[Verga et~al.(2024)Verga, Hofstatter, Althammer, Su, Piktus, Arkhangorodsky, Xu, White, and Lewis]{verga2024replacing}
Pat Verga, Sebastian Hofstatter, Sophia Althammer, Yixuan Su, Aleksandra Piktus, Arkady Arkhangorodsky, Minjie Xu, Naomi White, and Patrick Lewis.
\newblock Replacing judges with juries: Evaluating llm generations with a panel of diverse models, 2024.

\bibitem[Wang et~al.(2024)Wang, Li, Shao, Xu, Dai, Li, Chen, Wu, and Sui]{wang2024mathshepherd}
Peiyi Wang, Lei Li, Zhihong Shao, R.~X. Xu, Damai Dai, Yifei Li, Deli Chen, Y.~Wu, and Zhifang Sui.
\newblock Math-shepherd: Verify and reinforce llms step-by-step without human annotations, 2024.

\bibitem[Wang et~al.(2019)Wang, Dai, Póczos, and Carbonell]{wang2019characterizing}
Zirui Wang, Zihang Dai, Barnabás Póczos, and Jaime Carbonell.
\newblock Characterizing and avoiding negative transfer, 2019.

\bibitem[Wang et~al.(2020)Wang, Lipton, and Tsvetkov]{wang-etal-2020-negative}
Zirui Wang, Zachary~C. Lipton, and Yulia Tsvetkov.
\newblock On negative interference in multilingual models: Findings and a meta-learning treatment.
\newblock In Bonnie Webber, Trevor Cohn, Yulan He, and Yang Liu (eds.), \emph{Proceedings of the 2020 Conference on Empirical Methods in Natural Language Processing (EMNLP)}, pp.\  4438--4450, Online, November 2020. Association for Computational Linguistics.
\newblock \doi{10.18653/v1/2020.emnlp-main.359}.
\newblock URL \url{https://aclanthology.org/2020.emnlp-main.359}.

\bibitem[Williams(1992)]{williams1992simple}
Ronald~J. Williams.
\newblock Simple statistical gradient-following algorithms for connectionist reinforcement learning.
\newblock \emph{Machine Learning}, 8\penalty0 (3-4):\penalty0 229--256, 1992.

\bibitem[Wolfram(1997)]{wolfram1997issues}
Walt Wolfram.
\newblock Issues in dialect obsolescence: An introduction.
\newblock \emph{American speech}, 72\penalty0 (1):\penalty0 3--11, 1997.

\bibitem[Wu et~al.(2024)Wu, Balashankar, Kim, Eisenstein, and Beirami]{wu2024reuse}
Zhaofeng Wu, Ananth Balashankar, Yoon Kim, Jacob Eisenstein, and Ahmad Beirami.
\newblock Reuse your rewards: Reward model transfer for zero-shot cross-lingual alignment.
\newblock \emph{arXiv preprint arXiv:2404.12318}, 2024.

\bibitem[Xu et~al.(2024)Xu, Zhao, Zu, Gui, Zhang, and Huang]{xu2024advancing}
Nuo Xu, Jun Zhao, Can Zu, Tao Gui, Qi~Zhang, and Xuanjing Huang.
\newblock Advancing translation preference modeling with rlhf: A step towards cost-effective solution.
\newblock \emph{arXiv preprint arXiv:2402.11525}, 2024.

\bibitem[Yong et~al.(2024)Yong, Menghini, and Bach]{yong2024lowresource}
Zheng-Xin Yong, Cristina Menghini, and Stephen~H. Bach.
\newblock Low-resource languages jailbreak gpt-4, 2024.

\bibitem[Yu et~al.(2022)Yu, Sun, Zhang, and Jiang]{yu-etal-2022-translate}
Sicheng Yu, Qianru Sun, Hao Zhang, and Jing Jiang.
\newblock Translate-train embracing translationese artifacts.
\newblock In Smaranda Muresan, Preslav Nakov, and Aline Villavicencio (eds.), \emph{Proceedings of the 60th Annual Meeting of the Association for Computational Linguistics (Volume 2: Short Papers)}, pp.\  362--370, Dublin, Ireland, May 2022. Association for Computational Linguistics.
\newblock \doi{10.18653/v1/2022.acl-short.40}.
\newblock URL \url{https://aclanthology.org/2022.acl-short.40}.

\bibitem[Yuan et~al.(2024)Yuan, Pang, Cho, Li, Sukhbaatar, Xu, and Weston]{yuan2024selfrewarding}
Weizhe Yuan, Richard~Yuanzhe Pang, Kyunghyun Cho, Xian Li, Sainbayar Sukhbaatar, Jing Xu, and Jason Weston.
\newblock Self-rewarding language models, 2024.

\bibitem[Zampieri et~al.(2020)Zampieri, Nakov, and Scherrer]{zampieri2020natural}
Marcos Zampieri, Preslav Nakov, and Yves Scherrer.
\newblock Natural language processing for similar languages, varieties, and dialects: A survey.
\newblock \emph{Natural Language Engineering}, 26\penalty0 (6):\penalty0 595--612, 2020.

\bibitem[Zellers et~al.(2019)Zellers, Holtzman, Bisk, Farhadi, and Choi]{zellers-etal-2019-hellaswag}
Rowan Zellers, Ari Holtzman, Yonatan Bisk, Ali Farhadi, and Yejin Choi.
\newblock {H}ella{S}wag: Can a machine really finish your sentence?
\newblock In Anna Korhonen, David Traum, and Llu{\'\i}s M{\`a}rquez (eds.), \emph{Proceedings of the 57th Annual Meeting of the Association for Computational Linguistics}, pp.\  4791--4800, Florence, Italy, July 2019. Association for Computational Linguistics.
\newblock \doi{10.18653/v1/P19-1472}.
\newblock URL \url{https://aclanthology.org/P19-1472}.

\bibitem[Zhao et~al.(2024)Zhao, Dang, and Grover]{zhao2024group}
Siyan Zhao, John Dang, and Aditya Grover.
\newblock Group preference optimization: Few-shot alignment of large language models.
\newblock In \emph{The Twelfth International Conference on Learning Representations}, 2024.
\newblock URL \url{https://openreview.net/forum?id=DpFeMH4l8Q}.

\bibitem[Zheng et~al.(2023)Zheng, Chiang, Sheng, Zhuang, Wu, Zhuang, Lin, Li, Li, Xing, Zhang, Gonzalez, and Stoica]{zheng2023judging}
Lianmin Zheng, Wei-Lin Chiang, Ying Sheng, Siyuan Zhuang, Zhanghao Wu, Yonghao Zhuang, Zi~Lin, Zhuohan Li, Dacheng Li, Eric Xing, Hao Zhang, Joseph~E. Gonzalez, and Ion Stoica.
\newblock Judging {LLM}-as-a-judge with {MT}-bench and chatbot arena.
\newblock In \emph{Thirty-seventh Conference on Neural Information Processing Systems Datasets and Benchmarks Track}, 2023.
\newblock URL \url{https://openreview.net/forum?id=uccHPGDlao}.

\end{thebibliography}
\appendix
\clearpage
\newpage

\section{Background on DPO and RLOO}
\label{appendix:rlhf-background}

\textbf{(1) Instruction fine-tuning (SFT) stage:} A pre-trained LM is instruction-tuned using a dataset consisting of a given instruction prompt, and (typically) a human-written completion. The LM/policy is trained with a cross-entropy loss over the completion only. Often, the SFT model, denoted as $\pi^{\text{sft}}$ is used to initialize both the reward model (for online RL optimization) and the RLHF policy model. 

\textbf{(2) Preference optimization stage:} 
In this stage, the preference data such as rankings of model responses, are collected through humans or AI feedback. This data is then used to further fine-tune the SFT model (policy) to align the model with human feedback via the collected preferences data.   
Since collecting human feedback is often very expensive, many preference optimization methods train a separate reward model, on collected preference data to act as a proxy for human preferences, enabling for \emph{online} preference feedback on LLM responses without requiring human intervention. Preference optimization can be performed in a number of ways:

\textbf{Online Preference Optimization} includes training a reward model, typically through binary classification, which is then used to provide online feedback in the optimization of the policy with the following objective: 
\begin{equation*}\label{eq:rlhf}
\begin{split}
     & \max_{\pi_\theta} \mathbb{E}_{x \sim \mathcal{D}, y\sim \pi_\theta(.|x)} 
     [r_\phi(x,y) - \beta p_{KL}], \\
     & \text{with } p_{\text{KL}} = D_{\text{KL}} \pi_\theta(.|x) || \pi_{\text{ref}}(.|x)
\end{split}
\end{equation*}

where $\beta$ is meant to control the distance from the initial policy $\pi_{\text{ref}}$ during the optimization of reward $r_\theta(x,y)$ as proposed in \citep{stiennon2022learningRLHF}. The KL-penalty $p_{\text{KL}}$ is crucial as penalty-free optimization of the reward model leads to degradation in the coherence of the model. 

\textbf{Direct Preference Optimization \citep[DPO;][]{rafailov2023DPO}} collects pairwise preferences (often over LLM responses) and fine-tunes the policy to beimplicitly consistent with the collected preference pairs by using the following loss: 
\begin{equation*}
    - \log \sigma(\beta \log \frac{ \pi_\theta(y_+|x)}{\pi_{\text{ref}}(y_+|x)} - \beta \log \frac{ \pi_\theta(y_-|x)}{\pi_{\text{ref}}(y_-|x)}) 
\end{equation*}

Different from the online RL methods, DPO skips the reward modeling and enables preference optimization in an offline manner without requiring online samples from the policy during the training. At its core, DPO uses the analytical formulation of the canonical KL-controlled RLHF objective detailed in equation \ref{eq:rlhf} and the assumption that preferences can be modeled by the Bradelly-Terry model \citep{Bradley1952RankAO}, to for-go reward modeling, simplifying the problem into a supervised style classification task.

While reinforcement learning approaches share the components above, techniques differ in formulating the reward and the sample-based loss. 
In this work, we use \textbf{REINFORCE-Leave-one-out \citep[RLOO;][]{Kool2019Buy4R,ahmadian2024basics}} estimator as the online preference optimization method as 
 it is effective and more efficient than Proximal Policy Optimization (PPO) \citep{schulman2017proximal}. RLOO is a multi-sample extension of REINFORCE \citep{williams1992simple}, where 
multiple online generations are sampled from the policy per prompt which enables to reduce variance without requiring an additional network as opposed to PPO:
    \begin{equation*}
        \begin{split}
        \frac{1}{k}&\sum_{i=1}^k [R(y_{(i)},x) - \frac{1}{k-1}\sum_{j\ne{i}}R(y_{(j)},x)] \\ &\nabla \log \pi(y_{(i)}|x) \text{ for   } y_{(1)},...,y_{(k)} \overset{i.i.d}{\sim} \pi_\theta (.|x)\\        
        \end{split}
        \label{eq:RLOO}
    \end{equation*}
$\textsc{RLOO}_{k}$ considers each $y_{(i)}$ individually and uses the remaining $k-1$ samples to create an unbiased estimate of the expected return for the prompt, akin to a \emph{parameter-free} value-function, but estimated at each training step.

\section{Additional Win-Rate Results}



\begin{table*}[h]
\centering
\resizebox{0.8\textwidth}{!}{%
\begin{tabular}{llc|rrr|rrr}
\toprule
 &  & & \multicolumn{3}{c|}{\textbf{English}} &  \multicolumn{3}{c}{\textbf{Avg. 23 Langs.}}  \\
 & & \textbf{Num Examples} & 
\multicolumn{1}{c}{\textbf{Win\%}} & \multicolumn{1}{c}{\textbf{Loss\%}} & \multicolumn{1}{c|}{$\Delta$\textbf{W-L\%}} & 
\multicolumn{1}{c}{\textbf{Win\%}} & \multicolumn{1}{c}{\textbf{Loss\%}} & \multicolumn{1}{c}{$\Delta$\textbf{W-L\%}} \\ \midrule

& \textsc{EN-1} & 50K   & \multicolumn{1}{r}{37.5} & \multicolumn{1}{r}{46.5} & \multicolumn{1}{c|}{\cellcolor[HTML]{E67C73}-9.0} 

& \multicolumn{1}{r}{44.2} & \multicolumn{1}{r}{40.3} & \multicolumn{1}{c}{\cellcolor[HTML]{A5DBC0}4.0} \\

& \textsc{ML-5} & 50K & \multicolumn{1}{r}{50.5} & \multicolumn{1}{r}{40.0} & \multicolumn{1}{c|}{\cellcolor[HTML]{77C8A0}10.5} 

& \multicolumn{1}{r}{51.4} & \multicolumn{1}{r}{38.8} & \multicolumn{1}{c}{\cellcolor[HTML]{68C296}12.6} \\ 

& \textsc{ML-23} & 50K  & \multicolumn{1}{r}{51.0} & \multicolumn{1}{r}{39.5} & \multicolumn{1}{c|}{\cellcolor[HTML]{70C59C}11.5} & \multicolumn{1}{r}{50.4} & \multicolumn{1}{r}{41.3} & \multicolumn{1}{c}{\cellcolor[HTML]{81CCA7}9.1} \\ 

\multirow{-4}{*}{\textsc{RLOO vs DPO}} & \textsc{ML-23} & 230K & \multicolumn{1}{r}{50.0} & \multicolumn{1}{r}{35.0} & \multicolumn{1}{c|}{\cellcolor[HTML]{57BB8A}15.0} & \multicolumn{1}{r}{47.4} & \multicolumn{1}{r}{41.1} & \multicolumn{1}{c}{\cellcolor[HTML]{95D4B5}6.2} \\ 

\bottomrule
\end{tabular}
 }
\caption{Direct win-rate comparisons for RLOO and DPO models on Dolly. RLOO models consistently outperform their DPO counterparts across all dataset splits.} 
\label{tab:rloo-vs-dpo}
\end{table*}

\subsection{RLOO vs DPO}
To provide a head-to-head comparison between DPO and RLOO, Table \ref{tab:rloo-vs-dpo} shows the win-rate evaluation in open-ended generations between models trained with RLOO method with the models trained with DPO.

\subsection{XLSum Summarization}
Table \ref{tab:xlsum-overall} shows the win-rate scores of preference-trained models on 15 languages that are covered by our 23 language list. Win-rates are measured against the original Aya-23-8B model \citep{aryabumi2024aya}.  The average generation length for the base model, the best DPO, and the best RLOO models are 138, 234, and 171 tokens respectively. Length bias is a known property of DPO \citep{park2024dpolength} and can bias GPT-4 as an evaluator \citep{saito2023verbosity} accordingly. Because the base model and the RLOO model generation are similar in length, it is unlikely that the large gains in win-rate for the RLOO model against the base model are caused by GPT-4 as a judge preferring longer responses.
\begin{table}[t]
\centering
\small
\begin{tabular}{ll|rrr|}
\toprule
 &  & \multicolumn{3}{c}{\textbf{Average 15 Languages}}    \\
 & &
\multicolumn{1}{c}{\textbf{Win\%}} & \multicolumn{1}{c}{\textbf{Loss\%}} & \multicolumn{1}{c}{$\Delta$\textbf{W-L\%}} \\ 

\midrule

& \textsc{EN-1}    & 
\multicolumn{1}{r}{52.5} & \multicolumn{1}{r}{41.6} & \multicolumn{1}{c}{\cellcolor[HTML]{EFF9F4}10.9} \\

& \textsc{ML-5}  & 
\multicolumn{1}{r}{50.5} & \multicolumn{1}{r}{43.4} & \multicolumn{1}{c}{\cellcolor[HTML]{FFFFFF}7.1} \\ 

& \textsc{ML-23}   & 
\multicolumn{1}{r}{53.0} & \multicolumn{1}{r}{40.9} & \multicolumn{1}{c}{\cellcolor[HTML]{EAF7F0}12.1} \\ 

\multirow{-4}{*}{\textsc{DPO}} & \textsc{ML-23*} & 
\multicolumn{1}{r}{56.7} & \multicolumn{1}{r}{37.5} & \multicolumn{1}{c}{\cellcolor[HTML]{CAEADB}19.2} \\  

\midrule

& \textsc{EN-1}    & 
\multicolumn{1}{r}{58.0} & \multicolumn{1}{r}{35.8} & \multicolumn{1}{c}{\cellcolor[HTML]{BDE5D1}22.2} \\

& \textsc{ML-5}  & 
\multicolumn{1}{r}{65.1} & \multicolumn{1}{r}{30.2} & \multicolumn{1}{c}{\cellcolor[HTML]{86CEAB}34.9} \\ 

& \textsc{ML-23}  & 
\multicolumn{1}{r}{70.7} & \multicolumn{1}{r}{25.3} & \multicolumn{1}{c}{\cellcolor[HTML]{57BB8A}45.4} \\ 

\multirow{-4}{*}{\textsc{RLOO}} & \textsc{ML-23*}  & 
\multicolumn{1}{r}{68.9} & \multicolumn{1}{r}{26.7} & \multicolumn{1}{c}{\cellcolor[HTML]{66C194}42.2} \\

\bottomrule
\end{tabular}

\caption{15 language XLSum win-rate results for DPO/RLOO preference optimized Aya 23 8B on multiple training data mixtures: EN-1 (English Only), ML-5 (5 Languages), ML-23 (23 Langauges). All runs are done with 50K total training examples with the exception of ML-23*, which is done with 230K total training examples. We report results for the best  checkpoint across 2 epochs.} 
\label{tab:xlsum-overall}
\end{table}

\section{NLP Benchmark Results}
\label{appendix:disc-results}

We report benchmark results for the Base Aya 23 8B Model and preference-trained Aya 23 8B Models (DPO and RLOO) in Tables \ref{tab:m_mmlu}, \ref{tab:mgsm}, \ref{tab:discriminative-results} for mMMLU, MGSM, and discriminative tasks respectively.

\begin{table}[h]
    \centering
    \small
    \begin{tabular}{llcccc}
        \toprule
        & \multicolumn{4}{c}{Held out tasks (Accuracy \%)} \\
        \cmidrule{2-5}
        Model & XCOPA & XSC & XWG & \textbf{\underline{Avg}}\\
        \midrule
    
        Base Aya 23 8B & 59.8 & 62.3 & 80.7 & 67.6\\

        DPO Aya 23 8B & 59.9 & 62.6 & 80.7 & 67.7\\
        
        RLOO Aya 23 8B  & 59.4 & 62.8 & 81.1 & 67.8\\

        \bottomrule
    \end{tabular}
    
    \caption{Results for \textbf{discriminative unseen (held-out) task} evaluation. Results are reported as the zero-shot performance averaged across all languages of XCOPA, XStoryCloze, and XWinoGrad. DPO and RLOO checkpoints are for ML-23 230K runs}
    \label{tab:discriminative-results}
\end{table}
\begin{table*}
    \centering
\resizebox{0.95\textwidth}{!}{
\begin{tabular}{@{}lllllllllllllllll@{}}
\toprule
& en & ar & de & es & fr & hi & id & it & nl & pt & ro & ru & uk & vi & zh & \textbf{\underline{Avg}} \\
\midrule

Base Aya 23 8B & 54.6 & 45.1 & 50 & 50.9 & 51 & 39.7 & 48.8 & 50.7 & 49.7 & 50.8 & 49.9 & 47.8 & 46.8 & 46.5 & 47.1 & 48.2 \\

DPO Aya 23 8B & 54.9 & 45.7 & 50.0 & 51.1 & 51.3 & 40.0 & 49.0 & 51.2 & 49.8 & 51.1 & 49.9 & 48.0 & 47.0 & 46.8 & 47.6 & 48.5 \\

RLOO Aya 23 8B & 54.0 & 45.2 & 50.0 & 50.5 & 50.4 & 39.8 & 48.6 & 50.3 & 49.1 & 50.47 & 49.48 & 47.79 & 46.64 & 46.49 & 47.1 & 48.0 \\

\bottomrule
\end{tabular}
}
\caption{\textbf{Multilingual MMLU} (\texttt{5-shot}) results for base, DPO, and RLOO Aya 23 models.
}
\label{tab:m_mmlu}
\end{table*}

\begin{table*}
    \centering
    \resizebox{0.6\textwidth}{!}{
\begin{NiceTabular}{@{}lllllllll@{}}
\toprule
& de & en & es & fr & ja & ru & zh & \textbf{\underline{Avg}} \\
\midrule

Base Aya 23 8B & 40.4 & 48.0 & 45.2 & 38.8 & 12.8 & 38.0 & 32.8 & 36.6 \\

DPO Aya 23 8B & 39.6 & 45.6 & 44.4 & 41.2 & 8.4 & 37.6 & 35.2 & 36.1\\

RLOO Aya 23 8B & 39.6 & 46.4 & 38.4 & 39.6 & 14.0 & 34.8 & 33.2 & 35.1 \\

\bottomrule
\end{NiceTabular}
}
\caption{\textbf{Multilingual Grade School Math benchmark (MGSM)} results for . We use questions with answers followed by CoT prompt (\texttt{5-shot}) in the same language (\texttt{native\_cot}) as the dataset and \texttt{strict-match} score as the evaluation metric.}
\label{tab:mgsm}
\end{table*}

\section{Judge Prompt}
\label{appendix:judge-prompt}
\textbf{System preamble}:\\
\emph{You are a helpful following assistant whose goal is to select the preferred (least wrong) output for a given instruction in [LANGUAGE\_NAME].} 
\\ \\
\noindent \textbf{Prompt Template}:\\
\emph{Which of the following answers is the best one for given instruction in [LANGUAGE\_NAME]. A good answer should follow these rules:\\
1) It should be in [LANGUAGE\_NAME] \\
2) It should answer the request in the instruction \\
3) It should be factually and semantically comprehensible \\
4) It should be grammatically correct and fluent.}

\emph{Instruction: [INSTRUCTION]} \\
\emph{Answer (A): [COMPLETION A]} \\
\emph{Answer (B): [COMPLETION A]} 

\emph{FIRST provide a one-sentence comparison of the two answers, explaining which you prefer and why. SECOND, on a new line, state only `Answer (A)' or `Answer (B)' to indicate your choice. If the both answers are equally good or bad, state `TIE'. Your response should use the format:}

\emph{Comparison: <one-sentence comparison and explanation>}

\emph{Preferred: <`Answer (A)' or `Answer (B)' or `TIE'>}

\section{Full Language Set Win-Rates}
We provide full win-rate results broken down for all 23 languages against the Base Aya 23 8B Model for the ML-23-230K DPO run in Table \ref{tab:dpo-all} and the the ML-23-230K RLOO run in Table \ref{tab:rloo-all}. Additionally we provide win-rate results for the ML-23-230K RLOO run against Gemma-1.1-7B-it, Llama-3-8B-Instruct, and Mistral-7B-Instruct-v0.3 in Tables \ref{tab:gemma-all}, \ref{tab:llama-all}, and \ref{tab:mistral-all} respectively.

\begin{table}[t]
\small
\centering
\begin{tabular}{lrrrr}
\hline
& \multicolumn{1}{l}{\textbf{Win (\%)}} & \multicolumn{1}{l}{\textbf{Tie (\%)}} & \multicolumn{1}{l}{\textbf{Loss (\%)}} & \multicolumn{1}{l}{\textbf{$\Delta$W-L (\%)}} \\ \hline
en & 0.5300 & 0.0750 & 0.3950 & \cellcolor[HTML]{FFFFFF}0.1350 \\
vi & 0.6100 & 0.0400 & 0.3500 & \cellcolor[HTML]{DFF3E9}0.2600 \\
tr & 0.7300 & 0.0450 & 0.2250 & \cellcolor[HTML]{A1D9BD}0.5050 \\
pt & 0.7150 & 0.0100 & 0.2750 & \cellcolor[HTML]{B1E0C9}0.4400 \\
de & 0.6500 & 0.0650 & 0.2850 & \cellcolor[HTML]{C5E8D6}0.3650 \\
ar & 0.7950 & 0.0550 & 0.1500 & \cellcolor[HTML]{7DCBA4}0.6450 \\
cs & 0.7450 & 0.0350 & 0.2200 & \cellcolor[HTML]{9BD7BA}0.5250 \\
el & 0.8850 & 0.0200 & 0.0950 & \cellcolor[HTML]{57BB8A}0.7900 \\
es & 0.5850 & 0.0500 & 0.3650 & \cellcolor[HTML]{EAF7F0}0.2200 \\
fa & 0.8400 & 0.0550 & 0.1050 & \cellcolor[HTML]{66C194}0.7350 \\
fr & 0.6050 & 0.0600 & 0.3350 & \cellcolor[HTML]{DDF1E7}0.2700 \\
he & 0.7800 & 0.0400 & 0.1800 & \cellcolor[HTML]{88CFAC}0.6000 \\
hi & 0.6250 & 0.0700 & 0.3050 & \cellcolor[HTML]{D0ECDE}0.3200 \\
id & 0.7850 & 0.0000 & 0.2150 & \cellcolor[HTML]{90D2B2}0.5700 \\
it & 0.6500 & 0.0500 & 0.3000 & \cellcolor[HTML]{C8E9D9}0.3500 \\
ja & 0.6750 & 0.0450 & 0.2800 & \cellcolor[HTML]{BDE5D1}0.3950 \\
ko & 0.6400 & 0.0450 & 0.3150 & \cellcolor[HTML]{CFECDE}0.3250 \\
nl & 0.6900 & 0.0200 & 0.2900 & \cellcolor[HTML]{BCE4D0}0.4000 \\
pl & 0.7350 & 0.0200 & 0.2450 & \cellcolor[HTML]{A4DBC0}0.4900 \\
ro & 0.7550 & 0.0350 & 0.2100 & \cellcolor[HTML]{96D5B6}0.5450 \\
ru & 0.7100 & 0.0200 & 0.2700 & \cellcolor[HTML]{B1E0C9}0.4400 \\
uk & 0.6600 & 0.0550 & 0.2850 & \cellcolor[HTML]{C2E7D5}0.3750 \\
zh & 0.5900 & 0.0500 & 0.3600 & \cellcolor[HTML]{E7F6EF}0.2300 \\
\hline
\end{tabular}
\caption{RLOO ML-23-230K vs Gemma-1.1-7B-It}
\label{tab:gemma-all}
\end{table}
\begin{table}[t]
\small
\centering
\begin{tabular}{lrrrr}
\hline
& \multicolumn{1}{l}{\textbf{Win (\%)}} & \multicolumn{1}{l}{\textbf{Tie (\%)}} & \multicolumn{1}{l}{\textbf{Loss (\%)}} & \multicolumn{1}{l}{\textbf{$\Delta$W-L (\%)}} \\ \hline
en & 0.3750 & 0.0350 & 0.5900 & \cellcolor[HTML]{E67C73}-0.2150 \\
vi & 0.7550 & 0.0300 & 0.2150 & \cellcolor[HTML]{90D2B2}0.5400  \\
tr & 0.8400 & 0.0100 & 0.1500 & \cellcolor[HTML]{71C69C}0.6900  \\
pt & 0.5300 & 0.0450 & 0.4250 & \cellcolor[HTML]{EAF7F0}0.1050  \\
de & 0.6400 & 0.0400 & 0.3200 & \cellcolor[HTML]{BEE5D2}0.3200  \\
ar & 0.8350 & 0.0300 & 0.1350 & \cellcolor[HTML]{6FC59B}0.7000  \\
cs & 0.8250 & 0.0300 & 0.1450 & \cellcolor[HTML]{73C79E}0.6800  \\
el & 0.8100 & 0.0600 & 0.1300 & \cellcolor[HTML]{73C79E}0.6800  \\
es & 0.5150 & 0.0450 & 0.4400 & \cellcolor[HTML]{F0F9F5}0.0750  \\
fa & 0.8200 & 0.0450 & 0.1350 & \cellcolor[HTML]{72C69D}0.6850  \\
fr & 0.4600 & 0.0550 & 0.4850 & \cellcolor[HTML]{FCEFEE}-0.0250 \\
he & 0.8950 & 0.0150 & 0.0900 & \cellcolor[HTML]{5ABC8C}0.8050  \\
hi & 0.7800 & 0.0650 & 0.1550 & \cellcolor[HTML]{7FCBA6}0.6250  \\
id & 0.7050 & 0.0200 & 0.2750 & \cellcolor[HTML]{A7DCC2}0.4300  \\
it & 0.5600 & 0.0400 & 0.4000 & \cellcolor[HTML]{DFF2E9}0.1600  \\
ja & 0.8700 & 0.0450 & 0.0850 & \cellcolor[HTML]{5EBE8F}0.7850  \\
ko & 0.8900 & 0.0350 & 0.0750 & \cellcolor[HTML]{57BB8A}0.8150  \\
nl & 0.6900 & 0.0300 & 0.2800 & \cellcolor[HTML]{ABDDC5}0.4100  \\
pl & 0.8350 & 0.0200 & 0.1450 & \cellcolor[HTML]{71C69C}0.6900  \\
ro & 0.6750 & 0.0250 & 0.3000 & \cellcolor[HTML]{B2E0CA}0.3750  \\
ru & 0.7000 & 0.0300 & 0.2700 & \cellcolor[HTML]{A7DCC2}0.4300  \\
uk & 0.8150 & 0.0200 & 0.1650 & \cellcolor[HTML]{7AC9A2}0.6500  \\
zh & 0.8400 & 0.0500 & 0.1100 & \cellcolor[HTML]{69C397}0.7300  \\
\hline
\end{tabular}
\caption{RLOO ML-23-230K vs Llama-3-8B-Instruct}
\label{tab:llama-all}
\end{table}
\begin{table}[t]
\small
\centering
\begin{tabular}{lrrrr}
\hline
& \multicolumn{1}{l}{\textbf{Win (\%)}} & \multicolumn{1}{l}{\textbf{Tie (\%)}} & \multicolumn{1}{l}{\textbf{Loss (\%)}} & \multicolumn{1}{l}{\textbf{$\Delta$W-L (\%)}} \\ \hline
en & 0.3650 & 0.0600 & 0.5750 & \cellcolor[HTML]{E67C73}-0.2100 \\
vi & 0.9050 & 0.0250 & 0.0700 & \cellcolor[HTML]{6EC59A}0.8350  \\
tr & 0.8800 & 0.0250 & 0.0950 & \cellcolor[HTML]{77C8A0}0.7850  \\
pt & 0.6050 & 0.0300 & 0.3650 & \cellcolor[HTML]{D6EFE2}0.2400  \\
de & 0.7400 & 0.0250 & 0.2350 & \cellcolor[HTML]{A8DCC2}0.5050  \\
ar & 0.9150 & 0.0550 & 0.0300 & \cellcolor[HTML]{65C194}0.8850  \\
cs & 0.7600 & 0.0450 & 0.1950 & \cellcolor[HTML]{9DD8BB}0.5650  \\
el & 0.9700 & 0.0250 & 0.0050 & \cellcolor[HTML]{57BB8A}0.9650  \\
es & 0.6400 & 0.0350 & 0.3250 & \cellcolor[HTML]{C9E9D9}0.3150  \\
fa & 0.9500 & 0.0300 & 0.0200 & \cellcolor[HTML]{5EBE8F}0.9300  \\
fr & 0.5800 & 0.0550 & 0.3650 & \cellcolor[HTML]{DAF0E5}0.2150  \\
he & 0.9350 & 0.0350 & 0.0300 & \cellcolor[HTML]{62C092}0.9050  \\
hi & 0.9450 & 0.0250 & 0.0300 & \cellcolor[HTML]{60BF91}0.9150  \\
id & 0.8500 & 0.0200 & 0.1300 & \cellcolor[HTML]{82CDA8}0.7200  \\
it & 0.7050 & 0.0400 & 0.2550 & \cellcolor[HTML]{B1E0C9}0.4500  \\
ja & 0.7900 & 0.0550 & 0.1550 & \cellcolor[HTML]{91D3B3}0.6350  \\
ko & 0.8650 & 0.0500 & 0.0850 & \cellcolor[HTML]{78C9A1}0.7800  \\
nl & 0.7200 & 0.0100 & 0.2700 & \cellcolor[HTML]{B1E0C9}0.4500  \\
pl & 0.7450 & 0.0450 & 0.2100 & \cellcolor[HTML]{A2DABF}0.5350  \\
ro & 0.7700 & 0.0300 & 0.2000 & \cellcolor[HTML]{9CD7BA}0.5700  \\
ru & 0.7150 & 0.0250 & 0.2600 & \cellcolor[HTML]{B0DFC8}0.4550  \\
uk & 0.7600 & 0.0500 & 0.1900 & \cellcolor[HTML]{9CD7BA}0.5700  \\
zh & 0.7100 & 0.0500 & 0.2400 & \cellcolor[HTML]{AEDEC7}0.4700  \\
\hline
\end{tabular}
\caption{RLOO ML-23-230K vs Mistral-7B-Instruct v0.3}
\label{tab:mistral-all}
\end{table}

\begin{table}[t]
\small
\centering
\begin{tabular}{lrrrr}
\hline
& \multicolumn{1}{l}{\textbf{Win (\%)}} & \multicolumn{1}{l}{\textbf{Tie (\%)}} & \multicolumn{1}{l}{\textbf{Loss (\%)}} & \multicolumn{1}{l}{\textbf{$\Delta$W-L (\%)}} \\ \hline
en & 57.5                                  & 11.5                                  & 31.0                                   & \cellcolor[HTML]{57BB8A}26.5               \\
vi & 54.5                                  & 10.5                                  & 35.0                                   & \cellcolor[HTML]{84CDA9}19.5               \\
tr & 47.5                                  & 13.0                                  & 39.5                                   & \cellcolor[HTML]{CDEBDC}8.0                \\
pt & 51.0                                  & 7.0                                   & 42.0                                   & \cellcolor[HTML]{C6E8D8}9.0                \\
de & 50.0                                  & 10.0                                  & 40.0                                   & \cellcolor[HTML]{C0E6D3}10.0               \\
ar & 50.0                                  & 12.0                                  & 38.0                                   & \cellcolor[HTML]{B3E1CB}12.0               \\
cs & 45.0                                  & 11.5                                  & 43.5                                   & \cellcolor[HTML]{F6FCF9}1.5                \\
el & 48.0                                  & 6.5                                   & 45.5                                   & \cellcolor[HTML]{F0F9F4}2.5                \\
es & 39.5                                  & 10.5                                  & 50.0                                   & \cellcolor[HTML]{E67C73}-10.5              \\
fa & 51.5                                  & 10.5                                  & 38.0                                   & \cellcolor[HTML]{AADDC4}13.5               \\
fr & 47.0                                  & 10.0                                  & 43.0                                   & \cellcolor[HTML]{E6F5EE}4.0                \\
he & 48.5                                  & 10.5                                  & 41.0                                   & \cellcolor[HTML]{D0ECDE}7.5                \\
hi & 57.5                                  & 10.5                                  & 32.0                                   & \cellcolor[HTML]{5EBE8F}25.5               \\
id & 52.5                                  & 13.0                                  & 34.5                                   & \cellcolor[HTML]{8DD1B0}18.0               \\
it & 50.5                                  & 11.0                                  & 38.5                                   & \cellcolor[HTML]{B3E1CB}12.0               \\
ja & 53.0                                  & 15.5                                  & 31.5                                   & \cellcolor[HTML]{77C8A1}21.5               \\
ko & 51.0                                  & 12.5                                  & 36.5                                   & \cellcolor[HTML]{A4DABF}14.5               \\
nl & 49.0                                  & 14.5                                  & 36.5                                   & \cellcolor[HTML]{B0DFC8}12.5               \\
pl & 50.5                                  & 10.0                                  & 39.5                                   & \cellcolor[HTML]{BAE3CF}11.0               \\
ro & 58.0                                  & 8.5                                   & 33.5                                   & \cellcolor[HTML]{64C193}24.5               \\
ru & 46.5                                  & 8.0                                   & 45.5                                   & \cellcolor[HTML]{F9FDFB}1.0                \\
uk & 52.0                                  & 8.0                                   & 40.0                                   & \cellcolor[HTML]{B3E1CB}12.0               \\
zh & 45.0                                  & 13.0                                  & 42.0                                   & \cellcolor[HTML]{ECF8F2}3.0                \\ \hline
\end{tabular}
\caption{All language results for DPO ML-23-230K}
\label{tab:dpo-all}
\end{table}
\begin{table}[t]
\centering
\small
\begin{tabular}{lrrrr}
\hline
   & \multicolumn{1}{l}{\textbf{Win (\%)}} & \multicolumn{1}{l}{\textbf{Tie (\%)}} & \multicolumn{1}{l}{\textbf{Loss (\%)}} & \multicolumn{1}{l}{\textbf{$\Delta$W-L (\%)}} \\ \hline
en & 53.0                                  & 12.0                                  & 35.0                                   & 18.0                                       \\
vi & 58.5                                  & 6.5                                   & 35.0                                   & \cellcolor[HTML]{A0D9BD}23.5               \\
tr & 54.5                                  & 10.0                                  & 35.5                                   & \cellcolor[HTML]{BAE3CF}19.0               \\
pt & 54.5                                  & 11.0                                  & 34.5                                   & \cellcolor[HTML]{B4E1CB}20.0               \\
de & 54.0                                  & 10.5                                  & 35.5                                   & \cellcolor[HTML]{BDE5D1}18.5               \\
ar & 49.5                                  & 12.5                                  & 38.0                                   & \cellcolor[HTML]{E5F5ED}11.5               \\
cs & 57.5                                  & 8.0                                   & 34.5                                   & \cellcolor[HTML]{A3DABF}23.0               \\
el & 50.5                                  & 7.0                                   & 42.5                                   & \cellcolor[HTML]{FAFDFB}8.0                \\
es & 55.5                                  & 8.0                                   & 36.5                                   & \cellcolor[HTML]{BAE3CF}19.0               \\
fa & 56.0                                  & 12.0                                  & 32.0                                   & \cellcolor[HTML]{9DD8BB}24.0               \\
fr & 49.5                                  & 8.0                                   & 42.5                                   & \cellcolor[HTML]{FFFFFF}7.0                \\
he & 56.0                                  & 8.0                                   & 36.0                                   & \cellcolor[HTML]{B4E1CB}20.0               \\
hi & 62.0                                  & 12.0                                  & 26.0                                   & \cellcolor[HTML]{57BB8A}36.0               \\
id & 49.5                                  & 9.5                                   & 41.0                                   & \cellcolor[HTML]{F7FCF9}8.5                \\
it & 51.0                                  & 10.0                                  & 39.0                                   & \cellcolor[HTML]{E3F4EB}12.0               \\
ja & 58.5                                  & 10.5                                  & 31.0                                   & \cellcolor[HTML]{89CFAD}27.5               \\
ko & 50.5                                  & 9.5                                   & 40.0                                   & \cellcolor[HTML]{EBF7F1}10.5               \\
nl & 49.0                                  & 10.0                                  & 41.0                                   & \cellcolor[HTML]{FAFDFB}8.0                \\
pl & 52.5                                  & 5.5                                   & 42.0                                   & \cellcolor[HTML]{EBF7F1}10.5               \\
ro & 54.0                                  & 11.0                                  & 35.0                                   & \cellcolor[HTML]{BAE3CF}19.0               \\
ru & 51.5                                  & 6.0                                   & 42.5                                   & \cellcolor[HTML]{F4FBF7}9.0                \\
uk & 50.0                                  & 14.0                                  & 36.0                                   & \cellcolor[HTML]{D7EFE3}14.0               \\
zh & 50.5                                  & 10.0                                  & 39.5                                   & \cellcolor[HTML]{E8F6EF}11.0               \\ \hline
\end{tabular}
\caption{All language results for RLOO ML-23 230K}
\label{tab:rloo-all}
\end{table}

\section{Language List}

We provide a list and description of all languages supported by Aya 23 8B which we use to perform multilingual evaluations in Table \ref{tab:language_codes}.
\begin{table*}[t!]
\centering
\begin{tabular}{lllll}
\toprule
Code & Language & Script & Family & Subgrouping \\
\midrule 
ar & Arabic      & Arabic       & Afro-Asiatic  & Semitic    \\
cs & Czech       & Latin        & Indo-European & Balto-Slavic \\
de & German      & Latin        & Indo-European & Germanic  \\
el & Greek       & Greek        & Indo-European & Graeco-Phrygian \\
en & English     & Latin        & Indo-European & Germanic  \\
es & Spanish     & Latin        & Indo-European & Italic   \\
fa & Persian     & Arabic       & Indo-European & Iranian  \\
fr & French      & Latin        & Indo-European & Italic  \\
he & Hebrew       & Hebrew      & Afro-Asiatic & Semitic   \\
hi & Hindi       & Devanagari   & Indo-European & Indo-Aryan \\
id & Indonesian  & Latin        & Austronesian  & Malayo-Polynesian  \\
it & Italian     & Latin        & Indo-European & Italic \\
jp & Japanese    & Japanese     & Japonic       & Japanesic  \\
ko & Korean      & Hangul       & Koreanic      & Korean   \\
nl & Dutch       & Latin        & Indo-European & Germanic \\
pl & Polish      & Latin        & Indo-European & Balto-Slavic  \\
pt & Portuguese  & Latin        & Indo-European & Italic   \\
ro & Romanian    & Latin        & Indo-European & Italic   \\
ru & Russian     & Cyrillic     & Indo-European & Balto-Slavic  \\
tr & Turkish     & Latin        & Turkic        & Common Turkic  \\
uk & Ukrainian   & Cyrillic     & Indo-European & Balto-Slavic   \\
vi & Vietnamese  & Latin        & Austroasiatic & Vietic   \\
zh & Chinese     & Han \& Hant          & Sino-Tibetan  & Sinitic \\
\bottomrule
\end{tabular}
\caption{23 languages supported in Aya 23 model \cite{aryabumi2024aya} with each language's script, family, and subgrouping}
\label{tab:language_codes}
\end{table*}

\end{document}